%% file: pitra2022ec.tex
\documentclass[twoside]{article}
\usepackage{ecj,palatino,epsfig,latexsym,natbib}

\usepackage{amsmath}
\usepackage{amsfonts}
\usepackage{mathtools}

\usepackage{algorithmic}
\usepackage{algorithm}

\usepackage{graphicx}
\usepackage{booktabs}
\usepackage{multirow}
\usepackage{colortbl}
\usepackage{rotating}      
\usepackage{pgffor}        
\usepackage{calc}          
\usepackage{bm}
\usepackage{xfrac}         
\usepackage{hyperref}      
\hypersetup{
  hidelinks=true
}

\usepackage{pdftexcmds}    
\usepackage{boolexpr}      

\usepackage{tikz}
\usetikzlibrary{snakes,arrows,shapes}
\tikzset{rectangle/.append style={inner sep=7pt}}

\definecolor{InnerNodeColor}{HTML}{FFF2D3}
\definecolor{LeafColor}{HTML}{D1D8F6}

\newcommand{\eg}{e.\,g.,\ }
\newcommand{\ie}{i.\,e.,\ }
\newcommand{\aka}{a.\,k.\,a.\ }

\newcommand{\vmat}[1]{\mathrm{\mathbf{#1}}} 
\newcommand{\set}[1]{\mathcal{#1}}          
\newcommand{\dis}[1]{\mathcal{#1}}          
\newcommand{\norm}[1]{\left\|#1\right\|}    
\newcommand{\bignorm}[1]{\big\|{#1}\big\|}  
\newcommand{\BB}{\ensuremath{\mbox{$\mathfrak{b\!b}$}}} 
\newcommand{\natlog}{\ensuremath{\ln}}                  
\newcommand{\ev}{\ensuremath{\mathbb{E}}}               

\newcommand{\dm}{D}  
\newcommand{\npt}{N} 
\newcommand{\gen}{g} 
\newcommand{\lscale}{\ell}                

\newcommand{\RR}{\mathbb{R}}    
\newcommand{\NN}{\mathbb{N}}    

\newcommand{\xx}{\vmat{x}}
\newcommand{\yy}{\vmat{y}}
\newcommand{\XX}{\vmat{X}}
\newcommand{\XXs}{\set{X}}
\newcommand{\sS}{\set{S}}

\newcommand{\gn} [1]{ {#1^{(\gen)}} }   
\newcommand{\gnd}[2]{ {#1^{(#2)}} }     
\newcommand{\gnp}[1]{ {#1^{(\gen+1)}} } 
\newcommand{\CC}{\vmat{C}}              
\newcommand{\mm}{\vmat{m}}              

\newcommand{\feat}[1]{\ensuremath{\varphi_{#1}}} 
\newcommand{\nanin}{\circ}                   
\newcommand{\nanout}{\bullet}                

\newcommand{\archive}{\set{A}}             
\newcommand{\archivepred}{\archive_\predset} 
\newcommand{\trainset}{\set{T}}            
\newcommand{\predset}{\set{P}}             
\newcommand{\trainpredset}{\trainset_\predset} 
\newcommand{\testset}{\XX_\text{te}}       
\newcommand{\transnone}{{}}                
\newcommand{\transcma}{{}_\top}            
\newcommand{\trans}[1]{#1^{\transcma}}     

\newcommand{\featset}[1]{\Phi_\text{#1}}   
\newcommand{\ftBasic}{Basic}               
\newcommand{\ftCMA}{CMA}                   
\newcommand{\ftyDis}{y-D}                  
\newcommand{\ftLevel}{Lvl}                 
\newcommand{\ftMM}{MM}                     
\newcommand{\ftNBC}{NBC}                   
\newcommand{\ftInfo}{Inf}                  
\newcommand{\ftDisp}{Dis}                  

\newcommand{\featBasic}{\supptextref{\featset{\ftBasic}}{5}}    
\newcommand{\featCMA}{\supptextref{\featset{\ftCMA}}{6}}        
\newcommand{\featyDis}{\supptextref{\featset{\ftyDis}}{12}}     
\newcommand{\featLevel}{\supptextref{\featset{\ftLevel}}{9}}    
\newcommand{\featMM}{\supptextref{\featset{\ftMM}}{10}}         
\newcommand{\featNBC}{\supptextref{\featset{\ftNBC}}{11}}       
\newcommand{\featInfo}{\supptextref{\featset{\ftInfo}}{8}}      
\newcommand{\featDisp}{\supptextref{\featset{\ftDisp}}{7}}      

\newcommand{\artFeat}[4]{
  \def\tempset{#2}
  \def\tempftset{#4}
  \ifx\tempftset\empty
    \ifx\tempset\empty
      {\supptextref{\feat{\texttt{#1}}}{\supppage{\ftBasic}}}
    \else
      {\supptextref{\feat{\texttt{#1}}}{\supppage{\ftBasic}}\ensuremath{(#2^{#3})}}
    \fi
  \else
    \ifx\tempset\empty
      {\supptextref{\ensuremath{\feat{\texttt{#1}}^\text{#4}}}{\supppage{#4}}}
    \else
      {\supptextref{\ensuremath{\feat{\texttt{#1}}^\text{#4}}}{\supppage{#4}}\ensuremath{(#2^{#3})}}
    \fi
  \fi
}
\newcommand{\tableFeat}[4]{
  \artFeat{#1}{#2}{#3}{#4}
}

\newcommand{\ym}{\hat{{y}}}    
\newcommand{\model}{\set{M}} 
\newcommand{\modelsettings}{\boldsymbol{\psi}} 
\newcommand{\btheta}{\boldsymbol{\theta}} 
\newcommand{\tss}[1]{\protect\hyperlink{tss:#1}{TSS #1}}   
\newcommand{\gp}{\mathop{\rm GP}}
\newcommand{\cov}{\kappa}                   
\newcommand{\covf}{\mathop{\rm cov}}        
\newcommand{\covSE}{\cov_\mathrm{SE}}       
\newcommand{\covMat}[1]{\cov_\mathrm{Mat}^{\frac{#1}{2}}}  
\newcommand{\covNN}{\cov_\mathrm{NN}}       
\newcommand{\covRQ}{\cov_\mathrm{RQ}}       
\newcommand{\covLIN}{\cov_\mathrm{LIN}}     
\newcommand{\covQUAD}{\cov_\mathrm{Q}}   
\newcommand{\covSEQUAD}{\cov_\mathrm{SE+Q}} 
\newcommand{\covGIBBS}{\cov_\mathrm{Gibbs}} 
\newcommand{\splitf}{s}         
\newcommand{\splitS}{\sS}       

\newcommand{\mse}{\mathop{\rm MSE}}
\newcommand{\rde}{\mathop{\rm RDE}}

\newcommand{\refFormat}[1]{\textsc{#1}}
\newcommand{\sourceurl}{https://github.com/bajeluk/surrogate-cmaes/tree/meta}
\newcommand{\suppurl}{https://raw.githubusercontent.com/jdgregorian/surrogate-cmaes-pages/gh-pages/supp/ec2022/supp_mat.pdf}
\newcommand{\suppmat}[1]{\href{\suppurl}{\refFormat{#1}}}
\newcommand{\suppref}[2]{\href{\suppurl\#page=#2}{\refFormat{#1}}}
\newcommand{\supptextref}[2]{\href{\suppurl\#page=#2}{#1}}
\newcommand{\supppage}[1]{\switch
    \case{\pdfstrcmp{#1}{\ftBasic}}5
    \case{\pdfstrcmp{#1}{\ftCMA}}6
    \case{\pdfstrcmp{#1}{\ftDisp}}7
    \case{\pdfstrcmp{#1}{\ftInfo}}8
    \case{\pdfstrcmp{#1}{\ftLevel}}9
    \case{\pdfstrcmp{#1}{\ftMM}}10
    \case{\pdfstrcmp{#1}{\ftNBC}}11
    \case{\pdfstrcmp{#1}{\ftyDis}}12
  \endswitch
}

\graphicspath{{images/}}

\begin{document}

\ecjHeader{x}{x}{xxx-xxx}{201X}{Landscape Analysis for Surrogate Models in the Evol. B-B Context}{Z. Pitra, J. Koza, J. Tumpach, M. Hole\v{n}a}
\title{\bf Landscape Analysis for Surrogate Models in the Evolutionary Black-Box Context}  

\author{
        \name{\bf Zbyn\v{e}k Pitra} \hfill \addr{z.pitra@gmail.com}\\ 
        \addr{Faculty of Nuclear Sciences and Physical Engineering, Czech Technical University\\
          B\v{r}ehov\'{a} 7, 115\,19 Prague, Czech Republic
        }
\AND
       \name{\bf Jan Koza} \hfill \addr{koza@cs.cas.cz}\\
       \addr{Faculty of Information Technology, CTU in Prague, \\
         Th\'{a}kurova 9, 160\,00 Prague 6, Czech Republic
       }
\AND
       \name{\bf Ji\v{r}\'{i} Tumpach} \hfill \addr{tumpach@cs.cas.cz}\\
       \addr{Faculty of Mathematics and Physics, Charles University in Prague,\\
         Malostran. n\'{a}m. 25, 118\,00 Prague, Czech Republic
       }
\AND
       \name{\bf Martin Hole\v{n}a} \hfill \addr{martin@cs.cas.cz}\\
       \addr{Institute of Computer Science, Czech Academy of Sciences,\\
         Pod Vod\'{a}renskou v\v{e}\v{z}\'{i} 2, 182\,07 Prague, Czech Republic
       }
}

\maketitle

\begin{abstract}

Surrogate modeling has become a valuable
technique for black-box optimization tasks
with expensive evaluation of the objective function.
In this paper,
we investigate the relationship between the predictive accuracy of surrogate models
and features of the black-box function landscape.
We also study properties of features for landscape analysis
in the context of different transformations and ways of selecting the input data.
We perform the landscape analysis of a large set of data
generated using runs of a surrogate-assisted version of
the Covariance Matrix Adaptation Evolution Strategy
on the noiseless part of the Comparing Continuous Optimisers benchmark function testbed.

\end{abstract}


\section{Introduction}
\label{sec:intro}

When solving a real-world optimization problem
we often have no information about the analytic form of the objective function.
Evaluation of such \emph{black-box functions}
is frequently expensive in terms of time and money
\citep{baerns2009combinatorial,lee2016surrogate,zaefferer2016multi},
which has been for two decades the driving force of research
into \emph{surrogate modeling} of black-box objective functions
\citep{buche2005accelerating,forrester2009recent,jin2011surrogate}.
Given a set of observations, a surrogate model
can be fitted to approximate the landscape of the black-box function.

The \emph{Covariance Matrix Adaptation Evolution Strategy} (CMA-ES) by~\cite{hansen2006cma},
which we consider the state-of-the-art evolutionary black-box
optimizer, has been frequently combined with surrogate
models.
However,
different surrogate models can show significant differences
in the performance of the same optimizer on different data~\citep{pitra2021interaction}.
Such performance can also change during the algorithm run
due to varying landscape of the black-box function
in different parts of the search space.
Moreover, the predictive accuracy of individual models is
strongly influenced by the choice of their parameters.
Therefore, the investigation of the relationships between
the settings of individual models and the optimized data
is necessary for better understanding of the whole surrogate modeling task.

In last few years,
research into landscape analysis of objective functions
(cf. the overview by~\cite{kerschke2017comprehensive})
has emerged in the context of algorithm selection and algorithm configuration.
However, to our knowledge, such features have been
investigated in connection with surrogate models in the context
of black-box optimization using static settings only,
where the model is selected once at the beginning of the optimization process
\citep{saini2019automatic}.
Moreover, the analysis of landscape features
also had less attention so far than it deserves
and not in the context of surrogate models
\citep{renau2019expressiveness, renau2020exploratory}.

In this paper,
we study properties of features representing the fitness landscape
in the context of the data from actual runs of a surrogate-assisted version of the CMA-ES
on the noiseless part of
the Comparing Continuous Optimisers (COCO) benchmark function testbed~\citep{hansen2016coco}.
From the large number of available features,
we select a small number of representatives
for subsequent research,
based on their robustness to sampling and
similarity to other features.
Basic investigation in connection with landscape features of CMA-ES assisted by a surrogate model based on
Gaussian processes~\citep{rasmussen2006gaussian}
was recently presented in a conference paper by~\cite{pitra2019landscape}.
Here, we substantially more thoroughly investigate
the relationships between selected representatives of landscape features
and the error of
several kinds of surrogate models using different settings of their parameters and
the criteria for selecting points for their training.
Such study is crucial due to completely different properties of data from runs of a surrogate-assisted algorithm
compared with generally utilized sampling strategies,
which imply different values of landscape features
as emphasized in~\cite{renau2020exploratory}.

The paper is structured as follows.
Section~\ref{sec:back} briefly summarizes the basics of surrogate modeling in the context of the CMA-ES, surrogate models used in this investigation, and landscape features from the literature.
In Section~\ref{sec:lasm},
properties of landscape features are thoroughly tested
and their connection to surrogate models is analysed.
Section~\ref{sec:concl} concludes the paper and suggests a few future research objectives.

\section{Background}
\label{sec:back}

The goal of \emph{black-box} optimization is to find a point 
For this function,
we are only able to obtain the value $\BB(\xx)$ of an \emph{objective function} $\BB$,
\aka \emph{fitness}, in a point $\xx$,
but no analytic expression is known, neither an explicit one as a composition of known mathematical functions, nor an implicit one as a solution of an
equation (\eg algebraic or differential). 
In case of minimization:
\begin{equation}
 \xx^* = \operatorname{arg\,min}_{\xx \in \sS} \BB(\xx)\,.
\end{equation}

\subsection{Surrogate Modeling in the Context of the CMA-ES}
\label{ssec:sm}

Black-box optimization problems frequently appear in the real world, where the values of a fitness function $\BB(\xx)$ can be
obtained only empirically through experiments or via computer simulations. 
As recalled in the Introduction, such
empirical evaluation is sometimes very time-consuming or
expensive.
Thus, we assume that evaluating the fitness represents a considerably higher cost than training a regression model.

As a \emph{surrogate model}, we understand any regression model $\widehat{\BB}: \XXs \rightarrow \RR$ that is trained
on the already available input--output value pairs
stored in an \emph{archive}
$\archive = \{(\xx_i, y_i) \, | \, y_i = \BB(\xx_i), \, i=1,\dots,N\}$,
and is used instead of
the original expensive fitness to evaluate
some of the points needed by the optimization algorithm.

\subsubsection{The CMA-ES}
\label{sssec:cmaes}

The \emph{Covariance Matrix Adaptation Evolution Strategy} (CMA-ES) proposed 
by~\cite{hansen1996adapting} has become one of the most successful 
algorithms in the field of continuous black-box optimization.
Its pseudocode
is outlined in Algorithm~\ref{alg:wacmaes}. 

\begin{algorithm} [t]
  \begin{algorithmic}[1]

    \REQUIRE{
      $\lambda$ (population-size),
      $\BB$ (original fitness function),
      $(w_i)_{i=1}^\mu$, $\mu_w$ (selection and recombination parameters), 
      $c_\sigma$, $d_\sigma$ (step-size control parameters), 
      $c_c$, $c_I$, $c_\mu$ 
      (covariance matrix adaptation parameters)
    }
    \STATE{
      \textbf{init}
      $\gen = 0,
       \gnd{\sigma}{0} > 0,
       \gnd{\vmat{m}}{0} \in \RR^\dm,
       \gnd{\CC}{0} = \vmat{I},
       \gnd{\vmat{p}_\sigma}{0} = \vmat{0},
       \gnd{\vmat{p}_c}{0} = \vmat{0},
       \mu = \left\lfloor\lambda/2\right\rfloor
      $
    }
    \REPEAT
      \STATE{$\xx_k \leftarrow \gn{\mm} + \gn{\sigma} \dis{N}(\vmat{0}, \gn{\CC})$ \qquad $k = 1, \ldots, \lambda$}
      \label{ln:offspring}
      \STATE{$\BB_k \leftarrow \BB(\xx_k)$ \hspace{2.03cm} $k = 1, \ldots, \lambda$}
      \label{ln:eval}
      \STATE{$\gnp{\mm} \leftarrow \sum_{i=1}^\mu w_i \xx_{i:\lambda}$, where $\xx_{i:\lambda}$ is the $i$-th best individual out of $\xx_1, \ldots, \xx_\lambda$}
      \label{ln:mean}
      \STATE{$\gnp{\vmat{p}_\sigma} \leftarrow (1-c_\sigma)\gn{\vmat{p}_\sigma} +
              \sqrt{c_\sigma (2-c_\sigma) \mu_w}{\gn{\CC}}^{-1/2}\frac{\gnp{\mm}-\gn{\mm}}{\gn{\sigma}}$}
      \label{ln:pathsigma}
      \STATE{$h_\sigma \leftarrow \mathbb{I}\left(\|\gnp{\vmat{p}_\sigma}\| < \sqrt{1-(1-c_\sigma)^{2(\gen + 1)}}
              \left(1.4 + \frac{2}{\dm+1}\right) \ev\|\dis{N}(0,1))\|\right)$}
      \STATE{$\gnp{\vmat{p}_c} \leftarrow (1-c_c)\gn{\vmat{p}_c} + 
              h_\sigma \sqrt{c_c (2-c_c) \mu_w} \, \frac{\gnp{\mm}-\gn{\mm}}{\gn{\sigma}}$}
      \label{ln:pathC}
      \STATE{
        $\CC_\mu \leftarrow \sum_{i=1}^\mu w_i \gn{\sigma}^{-2}
        \big(\gnp{\xx_{i:\lambda}}-\gn{\mm}\big)
        \big(\gnp{\xx_{i:\lambda}}-\gn{\mm}\big)
        ^\top$
      }
      \label{ln:cmu}
      \STATE{$\gnp{\CC} \leftarrow (1-c_1-c_\mu) \gn{\CC} + c_1 \gnp{\vmat{p}_c}\gnp{\vmat{p}_c}^\top + c_\mu \CC_\mu$} 
      \label{ln:updateC}
      \STATE{$
        \gnp{\sigma} \leftarrow \gn{\sigma} 
        \exp\Big(\frac{c_\sigma}{d_\sigma}\Big(\frac{\|\gnp{\vmat{p}_\sigma}\|}{\ev\|\dis{N}(\vmat{0}, \vmat{I})\|} -
1 \Big)\Big)
        $}
      \label{ln:updatesigma}
      \STATE{$\gen \leftarrow \gen + 1$}
    \UNTIL{stopping criterion is met}
    \ENSURE{$\xx_\text{res}$ (resulting optimum)}

  \end{algorithmic}
  \caption{Pseudocode of the CMA-ES}
  \label{alg:wacmaes}
\end{algorithm}

After generating $\lambda$ new candidate solutions using 
the mean $\gn{\mm}$ of the mutation distribution added to 
a random Gaussian mutation with covariance matrix 
$\gn{\CC}$ in generation $\gen$ (step~\ref{ln:offspring}),
the fitness function $\BB$ is evaluated for the new offspring (step~\ref{ln:eval}).
The new mean $\gnp{\mm}$ of the mutation distribution is computed as 
the weighted sum of the $\mu$ best points among 
the $\lambda$ ordered offspring $\xx_1, \ldots, \xx_\lambda$ (step~\ref{ln:mean}).

The sum of consecutive successful mutation steps of the algorithm 
in the search space
$(\gnp{\mm}-\gn{\mm})/\gn{\sigma}$
is utilized to compute two \emph{evolution path} vectors  
 $\vmat{p}_\sigma$ and $\vmat{p}_c$ (step~\ref{ln:pathsigma}).
Successful steps are tracked in the sampling space 
and stored in $\vmat{p}_\sigma$ using the transformation $\gn{\CC}^{-1/2}$.
The evolution 
path $\vmat{p}_c$ is calculated similarly to $\vmat{p}_\sigma$; 
however, the coordinate system is not changed (step~\ref{ln:pathC}).
The two remaining evolution path elements are 
a decay factor $c_\sigma$ decreasing the impact of successful steps with increasing generations, and 
$\mu_w$ used to normalize the variance of $\vmat{p}_\sigma$.

The covariance matrix adaptation (step~\ref{ln:updateC}) is performed using \emph{rank-one}
and \emph{rank-$\mu$} updates.
The rank-one update utilizes $\vmat{p}_c$ to calculate
the covariance matrix $\vmat{p}_c\vmat{p}_c^\top$.
The rank-$\mu$ update cummulates successful steps of $\mu$ best 
individuals in matrix $\CC_\mu$ (step~\ref{ln:cmu}).

The step-size $\sigma$ (step~\ref{ln:updatesigma})
is updated according to the ratio between 
the evolution path length $\|\vmat{p}_\sigma\|$ and the expected length of a random 
evolution path.
If the ratio is greater than 1, the step-size is increasing, and decreasing otherwise.

The CMA-ES uses restart strategies to deal with multimodal fitness landscapes 
and to avoid being trapped in local optima. 
A multi-start strategy where the 
population size is doubled in each restart $n_\text{r}$ is referred to as IPOP-CMA-ES~\citep{auger2005restart}.

\subsubsection{Training Sets}
\label{sssec:tss}

The level of the model precision is highly determined by the selection of the points from the archive to the training set $\trainset$. Considering the surrogate models we use in this paper, we list the following training set selection (TSS) methods:
\begin{itemize}

  \item \hypertarget{tss:full}{\emph{TSS full}}, taking all the already evaluated points, \ie $\trainset=\archive$,

  \item \hypertarget{tss:knn}{\emph{TSS knn}}, selecting the union of the sets of $k$-nearest neighbors of all points for which the fitness should be predicted, where $k$ is
  user defined (\eg in \cite{kern2006local}),

  \item \hypertarget{tss:nearest}{\emph{TSS nearest}}, selecting the union of the sets of $k$-nearest neighbors of all points for which the fitness should be predicted,
    where $k$ is maximal such that the total number of selected points does not exceed a given maximum number of points $N_\text{max}$ and
    no point is further from current mean $\gn{\mm}$ than a given maximal distance $r_\text{max}$ (\eg in \cite{bajer2019gaussian}).
\end{itemize}

The models in this paper use the Mahalanobis distance given by the CMA-ES matrix $\sigma^2 \CC$
(see Equation \ref{eq:amt}) for selecting points into training sets, similarly to, \eg \cite{kruisselbrink2010robust}.
Considering the fact that \tss{knn} is the original TSS method of the model from~\cite{kern2006local} only (see Subsection~\ref{sssec:rs}),
we list it here and utilize it only in connection with this particular model.

\subsubsection{Response Surfaces}
\label{sssec:rs}

Basically, the purpose of surrogate modeling -- to
approximate an unknown functional dependence -- coincides with
the purpose of \emph{response surface modeling} in the design
of experiments
\citep{hosder2001polynomial,myers2009response}. Therefore, it is
not surprising that typical response surface models, \ie
\emph{low order polynomials}, also belong to most traditional
and most successful surrogate models
\citep{rasheed2005methods,kern2006local,auger2013benchmarking,hansen2019global}.
Here, we present two surrogate models using at most quadratic polynomial utilized in two successful surrogate-assisted variants of the CMA-ES:
the local-metamodel-CMA (lmm-CMA) by~\cite{kern2006local} and
the linear-quadratic CMA-ES (lq-CMA-ES) by~\cite{hansen2019global}.


\paragraph{Local-metamodel}
is a specific full quadratic model
$f_j:\RR^\dm\to\RR $ trained for
$\xx_j, j=1, \dots, \lambda$ used in the lmm-CMA~\citep{auger2013benchmarking},
\begin{equation}
  \label{eq:afq}
  f_j(\xx)= (\xx - \xx_j)^\top \vmat{A}_j(\xx-\xx_j)+ (\xx-\xx_j)^\top \vmat{b}_j + c_j
  \text{ with } \vmat{A}_j\in\RR^{\dm\times\dm}, \vmat{b}_j\in\RR^\dm, c_j\in\RR\,.
\end{equation}
The model $f_j$ is trained on the set $N_k
(\xx_j; \archive)$ of a given number $k$ of nearest neighbors
of $\xx_j$ with respect to a given archive $\archive$,
\begin{equation}
  \label{eq:ank}
  N_k(\xx_j; \archive) \subset \archive, |N_k (\xx_j; \archive)| = k \,, (\forall \xx \in N_k(\xx_j;\archive))
  (\forall \xx'\in\archive \setminus N_k (\xx_j; \archive))\, d (\xx,\xx_j)\le d (\xx', \xx_j) \,.
\end{equation}
As the distance $d$ in (\ref{eq:ank}), the Mahalanobis distance
for $\sigma^2\CC$ is used:
\begin{equation}
  \label{eq:amt}
  d_{\sigma^2\CC} (\xx, \yy) = \sqrt{ (\xx-\yy)^\top \sigma^{-2}\CC^{-1}(\xx-\yy) }\,.
\end{equation}

\paragraph{{}}\hspace{-0.2cm}The \emph{linear-quadratic model}
in the algorithm lq-CMA-ES differs from
the lmm-CMA in an important aspect:

Whereas surrogate models in the lmm-CMA are always full
quadratic, the lq-CMA-ES admits also pure quadratic or linear  
models. The employed kind of model depends on
the number of points in $\trainset$
(according to the employed TSS method).

\subsubsection{Gaussian Processes}
\label{sssec:gp}


A Gaussian process (GP) is a collection of random variables
$(f(\xx))_{\xx\in\set{X}}$, $\set{X} \subset \RR^\dm$,
any finite number of which has a joint Gaussian distribution~\citep{rasmussen2006gaussian}.
The GP is completely defined by a mean function $m:\set{X} \to \RR$, typically assumed to be some constant $m_{\gp}$, and by a covariance function $\cov : \set{X} \times \set{X} \to \RR$ such that
\begin{gather}
  \label{eq:amf}
  \ev f(\xx) = m_{\gp}\,, \quad \covf(f(\xx), f(\xx') ) = \cov(\xx, \xx')\,,\quad \xx, \xx'\in\set{X}\,.
\end{gather}
The value of $f(\xx)$ is typically
a noisy observation $y = f(\xx)+\varepsilon$, where
$\varepsilon$ is a zero-mean Gausssian noise with a variance
$\sigma_n^2 > 0$. Then
\begin{gather}
  \label{eq:acn}
  \covf(y,y') = \cov(\xx, \xx') + \sigma_n^2 \mathbb{I}(\xx = \xx')\,,
\end{gather}
where $\mathbb{I}(p)$ equals 1 when a proposition $p$ is true and 0 otherwise.

Consider now the prediction of the random variable $f(\xx_\star)$
in a point $\xx_\star\in\set{X}$ if we already know
$(\BB(\xx_1), \dots, \BB(\xx_n))^\top =
 (f(\xx_1) + \varepsilon_1, \dots, f(\xx_n) + \varepsilon_n)^\top$
in points $\xx_1, \dots, \xx_n$.
Introduce the vectors $\vmat{X} = (\xx_1, \dots, \xx_n)^\top$,
$\yy = (\BB(\xx_1), \dots, \BB(\xx_n))^\top$,
$\vmat{k}_\star = (\cov (\xx_1, \xx_\star),\dots$, $\cov(\xx_n, \xx_\star))^\top$
and the matrix $\vmat{K}\in\RR^{n\times n}$
such that $(\vmat{K})_{i,j}=\cov (\xx_i, \xx_j)$.
Then the probability density of the vector $\yy$ is
\begin{gather}
  \label{eq:apd}
  p (\yy; m(\vmat{X}), \cov, \sigma_n^2) = \frac{\exp \left(-\frac{1}{2}(\yy-m_{\gp}
  )^\top \vmat{K}^{-1} (\yy-m_{\gp}) )\right)}{\sqrt{
  (2\pi)^\dm\det(\vmat{K}+\sigma_n^2 \vmat{I}_n) }} \,,
\end{gather}
where $\det(\vmat{A})$ denotes the determinant of a matrix
$\vmat{A}$. Further, as a consequence of the assumption of Gaussian
joint distribution, also the conditional distribution of $f(\xx_\star)$
conditioned on $\yy$ is Gaussian:
\begin{gather}
  \label{eq:acd}
  \dis{N}(m(\xx_\star) + \vmat{k}_\star  \vmat{K}^{-1} (\yy - m_{\gp}),\,
  \cov(\xx_\star, \xx_\star) - \vmat{k}_\star^\top \vmat{K}^{-1}\vmat{k}_\star)\,.
\end{gather}

\subsubsection{Regression Forest}
\label{sssec:rf}

\emph{Regression forest}~\citep{breiman2001random} (RF) is
an ensemble of regression decision trees~\citep{breiman1984classification}.
Recently,
the \emph{gradient tree boosting}~\citep{friedman2001greedy}
has been shown
useful in connection with the CMA-ES~\citep{pitra2018boosted}.
Therefore, we will focus only on this method.

Let us consider binary regression trees, where
each observation $\xx$
passes through
a series of binary split functions $\splitf$ associated with internal nodes
and arrives in the leaf node containing a real-valued constant
trained to be the prediction of an associated function value $y$.
Let $\ym_i^{(t)}$ be the prediction of the $i$-th point of the $t$-th tree.
The $t$-th tree $f_t$ is obtained in the $t$-th iteration
of the boosting algorithm through optimization of
the following
function:
\begin{equation}
  \label{eq:reg_obj}
  \set{L}^{(t)}  = \sum_{i=1}^\npt l\left(y_i, \ym_i^{(t-1)} + f_t(\xx_i)\right) + \Omega(f_t) \,,
  \quad \text{where} \  \Omega(f)    = \gamma \, T_f + \frac{1}{2}\alpha\norm{w_f}^2\,, \nonumber
\end{equation}
$l$ is a differentiable convex loss function $l:\RR^2\to\RR$,
$T_f$ is the number of leaves in a tree $f$,
$w_f$ are weights of its individual leaves,
and $\alpha, \gamma \geq 0$ are penalization constants.
The gain can be derived using (\ref{eq:reg_obj}) as follows
(see~\cite{chen2016xgboost} for details):
\begin{equation}
\label{eq:xgb_split}
  \set{L}_{\text{split}} = \frac{1}{2} \left[r(\splitS_L) + r(\splitS_R) - r(\splitS_{L+R})\right] - \gamma \,,
  \quad r(\splitS) = \frac{\left(\sum_{y \in \splitS} g(y)\right)^2}{\sum_{y \in \splitS} h(y) + \alpha} \,,
\end{equation}
where
set $\splitS_{L+R}$ is split into $\splitS_L$ and $\splitS_R$,
$g(y) = \partial_{\ym^{(t-1)}}   l(y, \ym^{(t-1)})$ and
$h(y) = \partial_{\ym^{(t-1)}}^2 l(y, \ym^{(t-1)})$
are the first and second order derivatives of the loss function.


The overall boosted forest prediction is obtained through averaging
individual tree predictions, where each leaf $j$
in a $t$-th tree has weight
\begin{equation}
  w_j^{(t)} = - \frac{\sum_{y \in \splitS_j} g(y)}{\sum_{y \in \splitS_j} h(y) + \alpha}\,,
\end{equation}
where $S_j$ is the set of all training inputs that end in the leaf $j$.
As a prevention of overfitting, the random subsampling of input features or traning data can be employed.

\subsection{Landscape Features}
\label{ssec:landscape}

Let us consider a sample set $\sS$ of $\npt$ pairs of observations in the context of continuous black-box optimization
$\sS = \left\{(\xx_i, y_i) \in \RR^\dm \times \RR\cup\{\nanin\} \, \vert \,i = 1, \ldots, \npt \right\}$,
where $\nanin$ denotes missing $y_i$ value (\eg $\xx_i$ was not evaluated yet).
Then the sample set can be utilized to describe landscape properties using a 
\emph{landscape feature}
$\feat{}: \bigcup_{\npt\in\NN} \RR^{\npt, \dm} \times \left(\RR\cup\{\nanin\}\right)^{\npt, 1} \mapsto \RR \cup \{\pm\infty, \nanout\}$,
where $\nanout$ denotes impossibility of feature computation.

A large number of various landscape features have been
proposed in recent years in literature.
\cite{mersmann2011exploratory} proposed six easy to compute feature sets
(each containing a number of individual features for continuous domain)
representing different properties:
  \emph{y-Distribution} set with measures related to the
    distribution of the objective function values,
  \emph{Levelset} features capturing the relative
    position of each value with respect to objective quantiles,
  \emph{Meta-Model} features extracting the information from linear or
    quadratic regression models fitted to the sampled data,
  \emph{Convexity} set describing the level of function landscape convexity,
  \emph{Curvature} set with gradient and
    Hessian approximation statistics,
  and
  \emph{Local Search} features related to
    local searches conducted from sampled points.
The last three feature sets require additional objective
function evaluations.

The \emph{cell-mapping} (\emph{CM}) approach proposed by~\cite{kerschke2014cell}
discretizes the input space to a user-defined
number of blocks (\ie cells) per dimension.
Afterwards, the corresponding
features are based on the relations between the cells and points within.
Six cell-mapping feature sets were defined:
\emph{CM Angle},
\emph{CM Convexity},
\emph{CM Gradient Homogeneity},
\emph{Generalized CM},
\emph{Barrier} and \emph{SOO Trees} by \cite{flamm2002barrier} and \cite{derbel2019new}.
It should be noted that \emph{CM} approach is less useful in higher
dimensions where the majority of cells is empty and feature computation
can require a lot of time and memory.

\emph{Nearest better clustering} (\emph{NBC}) features proposed by~\cite{kerschke2015detecting} are based on the detection of funnel structures.
The calculation of such features relies on the comparison of distances from observations to their nearest
neighbors and their \emph{nearest better neighbors},
which are the nearest neighbors among the set of all observations with a better
objective value.
\cite{lunacek2006dispersion} proposed the set of
\emph{dispersion features} comparing the dispersion among the data points and
among subsets of these points from the sample set.
The \emph{information content} features of a continuous landscape by~\cite{munoz2015exploratory}
are derived
from methods for
calculating the information content of discrete landscapes.
In~\cite{kerschke2017comprehensive}, other three feature sets were proposed:
\emph{Basic} set with features such as the number of points, search space boundaries or dimension,
the proportion of \emph{Principal Components} for a given percentage of variance,
coefficients of \emph{Linear Model} fitted in each cell.
A comprehensive survey of landscape analysis can be found,
\eg in~\cite{munoz2015algorithm}.
A more detailed description of feature sets used in our research
can be found in the online available \suppref{Supplementary material}{4}\footnote{\url{\suppurl}}.

\section{Landscape Analysis of Surrogate Models}
\label{sec:lasm}

In recent years, several surrogate-model approaches have been developed
to increase the performance of the CMA-ES
(cf. the overview in \cite{pitra2017overview}).
Each such approach has two complementary aspects:
the employed regression model and
the so called \emph{evolution control}~\citep{jin2001managing}
managing when to evaluate the model and when the true
objective function
while replacing the fitness evaluation step
on line 4 in Algorithm \ref{alg:wacmaes}.
In~\cite{pitra2021interaction}, we have shown a significant influence of the surrogate model on the CMA-ES performance.
Thus, we are interested in relationships between
the prediction error of surrogate models using various
features of the fitness function landscape.

First, we state the problem and research question connected with the relations between surrogate models and landscape features.
Then, we present our set of new landscape features based on the CMA-ES state variables.
Afterwards, we investigate the properties of landscape features in the context of the training set selection methods and select the most convenient features for further research.
Finally, we analyse relationships between the selected features and measured errors of the surrogate models with various settings.

\subsection{Problem Statement}
\label{ssec:problem}

The problem can be formalized as follows:
In a generation $\gen$ of a surrogate-assisted version of the CMA-ES,
a set of surrogate models $\set{M}$ with hyperparameters $\btheta$ are trained
utilizing particular choices of the training set $\trainset$.
The training set $\trainset$ is selected out of an \emph{archive} $\archive$ ($\trainset \subset \archive$)
containing all points in which the fitness has been evaluated so far,
using some TSS method (see Section~\ref{sssec:tss}).
Afterwards, $\set{M}$ is tested on the set of points sampled using the CMA-ES distribution
$\testset = \big\{\xx_k \big\vert \xx_k \sim \dis{N}\big(\gn{\mm}, {\gn{\sigma}}^2\gn{\CC}\big), k = 1, \ldots, \alpha\big\}$,
where $\alpha \in \NN$ depends on the evolution control.
The research question connected to this problem is:
What relationships between the suitability of different models
for predicting the fitness and the considered landscape features
do the testing results indicate?


\subsection{CMA-ES Landscape Features}
\label{ssec:cmafeat}

To utilize additional information comprised in CMA-ES state variables,
we have proposed a set of features based on
the CMA-ES \citep{pitra2019landscape}.

In each CMA-ES generation $\gen$ during the fitness evaluation step
(Step~\ref{ln:eval}),
the following additional features $\feat{}$ are obtained
for the set of points $\XX = \left\{\xx_i\right\}_{i=1}^\npt$:

\begin{itemize}
  \item \textbf{Generation number} $\artFeat{generation}{}{}{\ftCMA} = \gen$ indicates the phase of the optimization process.
  \item \textbf{Step-size} $\artFeat{step\_size}{}{}{\ftCMA} = \gn{\sigma}$ provides an information about the extent of the approximated region.
  \item \textbf{Number of restarts} $\artFeat{restart}{}{}{\ftCMA} = \gn{n_\text{r}}$
        performed till generation $\gen$ may indicate landscape difficulty.
  \item \textbf{Mahalanobis mean distance}
        of the CMA-ES mean $\gn{\mm}$ to the sample mean $\mu_\XX$ of $\XX$
        \begin{equation}
          \artFeat{mean\_dist}{\XX}{}{\ftCMA} = \sqrt{(\gn{\mm} - \mu_\XX)^\top \CC_\XX^{-1} (\gn{\mm} - \mu_\XX)} \,,
        \end{equation}
        where $\CC_\XX$ is the sample covariance of $\XX$.
        This feature indicates suitability of $\XX$ for model training
        from the point of view of the current state of the CMA-ES algorithm.
  \item Square of the $\vmat{p}_c$ \textbf{evolution path} length $\artFeat{evopath\_c\_norm}{}{}{\ftCMA} = \bignorm{\gn{\vmat{p}_c}}^2$
        is the only possible non-zero eigenvalue of \emph{rank-one update} covariance matrix $\gnp{\vmat{p}_c}\gnp{\vmat{p}_c}^\top$
        (see Subsection \ref{sssec:cmaes}).
        That feature providing information about the correlations between consecutive CMA-ES steps indicates a similarity of function landscapes among subsequent generations.

  \item \textbf{$\vmat{p}_\sigma$ evolution path} ratio,
        \ie the ratio between the evolution path length $\bignorm{\gn{\vmat{p}_\sigma}}$
        and the expected length of a random
        evolution path used to update step-size.
        It provides a useful information about distribution changes:
        \begin{equation}
          \artFeat{evopath\_s\_norm}{}{}{\ftCMA} = \frac{\bignorm{\gn{\vmat{p}_\sigma}}}{\ev\norm{\dis{N}(\vmat{0}, \vmat{I})}} =
            \frac{\bignorm{\gn{\vmat{p}_\sigma}}\,\Gamma\!\left(\frac{\dm}{2}\right)}{\sqrt{2}\,\Gamma\!\left(\frac{\dm+1}{2}\right)}  \,.
        \end{equation}
  \item \textbf{CMA similarity likelihood}.
        The log-likelihood of the set of points $\XX$ with respect to the CMA-ES distribution
        may also serve as a measure of its suitability for training
        \begin{equation}
          \artFeat{cma\_lik}{\XX\!}{}{\ftCMA} = - \frac{\npt}{2}\!\left(\!\dm \natlog 2\pi {\gn{\sigma}}^2\!\!+ \natlog \det{ \gn{\CC} }\!\right)
                        -\frac{1}{2}\!\sum_{\xx \in \XX}\!\left(\!\frac{\xx - \gn{\mm}}{\gn{\sigma}}\!\!\right)^{\!\!\!\!\top}
                        \!\!\gn{\CC}^{-1}\!\!\left(\!\frac{\xx - \gn{\mm}}{\gn{\sigma}}\!\!\right)\!.
          \label{eq:feat_cma_lik}
        \end{equation}
\end{itemize}


\subsection{Landscape Features Investigation}
\label{ssec:featprop}

Based on the studies in \cite{bajer2019gaussian} and \cite{pitra2021interaction},
we have selected the Doubly trained surrogate CMA-ES (DTS-CMA-ES) \citep{pitra2016doubly}
as a successful representative of surrogate-assisted CMA-ES algorithms.


During the model training procedure in the DTS-CMA-ES,
shown in general in Algorithm \ref{alg:trainmodel},
a set of training points $\trainset$ is selected
and transformations of $\XX_\text{tr}$ using matrix $\frac{1}{\sigma}\CC^{-1/2}$
and $\yy_\text{tr}$ to zero mean and unit variance
($\trainset = (\XX_\text{tr}, \yy_\text{tr})$) are calculated in Steps~\ref{ln:sCbase} and \ref{ln:normalize}
before the model hyperparameters $\btheta$ are fitted.
In case of a successful fitting procedure,
the resulting model is tested for constancy
on an extra generated population
due to the CMA-ES restraints against stagnation.
Using the same transformation, the points in $\testset$
are transformed before obtaining the
model prediction of fitness values $\hat{\yy}_\text{te}$,
which is then inversely transformed to the original output space.

\begin{algorithm}[t]
  \begin{algorithmic}[1]
    \REQUIRE{
      $\archive$ (archive),
      TSS (training set selection) method,
      $\model$ (surrogate model),
      $\modelsettings$ (model settings),
      $\lambda$ (population size),
      $\gn{\mm}$, $\gn{\sigma}$, $\gn{\CC}$ (CMA-ES state variables)}
    \STATE{$(\XX_\text{tr}, \yy_\text{tr}) \leftarrow$ select points from $\archive$ using TSS method}
    \STATE{$\XX_\text{tr} \leftarrow$ transform $\XX_\text{tr}$ using matrix $\frac{1}{(\gn{\sigma})^2} {\gn{\CC}}^{-1/2}$} 
    \label{ln:sCbase}
    \STATE{$\yy_\text{tr} \leftarrow$ normalize $\yy_\text{tr}$ to zero mean and unit variance}
    \label{ln:normalize}
    \STATE{$\btheta \leftarrow$ fit the hyperparameters of $\model$ using $\modelsettings$, $\XX_\text{tr}$, $\yy_\text{tr}$}
    \label{ln:fit}
    \STATE{$\xx_k^\text{test} \leftarrow$ sample $\gn{\mm} + \gn{\sigma} \dis{N}(\vmat{0}, \gn{\CC})$, $k = 1, \ldots, \lambda$ (create testing population)}
    \STATE{$y_k^\text{test} \leftarrow \model_{\btheta}(\xx_k^\text{test})$, $k = 1, \ldots, \lambda$ (evaluate testing population using model with hyperparameters $\btheta$)}
    \label{ln:predict}
    \IF{$\max_k(y_k^\text{test})-\min_k(y_k^\text{test}) < \min(10^{-8}, 0.05(\max(\yy_\text{tr})-\min(\yy_\text{tr}))$}
      \STATE{$\model_{\btheta}$ is considered constant $\Rightarrow$ $\model_{\btheta}$ is marked as \emph{not trained}}
      \label{ln:constant}
    \ENDIF{}
    \ENSURE{$\model_{\btheta}$ (surrogate model with hyperparameters $\btheta$)}
  \end{algorithmic}
  \caption{Generalized model training in DTS-CMA-ES~\citep{pitra2016doubly}}
  \label{alg:trainmodel}
\end{algorithm}

\subsubsection{Investigation Settings}
\label{sssec:featpropset}

To investigate landscape features in the context of surrogate-assisted optimization,
we have generated\footnote{
  Source codes covering all mentioned datasets generation and experiments are available on
  \url{\sourceurl}.
}
a dataset $\set{D}$ of sample sets
using independent runs of the 8 model settings from \cite{pitra2019landscape} for the DTS-CMA-ES algorithm \citep{bajer2019gaussian,pitra2016doubly}
on the 24 noiseless single-objective benchmark functions from the COCO framework~\citep{hansen2016coco}.
All runs were performed in dimensions 2, 3, 5, 10, and 20 on instances 11--15.
These instances all stem from the same base problem,
and are obtained by translation and/or rotation in the input space
and also translation of the objective values.
The algorithm runs were terminated
if the target fitness value $10^{-8}$ was reached
or
the budget of $250$ function evaluations per dimension was depleted.
Taking into account the double model training in DTS-CMA-ES,
we have extracted archives $\archive$ and testing sets $\testset$
only from the first model training.
The DTS-CMA-ES was employed in the overall best non-adaptive settings from~\cite{bajer2019gaussian}.
To obtain 100 comparable archives and testing sets for landscape features investigation,
we have generated a new collection of sample sets $\set{D}_{100}$,
where points for new archives and new populations are created using the weighted sum of original archive distributions from $\set{D}$.
The $\gen$-th generated dataset uses
the weight vector $\gnd{\vmat{w}}{\gen} = \frac{1}{9}(0, \ldots, \underset{\gen-3}{0}, \underset{\gen-2}{1},
                       \underset{\gen-1}{2}, \underset{\gen}{3}, \underset{\gen+1}{2}, 
                       \underset{\gen+2}{1}, \underset{\gen+3}{0}, \ldots, 0)^\top$,
which provides distribution smoothing across
the available generations\footnote{
The weighted sum of the original archive distributions satisfies
$\sum_{n=0}^{\gen_\text{max}} \gn{w_n} \dis{N}\big(\gnd{\mm}{n}, \gnd{\CC}{n}\big)
=
\dis{N}\big(\sum_{n=0}^{\gen_\text{max}} \gn{w_n} \gnd{\mm}{n}, \sum_{n=0}^{\gen_\text{max}} (\gn{w_n})^2 \gnd{\CC}{n} \big)
$,
where $\gen_\text{max}$ is the maximal generation reached by the considered original archive
and $\gnd{\mm}{n}$ and $\gnd{\CC}{n}$ are the mean and covariance matrix in CMA-ES generation $n$.
}.
The data from well-known benchmarks were also used by~\cite{saini2019automatic} and \cite{renau2020exploratory}.

Because each TSS method in Subsection~\ref{sssec:tss}
results in a different training set $\trainset$ using identical $\archive$,
we have performed all the feature investigations for each TSS method separately.
By combining
the two \emph{basic sample sets} for feature calculation $\archive$ and $\trainset$
with a population $\predset$ consisting of the points without a known value of the original fitness to be evaluated
by the surrogate model,
we have obtained two new sets $\archivepred = \archive \cup \predset$ and $\trainpredset = \trainset \cup \predset$.
Step~\ref{ln:sCbase} of Algorithm~\ref{alg:trainmodel}
performing the transformation of the input could also influence
the landscape features.
Thus, we have utilized either transformed and non-transformed sets
for feature calculations, resulting in 8 different sample sets
(4 in case of \tss{full} due to $\trainset = \archive$):
$\archive$, $\trans{\archive}$, $\archivepred$, $\trans{\archivepred}$,
$\trainset$, $\trans{\trainset}$, $\trainpredset$, $\trans{\trainpredset}$,
where $\trans{{}}$ denotes the transformation.

From each generated run in $\set{D}_{100}$, we have uniformly selected 100 generations.
In each such generations, we have computed all features
from the following feature sets for all 3 TSS methods
on all sample sets:
\[
  \begin{array}{llll}
    \supptextref{\emph{Basic}}{5}                      & \featBasic, \qquad\quad {} &
    \supptextref{\emph{Levelset}}{9}                   & \featLevel, \\
    \supptextref{\emph{CMA features}}{6}               & \featCMA, &
    \supptextref{\emph{Meta-Model}}{10}                & \featMM, \\
    \supptextref{\emph{Dispersion}}{7}                 & \featDisp, &
    \supptextref{\emph{Nearest Better Clustering}}{11} & \featNBC, \\
    \supptextref{\emph{Information Content}}{8}        & \featInfo, &
    \supptextref{\emph{y-Distribution}}{12}            & \featyDis.
  \end{array}
\]

These sets do not require additional evaluations of the objective function
and can be computed even in higher dimensions.
Some features were not computed due to the following reasons
(see Section \suppref{2}{4} in \suppmat{Supplementary material} for feature definitions):
From $\featBasic$,
we have used only
$\artFeat{dim}{}{}{}$ and $\artFeat{obs}{}{}{}$
because the remaining ones were constant on the whole $\set{D}_{100}$ dataset.
$\artFeat{dim}{}{}{}$ is identical regardless the sample set.
$\artFeat{obs}{}{}{}$ and all features from $\featyDis$
were not computed using the transformation matrix
because it does not influence their resulting values.
The feature $\artFeat{lin\_simple\_intercept}{}{}{\ftMM} \in \featMM$ was excluded
because it is useless if fitness normalization is performed
(see Step~\ref{ln:normalize} in Algorithm~\ref{alg:trainmodel}).
The remaining features from $\featMM$,
all the features from $\featNBC$,
and all three features from $\featyDis$
were not computed on sample sets
with $\predset$
because it also does not influence their resulting values.

For the rest of the paper, we will consider features
which are independent of the sample set
(\ie $\artFeat{dim}{}{}{}$ and 5 $\featCMA$ features)
as a part of $\archive$-based features only.
This results in the total numbers of landscape features
equal to 197 for \tss{full} (all were $\archive$-based) and 388 for \tss{nearest} and \tss{knn}
(from which 191 features were $\trainset$-based).


\subsubsection{Feature Analysis Process and Its Results}
\label{sssec:featpropres}

The results of experiments concerning landscape features
are presented in Tables~\suppref{S1}{14}--\suppref{S6}{19}
in \suppmat{Supplementary material}.
In the paper, we report only main highlights of the results in
Tables~\ref{table:featProp} and~\ref{tab:cluster}.

First, we have investigated the impossibility of feature calculation
(\ie the feature value $\nanout$)
for each feature.
Such information can be valuable and we will consider it as a valid output of any feature.
On the other hand, large amount of $\nanout$ values on the tested dataset suggests low usability of the respective feature.
Therefore, we have excluded features yielding $\nanout$ in more than 25\% of all measured values,
which were
for all TSS methods the feature
  $\artFeat{eps\_s}{}{}{\ftInfo}$ on
  $\archivepred$, $\trans{\archivepred}$, $\trainpredset$, and $\trans{\trainpredset}$
and for the \tss{knn} also the features
  $\artFeat{ratio\_mean\_02}{}{}{\ftDisp}$,
  $\artFeat{ratio\_median\_02}{}{}{\ftDisp}$,
  $\artFeat{diff\_mean\_02}{}{}{\ftDisp}$,
  $\artFeat{diff\_median\_02}{}{}{\ftDisp}$
on $\trainset$, $\trans{\trainset}$, $\trainpredset$, and $\trans{\trainpredset}$,
as well as $\artFeat{quad\_w\_interact\_adj\_r2}{}{}{\ftMM}$
on $\trainset$ and $\trans{\trainset}$.
This decreased the numbers of features to 195
for \tss{full},
384 for \tss{nearest}, and 366 for \tss{knn}.
Many features are difficult to calculate using low numbers of points.
Therefore, for each feature we have measured the minimal number of points $\npt_\nanout$ in a particular combination of feature and sample set,
for which the calculation resulted in $\nanout$ in at most 1\% of cases.
All measured values can be found in Tables~\suppref{S1}{14}--\suppref{S6}{19} in \suppmat{Supplementary material}
and values of selected robust features (see below) in Table~\ref{tab:cluster}.
For the calculation of most of the features, the CMA-ES default population size value
in $2\dm$: $\npt_\nanout = \lambda_\text{def} = 4 + \lfloor 3 \natlog 2 \rfloor = 6$,
or initial point plus doubled default population size in $2\dm$:
$\npt_\nanout = 1 + 2\lambda_\text{def} = 13$
was sufficient
in all sample sets indexed with $\predset$.
Considering sample-set-independent features,
no points are needed,
because the values concern
the CMA-ES iteration or CMA-ES run as a whole, not particular points.
As can be seen from Tables~\suppref{S1}{14}--\suppref{S6}{19} in \suppmat{Supplementary material},
the quantile of function values used for splitting the sample set decreases
as the dispersion features $\featDisp$ require more points
to be computable ($\feat{} \neq \nanout$).
For quantile values, see Section~\suppref{2.3}{7} in \suppmat{Supplementary material}.
Finally,
$\artFeat{lin\_w\_interact\_adj\_r2}{}{}{\ftMM}$
and $\artFeat{quad\_w\_interact\_adj\_r2}{}{}{\ftMM}$
have also high values of $\npt_\nanout$,
which is plausible taking into account
their descriptions (see Section~\suppref{2.6}{10} in \suppmat{Supplementary material}).

To increase comparability of investigated features,
we normalize all the features to interval $[0, 1]$
using sigmoid function
\begin{equation}
  \feat{\text{norm}} = \frac{1}{1+e^{-k(\feat{}-\feat{0})}}\,,\quad
  k = \frac{2 \natlog 99}{Q_{0.99}-Q_{0.01}}\,,\quad
  \feat{0} = \frac{Q_{0.01} + Q_{0.99}}{2} \,,
\end{equation}
where $Q_{0.01}$ and $Q_{0.99}$ are $0.01$ and $0.99$ quantiles of feature $\feat{}$
on the whole $\set{D}_{100}$ dataset considering values $\feat{}\in\RR\cup\{\pm\infty\}$.
The normalization is derived to map feature quantiles to $0.01$ and $0.99$,
\ie $\feat{\text{norm}}(Q_{0.01}) = 0.01$ and $\feat{\text{norm}}(Q_{0.99}) = 0.99$.
Such normalization maps infinity values to $0$ and $1$
and increases comparability of features with large differences in possible values
(\eg $\artFeat{step\_size}{}{}{\ftCMA} \in [1.5\cdot 10^{-10}, 1.4\cdot 10^{15}]$,
whereas $\artFeat{nb\_cor}{}{}{\ftNBC} \in [-1,1]$).

We have tested the dependency of individual features on the dimension
using feature medians from 100 samples for each distribution from $\set{D}_{100}$.
The Friedman's test rejected the hypothesis
that the feature medians are independent of the dimension
for all features at the family-wise significance level 0.05 using the Bonferroni-Holm correction.
Moreover, for most of the features,
the subsequently performed pairwise tests rejected the hypothesis of equality of feature medians
for all pairs od dimensions.
There were only several features for which the hypothesis was not rejected for some pairs of dimensions
(see Tables~\suppref{S1}{14}--\suppref{S6}{19} in \suppmat{Supplementary material}).
Therefore, the influence of the dimension on the vast majority of features is essential.

\input{tex/relTable}

Our analysis of the influence of multiple landscape features
on the predictive error of surrogate models
requires high robustness of features against random
sampling of points.
To have a robust set of $k$ independent features,
which return identical values for input in 95\% cases,
we would like all features to be identical in $100\sqrt[k]{0.95}\%$ cases.
Thus, for our dataset $\set{D}_{100}$
with 100 samples for each distribution
even a small $k$
requires all values to be identical,
which is almost impossible to achieve for most of the investigated features.
Therefore, we define feature \emph{robustness}
as a proportion of cases
for which the difference between the 1st and 100th percentile
calculated after standardization on samples
from the same CMA-ES distribution
is $\leq 0.05$.
Table~\ref{table:featProp} lists numbers of features achieving different levels of robustness.
We have selected the robustness $\geq 0.9$ to be used for subsequent analyses.
The robustness calculated for individual features is listed in
Tables~\suppref{S1}{14}--\suppref{S6}{19} in \suppmat{Supplementary material}
and its values for features with robustness $\geq 0.9$ in Table~\ref{tab:cluster}.
The chosen level of robustness excluded from further computations
all features from $\featNBC$ and $\featyDis$ for every TSS
and all features from $\featInfo$ for \tss{full} and \tss{nearest}.
Both $\featNBC$ and $\featyDis$ are rather sensitive
to the input data,
where the influence of non-uniform sampling of data is probably
not negligible.
Features from $\featInfo$ have very varied robustness.
Whereas $\artFeat{eps\_max}{}{}{\ftInfo}$ provides high robustness
around 0.85 on transformed sample sets (up to 0.97 for \tss{knn}),
computations of $\artFeat{m0}{}{}{\ftInfo}$ and $\artFeat{eps\_s}{}{}{\ftInfo}$ resulted in the robustness
$0.004$ and $0.016$, respectively,
which were the two lowest among all features.
The majority of $\featCMA$ features provided
high robustness
caused mainly due to the independence of most of the features on the sample set.
Features from $\featLevel$ based on quadratic discriminant analysis (\texttt{qda}) showed nearly double robustness compared to the rest of $\featLevel$ features.
The transformation of the input increased robustness
of specific types of features from $\featDisp$ and $\featMM$.
In particular, it increased the robustness of
difference-based features from $\featDisp$ to more than 0.99
and also of features based on model coefficients from $\featMM$.
\tss{knn} is more sample dependent, therefore,
the number of points in a sample set can vary.
This also decreases the robustness of some ratio-based features.
On the other hand,
coefficients of simple models from $\featMM$ show robustness over 0.99
and $\artFeat{cma\_mean\_dist}{}{}{\ftCMA}$ mostly over 0.9.
A noticeable dependence of robustness on which of the sets $\archive$, $\trainset$, or $\predset$ is used was not observed.

\newlength{\corrdimtabw}                  
\setlength{\corrdimtabw}{0.36\textwidth}

\begin{table}
  \caption{
    Results of a medoid clustering into 14 clusters of the features for individual TSS methods,
    accoding to their Schweizer-Wolf measure $\sigma$.
    The clusters are separated by horizontal lines and their medoid
    representatives are marked as gray lines.
    $N_\nanout$ denotes
    the minimal number of points
    for which the feature calculation resulted in $\nanout$ in at most 1\% of cases.
    The third column shows the robustness
    --- the proportion of measured cases such that the feature values of samples from the same CMA-ES distribution
    did not differ more than~0.05.
  }
  \label{tab:cluster}

  \begin{center}
    TSS full
  \end{center}

  \begin{center}
    \hfill
    \resizebox{\corrdimtabw}{!}{%
      \input{tex/corr_dim_full_1}
    }
    \hfill
    \resizebox{\corrdimtabw}{!}{%
      \input{tex/corr_dim_full_2}
    }
  \hfill
  \!
  \end{center}

  \begin{center}
    TSS nearest (1/2)
  \end{center}

  \begin{center}
    \hfill
    \resizebox{\corrdimtabw}{!}{%
      \input{tex/corr_dim_nearest_1}
    }
    \hfill
    \resizebox{\corrdimtabw}{!}{%
      \input{tex/corr_dim_nearest_2}
    }
    \hfill
    \!
  \end{center}
\end{table}

\setcounter{table}{1}

\begin{table}
  \caption{
    \hspace{-1ex}\textbf{(continued)}\hspace{1ex}
  }
  \begin{center}
    TSS nearest (2/2)
  \end{center}

  \begin{center}
    \hfill
    \resizebox{\corrdimtabw}{!}{%
      \input{tex/corr_dim_nearest_3}
    }
    \hfill
    \resizebox{\corrdimtabw}{!}{%
      \input{tex/corr_dim_nearest_4}
    }
    \hfill
    \!
  \end{center}

  \begin{center}
    TSS knn
  \end{center}

  \begin{center}
    \hfill
    \resizebox{\corrdimtabw}{!}{%
      \input{tex/corr_dim_knn_1}
    }
    \hfill
    \resizebox{\corrdimtabw}{!}{%
      \input{tex/corr_dim_knn_2}
    }
    \hfill
    \!
  \end{center}
\end{table}

%

The large number of features suggests that
for the purpose of investigation of their relationships to surrogate models,
they should be clustered into a smaller number of groups of similar features.
To this end, we have performed agglomerative hierarchical clustering according to $1 - \sigma_{\text{SW}}(\feat{i}, \feat{j})$,
where $\sigma_{\text{SW}}$ is the \emph{measure $\sigma$} by~\cite{schweizer1981nonparametric}
and $\feat{i}$, $\feat{j}$ are the vectors of all medians from $\set{D}_{100}$ for the features $i$ and $j$.
To compensate for the ordering-dependency of agglomerative hierarchical clustering,
we have performed 5 runs of clustering for each TSS method to find the optimal number of clusters.
The number of clusters exceeding a threshold 0.9 for $\sigma_{\text{SW}}$,
averaged over all 15 runs, was 14,
that number was subsequently used as the value of $k$ for subsequent $k$-medoid clustering
using again Schweizer-Wolff measure $\sigma$ as a similarity.
The features that are medoids of those 14 clusters are listed in Table~\ref{tab:cluster}.
Even such a small number of feature representatives can be sufficient
for achieving excellent performance in a subsequent investigation~\citep{hoos2018portfolio, renau2021towards}.
A majority of clusters contain features from the same feature set.
Sometimes, the whole cluster is composed of the same features calculated only on different sample sets
which suggests that the influence of feature calculation
on various sample sets might be negligible in those clusters.
On the other hand,
feature clusters for \tss{knn} often all share the base sample set
($\archive$ or $\trainset$), or the same transformation
even if the features are not from the same set of features.
Considering the large numbers of available features,
it is worth noticing that most of medoids are
identical or at least very similar for all TSS methods.
\tss{full} and \tss{nearest} medoids share identical features,
where only 4
($\artFeat{obs}{}{}{}$, $\artFeat{diff\_median\_02}{}{}{\ftDisp}$,
$\artFeat{lda\_qda\_25}{}{}{\ftLevel}$, $\artFeat{quad\_simple\_cond}{}{}{\ftMM}$)
differ in the sample set
($\archive$ vs. $\trainset$, $\trans{\archive}$ vs. $\trans{\trainset}$, $\archivepred$ vs. $\trainpredset$, and $\trans{\archive}$ vs. $\trans{\trainset}$ respectively).
Notice that sample sets differ only in using $\archive$ or $\trainset$.
Moreover, 10 out of 14 representatives
($\artFeat{dim}{}{}{}$,
 $\artFeat{obs}{}{}{}$,
 $\artFeat{evopath\_s\_norm}{}{}{\ftCMA}$,
 $\artFeat{restart}{}{}{\ftCMA}$,
 $\artFeat{cma\_lik}{}{}{\ftCMA}$,
 $\artFeat{diff\_median\_02}{}{}{\ftDisp}$,
 $\artFeat{diff\_mean\_05}{}{}{\ftDisp}$,
 $\artFeat{lda\_qda\_10}{}{}{\ftLevel}$,
 $\artFeat{lda\_qda\_25}{}{}{\ftLevel}$,
 $\artFeat{quad\_simple\_cond}{}{}{\ftMM}$)
are the same for all considered TSS methods,
whereas sample sets utilized for feature calculation sometimes differ.
Such similarity can indicate great importance of those features
for characterizing the fitness landscape
in the CMA-ES surrogate modeling context.

\subsection{Relationships of Landscape Features and Surrogate Models}
\label{ssec:connect}

To investigate the relationships between surrogate model errors
and 14 landscape features selected for each TSS in the previous subsection,
we have utilized 4 surrogate models: lq, lmm, GPs, and RFs.
The latter two models in 8 and 9 different settings respectively.
In~\citep{saini2019automatic}, $\featMM$ features were utilized to investigate connection with several
surrogate models in default settings
in static scenario only,
where the model is selected once for a specific problem regardless the possible following optimization process.

\subsubsection{Settings}
\label{sssec:connectset}

We have split the dataset\,$\set{D}$ into validation and testing parts
($\set{D}_\text{val}$\,and\,$\set{D}_\text{test}$)
on the covariance function level uniformly at random
(considering following levels of $\set{D}$:
dimension, function, instance, covariance function).
More specifically,
for each dimension, function, and instance in $\set{D}$,
we have uniformly selected runs using
7 covariance functions to $\set{D}_\text{test}$ and
1 covariance to $\set{D}_\text{val}$.
In each of those runs,
data from 25 uniformly selected generations were used.

The two error measures utilized in our research each follow different aspects of model precision:
The \emph{mean-squared error} ($\mse$) measures how much the model differs directly from the objective function landscape.
On the other hand,
the \emph{ranking difference error} ($\rde$)~\citep{bajer2019gaussian}
reflects that the CMA-ES state variables are adjusted according to the
ordering of $\mu$ best points from the current population
due to the invariance of the CMA-ES
with respect to monotonous transformations.
The $\rde$ of
$\yy\in\RR^\lambda$ with respect to
$\yy'\in\RR^\lambda $ considering $\mu$ best components is
defined:
\begin{gather}
  \rde_{\mu} (\yy, \yy') =
    \frac{\sum_{i,(\rho (\yy'))_i\le \mu} |(\rho (\yy'))_i-(\rho (\yy))_i|}{
    \max_{\pi} \sum_{i=1}^\mu|i-\pi^{-1} (i)| }\,,
\end{gather}
where $(\rho(\yy))_i$ is the rank of $y_i$ among the components of $\yy$.

The regression model from lmm-CMA
was used in its improved version published by~\cite{auger2013benchmarking}.
The lmm model operates with the transformation matrix in its own way,
thus, the transformation step during training
(step~\ref{ln:sCbase} in Algorithm~\ref{alg:trainmodel})
is not performed for this model.

The linear-quadratic model was used
in the version published by~\cite{hansen2019global}.
The original version utilizes all data
without a transformation, therefore,
the input data for lq model are not transformed for \tss{full}.

The GP regression model was employed in the version published by~\cite{bajer2019gaussian}
using 8~different covariance functions 
described in Table~\ref{tab:gpcov}.

\input{tex/gpcovTable}

The RF model was utilized in 9 different settings described in detail in Table~\ref{tab:rfSet}.

\input{tex/rfSetTable}

\subsubsection{Results}
\label{sssec:connectres}

\begin{figure*}
  \begin{center}
    \begin{tikzpicture}
      \node[anchor=south west,inner sep=0] at (0,0) {\includegraphics[clip, width=\textwidth, trim=595 595 646 646]{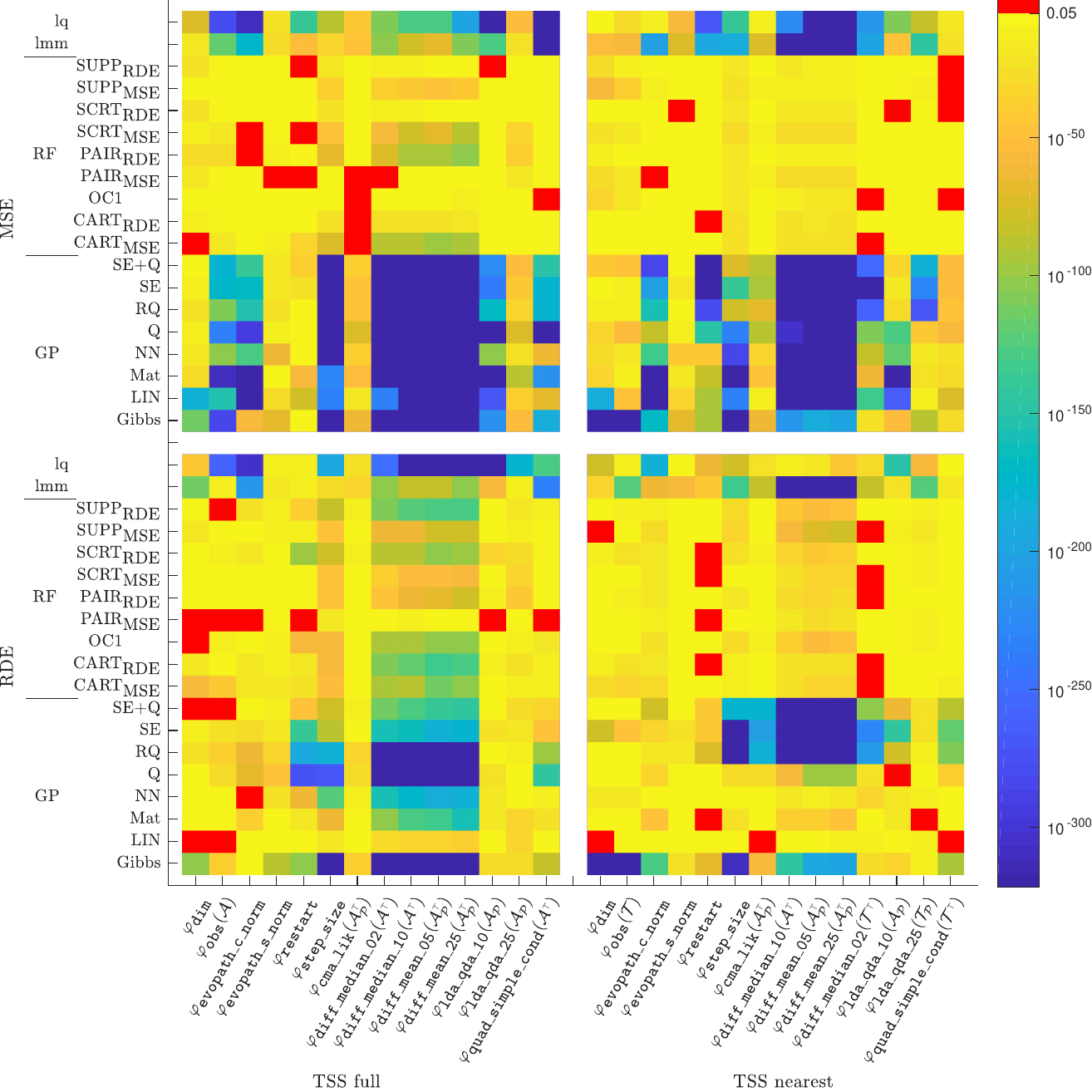}};
      \draw[white,fill=white] (1.8,7.89) rectangle ++(0.5, 0.14);
      \draw[white,fill=white] (6.97,2.3) rectangle ++(0.14,0.5);
    \end{tikzpicture}
  \end{center}

  \hspace{3.7cm}
  \includegraphics[clip, width=0.24\textwidth, trim=200 250 265 269 ]{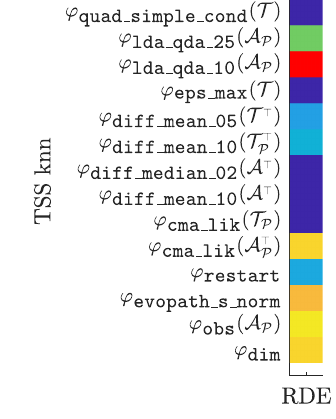}
  \includegraphics[clip, width=0.0388\textwidth, trim=335 250 265 269 ]{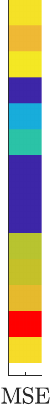}

  \caption{The visualisation of the p-values of the Kolmogorov-Smirnov test
           comparing the equality of probability distributions of individual features on all data
           and on those data on which a particular model setting scored best.
           Non-red colored squares denote KS test results rejecting the equality of distributions with the Holm correction
           at the family-wise significance level $\alpha=0.05$,
           otherwise, p-values are visualised as red squares.
           \tss{knn} was employed only in connection with lmm model.
  }
  \label{fig:ksFig}
\end{figure*}

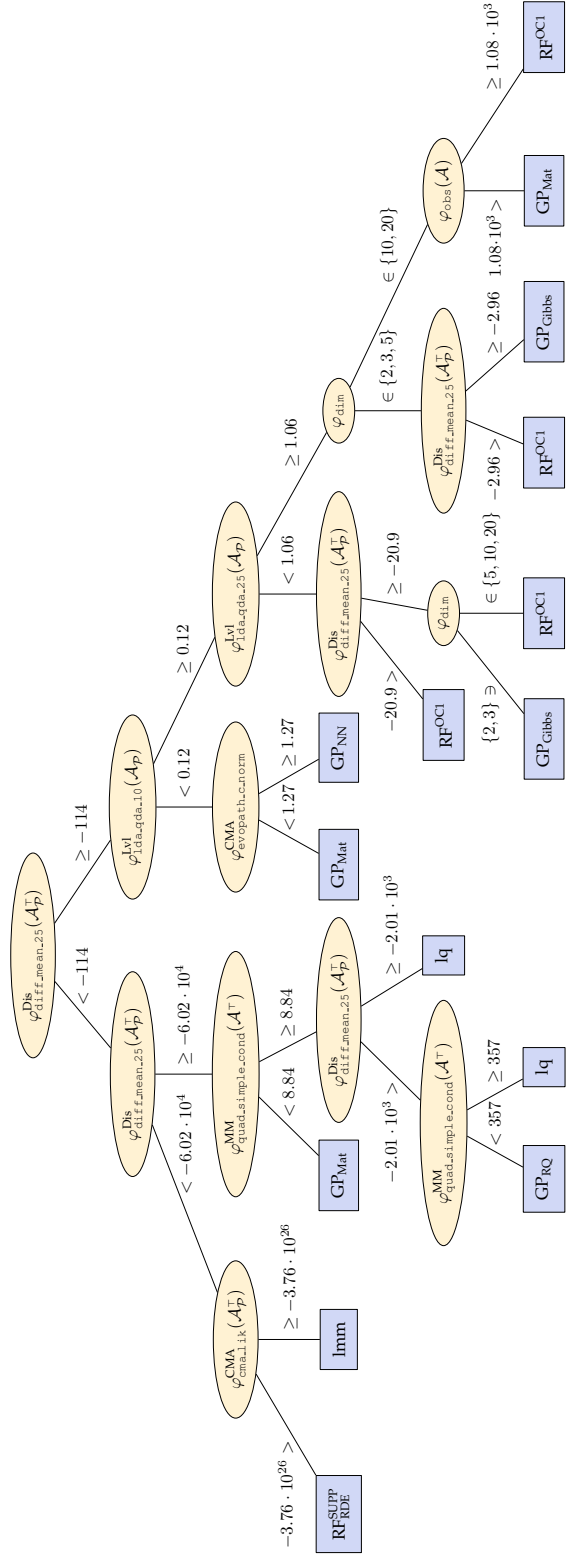
\begin{sidewaysfigure*}
 \resizebox{\textwidth}{!}{%
    \input{tex/tree_full_postproc.tex}
 }
  \caption{
    Classification tree analysing the influence of landscape features on the most suitable model and its setting using \tss{full}.
  }
  \label{fig:dec_tree_full}
\end{sidewaysfigure*}

\begin{figure*}
 \centering
 \resizebox{0.89\textwidth}{!}{%
    \input{tex/tree_nearest_postproc.tex}
 }
  \caption{
    Classification tree analysing the influence of landscape features on the most suitable model and its setting using \tss{nearest}.
  }
  \label{fig:dec_tree_nearest}
\end{figure*}
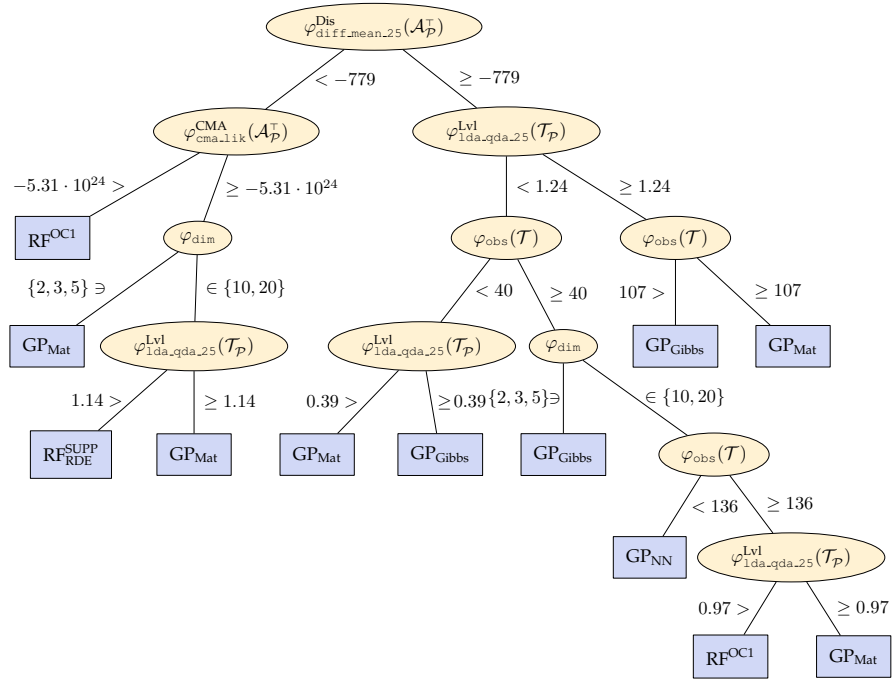

The results of analysing relationships between landscape features
and two error measures for selected surrogate models with the appropriate settings are presented
in Figures~\ref{fig:ksFig}--\ref{fig:dec_tree_nearest},
Tables~\ref{tab:modelNanTable} and~\ref{tab:duelTable_rde} and
Tables~\suppref{S7}{21}--\suppref{S13}{27} in \suppmat{Supplementary material}.

\input{tex/defFile}
\input{tex/modelNanTable}

We have summed up the cases when the model did not provide the prediction,
\ie the error value is not available,
in Table~\ref{tab:modelNanTable}.
Such cases can occur when hyperparameters fitting fails, fitness prediction fails, or the model is considered constant
(Steps~\ref{ln:fit}, \ref{ln:predict}, and \ref{ln:constant} in Algorithm~\ref{alg:trainmodel} respectively).
The numbers more or less confirm 
that more complex methods are more likely to fail.
Whereas the lq model, the most simple model among all tested,
provided predictions in almost 98\% of all cases,
where the missing results can be attributed to constant predictions,
the GP model with Gibbs covariance function provided only
55\% of the required predictions.
Generally, the models were able to provide predictions
more often when using \tss{full} than when using \tss{nearest},
probably due to the locality of the training set,
considering that it is easier to train a constant model
on a smaller area,
where the differences between objective values are more likely
negligible.
The \tss{knn} was the most successful selection method for lmm,
probably because it was designed directly for this model.
In the following investigation,
the cases where the error values were missing for a particular model,
were excluded from all comparisons involving that model.

\input{tex/duelTable_rde}

We have tested the statistical significance of
the $\mse$ and $\rde$
differences for 19 surrogate model settings
using \tss{full} and \tss{nearest} methods
and also the lmm surrogate model utilizing \tss{knn},
\ie 39 different combinations of model settings $\modelsettings$ and TSS methods $(\modelsettings, \text{TSS})$,
on all available sample sets
using the Friedman test.
The resulting p-values of the two Friedman tests, one for each error measure, are below the smallest double precision number.
A pairwise comparison of the model settings with respect to $\mse$ and $\rde$
revealed significant differences among the vast majority of pairs of model settings
according to the non-parametric two-sided Wilcoxon signed rank test with the Holm correction for the family-wise error.
To better illustrate the differences between individual settings,
we also count the percentage of cases at which one setting
had the error lower than the other.
The pairwise score and the statistical significance
of the pairwise
differences are summarized in Table~\ref{tab:duelTable_rde} for $\rde$
and Tables~\suppref{S7}{21} and \suppref{S8}{22} in \suppmat{Supplementary material} for both error measures.
Results of statistical tests confirmed that the obtained $\mse$ and $\rde$ values
are sufficiently diverse for further investigation of model settings suitability.
The best overall results were provided by GPs with all covariances except LIN.
Especially, GP using SE+Q as a covariance function in \tss{nearest} significantly outperformed all other
$(\modelsettings, \text{TSS})$ combinations.
The polynomial models using \tss{nearest} take the second place in such comparison, being outperformed only by the GP models mentioned above.
Generally, models using \tss{nearest} provided better results than when using \tss{full}
(in case of lmm also better than \tss{knn}).
The percentages of $\rde$ show smaller differences in precision than $\mse$ due to the larger probability of error equality on the limited number of possible $\rde$ values
compared to the infinite number of possible $\mse$ values.

To compare the suitability of individual features as descriptors, we separated areas where
the surrogate model $\model$ with a particular $(\modelsettings, \text{TSS})$ combination
has the best performance.
Subseqently, we compared the distributions of each feature value on the whole $\set{D}_\text{test}$
with a distribution of each feature value of the selected part using Kolmogorov-Smirnov test (KS test)
for MSE and RDE separately.
The hypothesis of distribution equality is tested
at the family-wise significance level $\alpha=0.05$ after the Holm correction.
The resulting p-values summarized in Tables~\suppref{S9}{23}--\suppref{S13}{27} in \suppmat{Supplementary material}
and visualised in color in Figure~\ref{fig:ksFig}
show significant differences
in distribution
among combinations of model settings
and TSS method $(\modelsettings, \text{TSS})$
for the vast majority of considered features.
Features from $\featDisp$ and $\artFeat{step\_size}{}{}{\ftCMA}$ provided the most significant differences for almost all models.
For the GP models and the sample set leading to the lowest $\mse$, the p-values were often even below the smallest double precision number.
These features also provided very low p-values for lmm and lq model.
The only exception is $\artFeat{diff\_median\_02}{\trainset}{\transcma}{\ftDisp}$ providing notably higher p-values for all model settings.
Moreover, features calculated on $\trainset$-based
sample sets for \tss{nearest} provided in most cases
higher p-values
than when calculated on $\archive$-based.
From the model point of view, differences of RF model settings are much lower than the rest of settings for all the tested features.
This might suggest the lower ability to distinguish between the individual RF settings maybe due to the low performance of these settings on the dataset.
The p-values for the $\mse$ and the $\rde$ also differ notably
mainly due to the different ranges of values of these two error measures.
As its consequence, p-values for $\mse$ more clearly suggest
the distinctive power of individual features on model precision,
regardless the fact that $\rde$ is more convenient for CMA-ES related problems.
Overall, the results of the KS test have shown that there is no tested landscape feature representative
for which the selection of the covariance function would be negligible for the vast majority of model settings.

To perform a multivariate statistical analysis we have built two classification trees (CTs): one for \tss{full} and one for \tss{nearest}.
The data for each TSS method described by 14 features
selected in Subsection~\ref{ssec:featprop}
were divided into 19 classes according to which of the 19 considered surrogate model settings 
achieved the lowest $\rde$.
In case of a tie,
the model with the lowest $\mse$ among models with the lowest $\rde$ was the chosen one.
The CT trained on those data is the MATLAB implementation of the CART \citep{breiman1984classification},
where the minimal number of cases in leaf was set to 5000,
the twoing rule was used as a splitting criterion,
$\artFeat{dim}{}{}{}$ was considered as a discrete variable and the remaining 13 features were considered as continuous.
The resulting CTs are depicted in Figures~\ref{fig:dec_tree_full} and \ref{fig:dec_tree_nearest}.



The most successful kind of models are GPs, present in most of the leaves proving the leading role shown in pairwise comparisons
of model setting errors.
Gibbs and $\mathrm{Mat}^{\sfrac{5}{2}}$ covariances of the GP models constitute the winning GP settings regardless the fact
that both provided the lowest numbers of predictions (see Table~\ref{tab:modelNanTable}).
This is probably caused by the removal of cases where
predictions of any model setting were missing.
The winning model setting of RF is OC1 method being the most often selected leaf of the CT for \tss{full} method.
The polynomial models are not present in the resulting CT for \tss{nearest}.
The feature $\artFeat{diff\_mean\_25}{\archivepred}{\transcma}{\ftDisp}$ plays possibly the most important role in the CT for \tss{full},
being in the root node as well as in 4 other nodes, and also very important in the \tss{nearest} CT, where it is in the root node.
This confirms the very strong decisive role of $\featDisp$ features indicated in the results of KS test.
Basic features $\artFeat{dim}{}{}{}$ and $\artFeat{obs}{}{}{}$ in few nodes provide the decisions resulting in GP model
if both feature values are small and in RF model otherwise.
Such dependency coincides with
the better ability of RF models to preserve the prediction quality with the growing dimension than the ability of GP models.
Features from the $\featLevel$ are also very important at least in connection with the \tss{nearest}.
The $\featCMA$ features are present only in very few nodes.
We can observe the connection between the feature $\artFeat{quad\_simple\_cond}{\archive}{\transcma}{\ftMM}$
and the lq model in the CT for \tss{full}
showing the possible ability of $\featMM$ features
to detect convenience of polynomial model usage.

\section{Conclusion}
\label{sec:concl}

This paper has presented a study of relationships between the prediction error of the surrogate models, their settings,
and features describing the fitness landscape
during evolutionary optimization by the CMA-ES
state-of-the-art optimizer for expensive continuous black-box tasks.
Four models in 39 different settings for the DTS-CMA-ES,
a surrogate-assisted version of the CMA-ES,
were compared using three sets of 14 landscape features
selected according to their properties, especially according to their robustness and similarity
to other features.
We analysed $\mse$ and $\rde$ dependence
of various models and model settings on the features
calculated using three methods selecting sample sets extracted from
DTS-CMA-ES runs on noiseless benchmarks from the COCO framework.
Up to our knowledge,
this analysis of landscape feature properties and their connection 
to surrogate models in evolutionary optimization
is more profound than any other published so far.

Most of the landscape features were designed to be calculated
on some initial sample set from the more or less uniform
distribution covering the whole search space to easily
identify the optimized function and/or the algorithm
with the highest performance on such function.
In this paper, we operate on a more local
level using only data from the actual runs of the optimization
algorithm generated in each generation using Gaussian distribution,
\ie data completely different from the original design of those features.
Therefore, the low number of robust and mostly similar features
confirms the findings of \cite{renau2020exploratory} that the distribution of the initial design has a notable
impact on landscape analysis.

The statistical analysis has shown significant differences
in the error values among all 39 model settings using different TSS methods for both $\mse$ and $\rde$.
The GP model settings provided the highest perfomance followed by polynomial models.
Most of the investigated features had distributions on sample sets with particular model settings
achieving the lowest error values significantly different from the overall distribution of all data for both error measures and all 3 TSS methods.
Finally, the overall results have shown the dispersion features
as highly influential on the model settings error values,
followed by
features based on CMA state variables,
features describing
the similarity of fitness function to some simple model,
features splitting space according to the black-box function values,
and simple features such as dimension and number of observations.

The results of this research could be utilized to design
a metalearning-based system~\citep{kerschke2019automated, saini2019automatic}
for selection of surrogate models
in surrogate-assisted evolutionary optimization algorithms context.
The landscape features investigated in this paper are in metalearning known as metafeatures.

\small

\bibliographystyle{apalike}
\bibliography{pitra2022ec}

\end{document}


\ecjHeader{x}{x}{xxx-xxx}{201X}{Landscape Analysis for Surrogate Models in the Evol. B-B Context (Supp. Mat.)}{Z. Pitra, J. Koza, J. Tumpach, M. Hole\v{n}a}
\title{\bf Landscape Analysis for Surrogate Models in the Evolutionary Black-Box Context (Supplementary Material)}

\author{
        \name{\bf Zbyn\v{e}k Pitra} \hfill \addr{z.pitra@gmail.com}\\ 
        \addr{Faculty of Nuclear Sciences and Physical Engineering, Czech Technical University\\
          B\v{r}ehov\'{a} 7, 115\,19 Prague, Czech Republic
        }
\AND
       \name{\bf Jan Koza} \hfill \addr{koza@cs.cas.cz}\\
       \addr{Faculty of Information Technology, CTU in Prague, \\
         Th\'{a}kurova 9, 160\,00 Prague 6, Czech Republic
       }
\AND
       \name{\bf Ji\v{r}\'{i} Tumpach} \hfill \addr{tumpach@cs.cas.cz}\\
       \addr{Faculty of Mathematics and Physics, Charles University in Prague,\\
         Malostran. n\'{a}m. 25, 118\,00 Prague, Czech Republic
       }
\AND
       \name{\bf Martin Hole\v{n}a} \hfill \addr{martin@cs.cas.cz}\\
       \addr{Institute of Computer Science, Czech Academy of Sciences,\\
         Pod Vod\'{a}renskou v\v{e}\v{z}\'{i} 2, 182\,07 Prague, Czech Republic
       }
}

\maketitle

\begin{abstract}

Surrogate modeling has become a valuable
technique for black-box optimization tasks
with expensive evaluation of the objective function.
In this supplementary material for the article
\emph{Landscape Analysis for Surrogate Models in the Evolutionary Black-Box Context} \citep{pitra2022landscape},
we report additional results from the
investigation of the relationship between the predictive accuracy of surrogate models
and features of the black-box function landscape.
We also report additional results from the study of properties of features for landscape analysis
in the context of different transformations and ways of selecting the input data.

\end{abstract}

\begin{keywords}

Black-box optimization,
Surrogate modeling,
Landscape analysis,
Metalearning CMA-ES

\end{keywords}

\section{Introduction}

\begin{algorithm} [t]
  \begin{algorithmic}[1]

    \REQUIRE{
      $\lambda$ (population-size),
      $\BB$ (fitness function),
      $(w_i)_{i=1}^\mu$, $\mu_w$ (selection and recombination parameters), 
      $c_\sigma$, $d_\sigma$ (step-size control parameters), 
      $c_c$, $c_I$, $c_\mu$ 
      (covariance matrix adaptation parameters)
    }
    \STATE{
      \textbf{init}
      $\gen = 0,
       \gnd{\sigma}{0} > 0,
       \gnd{\vmat{m}}{0} \in \RR^\dm,
       \gnd{\CC}{0} = \vmat{I},
       \gnd{\vmat{p}_\sigma}{0} = \vmat{0},
       \gnd{\vmat{p}_c}{0} = \vmat{0},
       \mu = \left\lfloor\lambda/2\right\rfloor
      $
    }
    \REPEAT
      \STATE{$\xx_k \leftarrow \gn{\mm} + \gn{\sigma} \dis{N}(\vmat{0}, \gn{\CC})$ \qquad $k = 1, \ldots, \lambda$}
      \label{ln:offspring}
      \STATE{$\BB_k \leftarrow \BB(\xx_k)$ \hspace{2.03cm} $k = 1, \ldots, \lambda$}
      \label{ln:eval}
      \STATE{$\gnp{\mm} \leftarrow \sum_{i=1}^\mu w_i \xx_{i:\lambda}$, where $\xx_{i:\lambda}$ is the $i$-th best individual out of $\xx_1, \ldots, \xx_\lambda$}
      \label{ln:mean}
      \STATE{$\gnp{\vmat{p}_\sigma} \leftarrow (1-c_\sigma)\gn{\vmat{p}_\sigma} +
              \sqrt{c_\sigma (2-c_\sigma) \mu_w}{\gn{\CC}}^{-1/2}\frac{\gnp{\mm}-\gn{\mm}}{\gn{\sigma}}$}
      \label{ln:pathsigma}
      \STATE{$h_\sigma \leftarrow \mathbb{I}\left(\|\gnp{\vmat{p}_\sigma}\| < \sqrt{1-(1-c_\sigma)^{2(\gen + 1)}}
              \left(1.4 + \frac{2}{\dm+1}\right) \ev\|\dis{N}(0,1))\|\right)$}
      \STATE{$\gnp{\vmat{p}_c} \leftarrow (1-c_c)\gn{\vmat{p}_c} + 
              h_\sigma \sqrt{c_c (2-c_c) \mu_w} \, \frac{\gnp{\mm}-\gn{\mm}}{\gn{\sigma}}$}
      \label{ln:pathC}
      \STATE{
        $\CC_\mu \leftarrow \sum_{i=1}^\mu w_i \gn{\sigma}^{-2}
        \big(\gnp{\xx_{i:\lambda}}-\gn{\mm}\big)
        \big(\gnp{\xx_{i:\lambda}}-\gn{\mm}\big)
        ^\top$
      }
      \label{ln:cmu}
      \STATE{$\gnp{\CC} \leftarrow (1-c_1-c_\mu) \gn{\CC} + c_1 \gnp{\vmat{p}_c}\gnp{\vmat{p}_c}^\top + c_\mu \CC_\mu$} 
      \label{ln:updateC}
      \STATE{$
        \gnp{\sigma} \leftarrow \gn{\sigma} 
        \exp\Big(\frac{c_\sigma}{d_\sigma}\Big(\frac{\|\gnp{\vmat{p}_\sigma}\|}{\ev\|\dis{N}(\vmat{0}, \vmat{I})\|} -
1 \Big)\Big)
        $}
      \label{ln:updatesigma}
      \STATE{$\gen \leftarrow \gen + 1$}
    \UNTIL{stopping criterion is met}
    \ENSURE{$\xx_\text{res}$ (resulting optimum)}

  \end{algorithmic}
  \caption{Pseudocode of the CMA-ES}
  \label{alg:wacmaes}
\end{algorithm}
\label{sec:intro}

In~\citep{pitra2022landscape} we have presented an investigation of relationships
between the prediction error of the surrogate models, their settings,
and features describing the fitness landscape
during evolutionary optimization by the Covariance Matrix Adaptation Evolution Strategy (CMA-ES)
state-of-the-art optimizer for expensive continuous black-box tasks
\citep{hansen1996adapting, hansen2006cma}.
Its pseudocode
is outlined in Algorithm~\ref{alg:wacmaes}. 
Four models in 39 different settings for the Doubly Trained Surrogate CMA-ES (DTS-CMA-ES) by \cite{pitra2016doubly},
a surrogate-assisted version of the CMA-ES,
were compared using three sets of 14 landscape features
selected according to their properties, especially according to their robustness and similarity
to other features.
We analysed \emph{mean squared error} ($\mse$) and \emph{ranking difference error} ($\rde$)
\citep{bajer2019gaussian}
dependence
of various models and model settings on the features
calculated using three methods selecting sample sets extracted from
DTS-CMA-ES runs on noiseless benchmarks
from the Comparing Continuous Optimisers (COCO) testbed~\citep{hansen2016coco}.
Up to our knowledge,
this analysis of landscape feature properties and their connection 
to surrogate models in evolutionary optimization
is more profound than any other published so far.

The employed extracted \emph{basic sample sets} were:
\begin{itemize}
  \item \emph{archive} $\archive$ containing all points evaluated so far by the fitness;
  \item \emph{training set} $\trainset$ selected from $\archive$ used for model training.
\end{itemize}
By combining
the two basic sample sets
with a population $\predset$ consisting of the points without a known value of the fitness to be evaluated
by the surrogate model,
we have obtained two new sets $\archivepred = \archive \cup \predset$ and $\trainpredset = \trainset \cup \predset$.

The employed training set selection (TSS) methods were:
\begin{itemize}

  \item \hypertarget{tss:full}{\emph{TSS full}}, taking all the already evaluated points, \ie $\trainset=\archive$,

  \item \hypertarget{tss:knn}{\emph{TSS knn}}, selecting the union of the sets of $k$-nearest neighbors of all points for which the fitness should be predicted, where $k$ is
  given (\eg in \cite{kern2006local}),

  \item \hypertarget{tss:nearest}{\emph{TSS nearest}}, selecting the union of the sets of $k$-nearest neighbors of all points for which the fitness should be predicted,
    where $k$ is maximal such that the total number of selected points does not exceed a given maximum number of points $N_\text{max}$ and
    no point is further from current mean $\gn{\mm}$ than a given maximal distance $r_\text{max}$ (\eg in \cite{bajer2019gaussian}).
\end{itemize}
We have performed all the feature investigations for each TSS method separately.

We have utilized sets either transformed or non-transformed into the $\sigma^2 \CC$ basis
for feature calculations, resulting in 8 different sample sets
(4 in case of \tss{full} due to $\trainset = \archive$):
$\archive$, $\trans{\archive}$, $\archivepred$, $\trans{\archivepred}$,
$\trainset$, $\trans{\trainset}$, $\trainpredset$, $\trans{\trainpredset}$,
where $\trans{{}}$ denotes the transformation.

The supplementary material is structured as follows.
Section~\ref{sec:featdef} describes landscape features utilized in~\citep{pitra2022landscape} in more detail.
Sections~\ref{sec:featres} and \ref{sec:connect} complement
the main paper results of testing and analysis, respectively,
with more detailed results.

\clearpage

\section{Feature Definitions}
\label{sec:featdef}

Let us consider an input set $\XX$ of $\npt$ points in $\dm$-dimensional input space
\begin{equation*}
  \XX = \big\{\xx_i \big|\,\xx_i \in \bigtimes_{j=1}^\dm [l_j, u_j], l_j, u_j \in \RR, l_j< u_j\big\}_{i=1}^N\,.  
\end{equation*}
Then we can define a \emph{sample set} $\sS$ of $\npt$ pairs of observations
\begin{equation*}
  \sS = \left\{(\xx_i, y_i) \, \vert \, \xx_i \in \XX, y_i \in \RR\cup\{\nanin\}, i = 1, \ldots, \npt \right\}\,,
\end{equation*}
where $\nanin$ denotes missing $y_i$ value (\eg $\xx_i$ was not evaluated yet).
Then the sample set can be utilized to describe landscape properties using a 
\emph{landscape feature} $\feat{}$
\begin{equation*}
  \feat{}: \bigcup_{\npt\in\NN} \RR^{\npt, \dm} \times \left(\RR\cup\{\nanin\}\right)^{\npt, 1} \mapsto \RR \cup \{\pm\infty, \nanout\}\,,
\end{equation*}
where $\nanout$ denotes impossibility of feature computation.
The $y_i$ values can be gathered to a set $\yy = \{y_i\}_{i=1}^\npt$.

The following features were reimplemented in Matlab according to the R-package
\texttt{flacco} by~\cite{kerschke2017flacco}.
In the words of \cite{kerschke2017automated},
\emph{
  ``It is important to notice that, independent of the research domain, most of these features do not provide intuitively understandable numbers.
  Therefore, we strongly recommend not to interpret them on their own.
  Moreover, some of them are stochastic and hence should be evaluated multiple times on an instance and afterwards be aggregated in a reasonable manner.
  Nevertheless, they definitely provide information that can be of great importance to scientific models, such as machine learning algorithms in general or algorithm selectors in particular.''
}
Following the remark about stochasticity,
we have fixed random seeds which are not fixed in the original implementation
to always return the same values for identical input.

\newpage

\subsection{Basic Features $\featBasic$}
\label{apps:\ftBasic}

\cite{kerschke2017comprehensive} summarized the following features
providing obvious information about the input space:

\begin{itemize}
  \item \textbf{Dimension} of the input space $\artFeat{dim}{}{}{} = \dm$.
  \item \textbf{Number of observations} $\artFeat{obs}{}{}{} = \npt$.
  \item \textbf{Minimum} and \textbf{maximum of lower bounds}
    $\artFeat{lower\_min}{}{}{} = \min_{i\in\dm} l_i$,
    $\artFeat{lower\_max}{}{}{} = \max_{i\in\dm} l_i$.
  \item \textbf{Minimum} and \textbf{maximum of upper bounds}
    $\artFeat{upper\_min}{}{}{} = \min_{i\in\dm} u_i$,
    $\artFeat{upper\_max}{}{}{} = \max_{i\in\dm} u_i$.
  \item \textbf{Minimum} and \textbf{maximum of $y$ values}
    $\artFeat{objective\_min}{}{}{} = \min_{i\in\npt} y_i$,
    $\artFeat{objective\_max}{}{}{} = \max_{i\in\npt} y_i$.
  \item \textbf{Minimum} and \textbf{maximum of cell blocks per dimension}
    $\artFeat{blocks\_min}{}{}{}$,
    $\artFeat{blocks\_max}{}{}{}$.
  \item \textbf{Total number of cells}
    $\artFeat{cells\_total}{}{}{}$.
  \item \textbf{Number of filled cells}
    $\artFeat{cells\_filled}{}{}{}$.
  \item Binary flag stating whether the \textbf{objective function}
                  should be \textbf{minimized}
    $\artFeat{minimize\_fun}{}{}{}$.
\end{itemize}

In this research, we have used only
$\artFeat{dim}{}{}{}$ and $\artFeat{obs}{}{}{}$
because the remaining ones were constant on the whole $\set{D}_{100}$ dataset.
$\artFeat{dim}{}{}{}$ is identical regardless the sample set and
$\artFeat{obs}{}{}{}$ was not computed in the $\sigma^2 \CC$ basis
because it does not influence the resulting feature values.

\newpage

\subsection{CMA Features $\featCMA$}
\label{apps:\ftCMA}

In each CMA-ES generation $\gen$ during the fitness evaluation step
(Algorithm~\ref{alg:wacmaes}, step~\ref{ln:eval}),
we have defined the following features in~\cite{pitra2019landscape}
for the set of points $\XX$:

\begin{itemize}
  \item \textbf{Generation number} $\artFeat{generation}{}{}{\ftCMA} = \gen$ indicates the phase of the optimization process.
  \item \textbf{Step-size} $\artFeat{step\_size}{}{}{\ftCMA} = \gn{\sigma}$ provides an information about the extent of the approximated region.
  \item \textbf{Number of restarts} $\artFeat{restart}{}{}{\ftCMA} = \gn{n_\text{r}}$
        performed till generation $\gen$ may indicate landscape difficulty.
  \item \textbf{Mahalanobis mean distance}
        of the CMA-ES mean $\gn{\mm}$ to the sample mean $\mu_\XX$ of $\XX$
        \begin{equation}
          \artFeat{mean\_dist}{\XX}{}{\ftCMA} = \sqrt{(\gn{\mm} - \mu_\XX)^\top \CC_\XX^{-1} (\gn{\mm} - \mu_\XX)} \,,
        \end{equation}
        where $\CC_\XX$ is the sample covariance of $\XX$.
        This feature indicates suitability of $\XX$ for model training
        from the point of view of the current state of the CMA-ES algorithm.
  \item Square of the $\vmat{p}_c$ \textbf{evolution path} length $\artFeat{evopath\_c\_norm}{}{}{\ftCMA} = \bignorm{\gn{\vmat{p}_c}}^2$
        is the only possible non-zero eigenvalue of \emph{rank-one update} covariance matrix $\gnp{\vmat{p}_c}\gnp{\vmat{p}_c}^\top$
        (see step~\ref{ln:updateC} in Algorithm~\ref{alg:wacmaes}).
        That feature providing information about the correlations between consecutive CMA-ES steps indicates a similarity of function landscapes among subsequent generations.

  \item \textbf{$\vmat{p}_\sigma$ evolution path} ratio,
        \ie the ratio between the evolution path length $\bignorm{\gn{\vmat{p}_\sigma}}$
        and the expected length of a random
        evolution path used to update step-size.
        It provides a useful information about distribution changes:
        \begin{equation}
          \artFeat{evopath\_s\_norm}{}{}{\ftCMA} = \frac{\bignorm{\gn{\vmat{p}_\sigma}}}{\ev\norm{\dis{N}(\vmat{0}, \vmat{I})}} =
            \frac{\bignorm{\gn{\vmat{p}_\sigma}}\,\Gamma\!\left(\frac{\dm}{2}\right)}{\sqrt{2}\,\Gamma\!\left(\frac{\dm+1}{2}\right)}  \,.
        \end{equation}
  \item \textbf{CMA similarity likelihood}.
        The log-likelihood of the set of points $\XX$ with respect to the CMA-ES distribution
        may also serve as a measure of its suitability for training
        \begin{align}
          \artFeat{cma\_lik}{\XX}{}{\ftCMA} = & - \frac{\npt}{2} \left(\dm \natlog 2\pi {\gn{\sigma}}^2 + \natlog \det{ \gn{\CC} } \right) \nonumber \\ 
                                              & - \frac{1}{2} \sum_{\xx \in \XX} \left(\frac{\xx - \gn{\mm}}{\gn{\sigma}}\right)^\top 
                                                                \gn{\CC}^{-1} \left(\frac{\xx - \gn{\mm}}{\gn{\sigma}}\right) \,.
        \end{align}
\end{itemize}

\newpage

\subsection{Dispersion Features $\featDisp$}
\label{apps:\ftDisp}

The dispersion features by \cite{lunacek2006dispersion} compare the dispersion among observations from $\sS$ and among a subset of these points. The subsets are created based on tresholds using the \texttt{quantile}s 0.02, 0.05, 0.1, and 0.25 of the objective values $\yy$.

The set of all distances within the set $\sS$:
\begin{equation}
  D_\text{all} = \{\operatorname{dist}(\xx_i, \xx_j)\,|\,i, j = 1, \ldots, \npt\}\,.
\end{equation}

The quantile subset of all distances:
\begin{equation}
  D_\texttt{quantile} = \{\operatorname{dist}(\xx_i, \xx_j)\,|\,y_i, y_j <= Q_\texttt{quantile}(\yy), \}\,.
\end{equation}

\begin{itemize}
  \item \textbf{Ratio of} the quantile subset and all points \textbf{median distances}
    \begin{equation}
      \artFeat{ratio\_median\_[quantile]}{\sS}{}{\ftDisp} = \frac{\operatorname{median} D_\texttt{quantile}}{\operatorname{median} D_\text{all}}\,.
    \end{equation}
  \item \textbf{Ratio of} the quantile subset and all points \textbf{mean distances}
    \begin{equation}
      \artFeat{ratio\_mean\_[quantile]}{\sS}{}{\ftDisp} =\frac{\operatorname{mean} D_\texttt{quantile}}{\operatorname{mean} D_\text{all}}\,.
    \end{equation}
  \item \textbf{Difference between} the quantile subset and all points \textbf{median distances}
    \begin{equation}
      \artFeat{diff\_median\_[quantile]}{\sS}{}{\ftDisp} = \operatorname{median} D_\texttt{quantile}-\operatorname{median} D_\text{all}\,.
    \end{equation}
  \item \textbf{Difference between} the quantile subset and all points \textbf{mean distances}
    \begin{equation}
      \artFeat{diff\_mean\_[quantile]}{\sS}{}{\ftDisp} = \operatorname{mean} D_\texttt{quantile}-\operatorname{mean} D_\text{all}\,.
    \end{equation}
\end{itemize}

\newpage

\subsection{Information Content Features $\featInfo$}
\label{apps:\ftInfo}

The Information Content of Fitness Sequences by~\cite{munoz2015exploratory} approach is based on a symbol sequence $\Psi = \{\psi_1, \ldots, \psi_{\npt-1}\}$, where
\begin{equation}
  \psi_i =
  \left\{
  \begin{array}{llcrr}
    \bar{1} \,, & \quad \text{if} &       \frac{y_{i+1} - y_i}{\norm{\xx_{i+1} - \xx_i}} & < & -\varepsilon \,, \\
    0 \,,       & \quad \text{if} & \left|\frac{y_{i+1} - y_i}{\norm{\xx_{i+1} - \xx_i}}\right| & \leq & \varepsilon \,, \\
    1 \,,       & \quad \text{if} &       \frac{y_{i+1} - y_i}{\norm{\xx_{i+1} - \xx_i}} & > & \varepsilon \,.
  \end{array}
  \right.
\end{equation}
The sequence considers observations from a sample set $\sS$
as a random walk accross the objective landscape
and depends on the information sensitivity parameter $\varepsilon > 0$.

The symbol sequence $\Psi$ is aggregated by the information content 
$H(\varepsilon) = - \sum_{i \neq j} p_{ij}\log_6 p_{ij}$,
where $p_{ij}$ is the probability of having the “block” $\psi_i \psi_j$,
where $\psi_i, \psi_j \in \{\bar{1}, 0, 1\}$,
within the sequence.
Note that the base of the logarithm was set to six 
as this equals the number of possible blocks $\psi_i \psi_j$ for which $\psi_i \neq \psi_j$, \ie $\psi_i \psi_j \in \{\bar{1}0, 0\bar{1}, \bar{1}1, 1\bar{1}, 01, 10\}$.

Another aggregation of the information is the so-called partial 
information content $M(\varepsilon) = |\Psi'|/(\npt - 1)$,
where $\Psi'$ is the symbol sequence
of alternating $\bar{1}$’s and $1$’s,
which is derived from $\Psi$ by removing all $0$
and repeated symbols.

When we do consider $\nanin$ as a valid state of $y$,
the sequence $\Psi$ consists of
$\psi_i \in \{\bar{1}, 0, 1, \bar{N}\}$,
where $\psi_i = \bar{N}$, if $y_{i+1} = \nanin$ or $y_i = \nanin$. Thus,
$H(\varepsilon) = - \sum_{i \neq j} p_{ij}\log_{12} p_{ij}$
due to the increased number of
possible $\psi_i \psi_j$ blocks, \ie $\psi_i \psi_j \in
\{
        \bar{1}0,       0\bar{1},
        \bar{1}1,       1\bar{1},
              01,             10,
  \bar{N}\bar{1}, \bar{1}\bar{N},
        \bar{N}0,       0\bar{N},
        \bar{N}1,       1\bar{N}
\}$.
Therefore, features based on $\Psi'$ can be utilized
only when $\nanin$ state is not present in $\yy$.

Based on sequences $\Psi$ and $\Psi'$ the following features can be defined according to~\cite{munoz2015exploratory}:
\begin{itemize}
  \item \textbf{Maximum information content}
    $\artFeat{h\_max}{\sS}{}{\ftInfo} = \max_\varepsilon H(\varepsilon)$.
  \item \textbf{Settling sensitivity}
    $\artFeat{eps\_s}{\sS}{}{\ftInfo} = \log_{10} \min (\varepsilon|H(\varepsilon) < s)$, where default $s = 0.05$ (see~\cite{munoz2015exploratory}).
  \item \textbf{Maximum sensitivity}
    $\artFeat{eps\_max}{\sS}{}{\ftInfo} = \arg\,\max_\varepsilon H(\varepsilon)$.
  \item \textbf{Initial partial information}
    $\artFeat{m0}{\sS}{}{\ftInfo} = M(\varepsilon = 0)$.
  \item \textbf{Ratio of partial information sensitivity}
    $\artFeat{eps\_ratio}{\sS}{}{\ftInfo} = \log_{10} \max (\varepsilon | M(\varepsilon) > r \artFeat{m0}{\sS}{}{\ftInfo}))$,
    where default $r = 0.5$ (see also~\cite{munoz2015exploratory}).
\end{itemize}

\newpage

\subsection{Levelset Features $\featLevel$}
\label{apps:\ftLevel}

In Levelset features by~\cite{mersmann2011exploratory},
the sample set $\sS$ is split into two classes by a specific treshold
calculated using the quantiles 0.1, 0.25, and 0.5.
Linear, quadratic, and mixture discriminant analysis
(\texttt{lda}, \texttt{qda}, and \texttt{mda}) are used to predict whether the objective values $\yy$ fall below 
or exceed the calculated threshold.
The extracted features are based on the distribution of the resulting cross-validated mean misclassification errors of each classifier.

\begin{itemize}
  \item \textbf{Mean misclassification error} of an appropriate discriminant analysis \texttt{method} using defined \texttt{quantile}
    $\artFeat{mmce\_[method]\_[quantile]}{\sS}{}{\ftLevel}$.
  \item \textbf{Ratio between} mean misclassification \textbf{errors} of two discriminant analysis \texttt{method}s using a given \texttt{quan\-tile}
    $\artFeat{[method1]\_[method2]\_[quantile]}{\sS}{}{\ftLevel}$.
\end{itemize}

\newpage

\subsection{Metamodel Features $\featMM$}
\label{apps:\ftMM}

To calculate metamodel features by~\cite{mersmann2011exploratory},
linear and quadratic regression models (\texttt{lin} and \texttt{quad}) with or without interactions are fitted to a sample set $\sS$.
The adjusted coefficient of determination $R^2$ and features reflecting the size relations of the model coefficients are extracted:

\begin{itemize}
  \item \textbf{Adjusted} $R^2$ of a \textbf{simple model}
    $\artFeat{[model]\_simple\_adj\_r2}{\sS}{}{\ftMM}$.
  \item \textbf{Adjusted} $R^2$ of a \textbf{model with interactions}
    $\artFeat{[model]\_w\_interact\_adj\_r2}{\sS}{}{\ftMM}$.
  \item \hypertarget{ft:MM:lsi}{\textbf{Intercept} of a simple \textbf{linear model}}
    $\artFeat{lin\_simple\_intercept}{\sS}{}{\ftMM}$.
  \item \textbf{Minimal} absolute \textbf{value of linear model coefficients}
    $\artFeat{lin\_simple\_coef\_min}{\sS}{}{\ftMM}$.
  \item \textbf{Maximal} absolute \textbf{value of linear model coefficients}
    $\artFeat{lin\_simple\_coef\_max}{\sS}{}{\ftMM}$.
  \item \textbf{Ratio} of \textbf{maximal} and \textbf{minimal} abs. \textbf{value of linear model coefficients}
    $\artFeat{lin\_simple\_coef\_max\_by\_min}{\sS}{}{\ftMM}$.
  \item \textbf{Ratio} of \textbf{maximal} and \textbf{minimal} abs. \textbf{value of quadratic model coefficients}
    $\artFeat{quad\_simple\_cond}{\sS}{}{\ftMM}$.
\end{itemize}

The feature $\artFeat{lin\_simple\_intercept}{}{}{\ftMM}$ was excluded
because it is useless if fitness normalization is performed.
The remaining features
were not computed on sample sets
with $\predset$
because it does not influence the resulting feature values.

\newpage

\subsection{Nearest Better Clustering Features $\featNBC$}
\label{apps:\ftNBC}

Nearest better clustering features by~\cite{kerschke2015detecting}
extract information based on the comparison of the sets of distances from all observations towards their nearest neighbors ($D_\text{nn}$)
and their nearest better neighbors ($D_\text{nb}$).

The distance to the nearest neighbor of a search point $\xx_i$, $i=1,\ldots, \npt$ from a 
set of points $\XX$:
\begin{equation}
  d_\text{nn}(\xx_i, \XX) = \min({\operatorname{dist}(\xx_i, \xx_j) | \xx_j \in \XX, i\neq j})\,.
\end{equation}

The distance to the nearest better neighbor:
\begin{equation}
  d_\text{nb}(\xx_i, \XX) = \min({\operatorname{dist}(\xx_i, \xx_j) | y_j < y_i, \xx_j \in \XX)})\,.
\end{equation}

The set of all nearest neighbor distances within the set $\XX$:
\begin{equation}
  D_\text{nn} = \{d_\text{nn}(\xx_i, \XX) | i = 1, \ldots, \npt\}\,.
\end{equation}

The set of all nearest better distances:
\begin{equation}
  D_\text{nb} = \{d_\text{nb}(\xx_i, \XX) | i = 1, \ldots, \npt\}\,.
\end{equation}

The features are defined as follows:
\begin{itemize}
  \item \textbf{Ratio of the standard deviations} between the $D_\text{nn}$ and $D_\text{nb}$
    $\artFeat{nb\_std\_ratio}{\sS}{}{\ftNBC} = \frac{\operatorname{std} D_\text{nn}}{\operatorname{std} D_\text{nb}}$.
  \item \textbf{Ratio of the means} between the $D_\text{nn}$ and $D_\text{nb}$
    $\artFeat{nb\_mean\_ratio}{\sS}{}{\ftNBC} = \frac{\operatorname{mean} D_\text{nn}}{\operatorname{mean} D_\text{nb}}$.
  \item \textbf{Correlation between the distances} of the nearest neighbors and nearest better neighbors
    $\artFeat{nb\_cor}{\sS}{}{\ftNBC} = \operatorname{corr}(D_\text{nn}, D_\text{nb})$.
  \item \textbf{Coefficient of} variation of the \textbf{distance ratios}
    $\artFeat{dist\_ratio}{\sS}{}{\ftNBC} = \frac{\operatorname{std} \set{W}}{\operatorname{mean} \set{W}}$,
    where $\set{W} = \left\{\frac{d_\text{nn}(\xx, \XX)}{d_\text{nb}(\xx, \XX)}\middle| \xx \in \XX\right\}$.
  \item \textbf{Correlation between the fitness value,
    and the count of observations} to whom the current observation
    is the nearest better neighbor
    $\artFeat{nb\_fitness\_cor}{\sS}{}{\ftNBC} =
    - \operatorname{corr}(\{\operatorname{deg}^{-}(\xx_i), y_i | i = 1, \ldots, \npt\})$,
    where $\operatorname{deg}^{-}(\xx_i)$ is the indegree
    of $\xx_i$ in the nearest better graph
    (the number of points for which a certain point is the nearest better point).
\end{itemize}

The features from $\featNBC$
were not computed on sample sets
with $\predset$
because it does not influence the resulting feature values.

\newpage

\subsection{$\yy$-Distribution Features $\featyDis$}
\label{apps:\ftyDis}

The $\yy$-distribution features from~\cite{mersmann2011exploratory} compute the basic statistics of the fitness values $\yy$.

\begin{itemize}
  \item \textbf{Skewness} and \textbf{kurtosis} of $\yy$ values
    $\artFeat{skewness}{\yy}{}{\ftyDis}$ and
    $\artFeat{kurtosis}{\yy}{}{\ftyDis}$.
  \item \textbf{Number of peaks} of the kernel-based estimation of the density of $\yy$-distribution
    $\artFeat{number\_of\_peaks}{\yy}{}{\ftyDis}$,
    where the peak is defined according to \cite{kerschke2017flacco}.
\end{itemize}

All three features
were computed only on $\archive$ and $\trainset$
because neither
$\predset$ nor the $\sigma^2 \CC$ basis
influence the resulting feature values.

\newpage

\section{Landscape Feature Investigation Results}
\label{sec:featres}


First, we have investigated the impossibility of feature calculation
(\ie the feature value $\nanout$)
for each feature.
Considering that large amount of $\nanout$ values on the tested dataset suggests low usability of the respective feature,
we have excluded features yielding $\nanout$ in more than 25\% of all measured values.
Many features are difficult to calculate using low numbers of points.
Therefore, for each feature we have measured the minimal number of points $\npt_\nanout$ in a particular combination of feature and sample set,
for which the calculation resulted in $\nanout$ in at most 1\% of cases.
All measured values can be found in Tables~\ref{tab:featProp_full_1}--\ref{tab:featProp_knn_2}.

We have tested the dependency of individual features on the dimension
using feature medians from 100 samples for each distribution from the dataset.
The Friedman's test rejected the hypothesis
that the feature medians are independent of the dimension
for all features at the family-wise significance level 0.05 using the Bonferroni-Holm correction.
Moreover, for most of the features,
the subsequently performed pairwise tests rejected the hypothesis of equality of feature medians
for all pairs od dimensions.
There were only several features for which the hypothesis was not rejected for some pairs of dimensions
(see Tables~\ref{tab:featProp_full_1}--\ref{tab:featProp_knn_2}).

Any analysis of the influence of multiple landscape features
on the predictive error of surrogate models
requires high robustness of features against random
sampling of points.
Therefore, we define feature \emph{robustness}
as a proportion of cases
for which the difference between the 1st and 100th percentile
calculated after standardization on samples
from the same CMA-ES distribution
is $\leq 0.05$.
The robustness calculated for individual features is listed in
Tables~\ref{tab:featProp_full_1}--\ref{tab:featProp_knn_2}.

\input{tex/feat_full_1}
\input{tex/feat_nearest_1}
\input{tex/feat_knn_1}

\newpage

\section{Connection of Landscape Features and Model Error Results}
\label{sec:connect}

We have tested the statistical significance of
the $\mse$ and $\rde$
differences for 19 surrogate model settings
using \tss{full} and \tss{nearest} methods
and also the lmm surrogate model utilizing \tss{knn},
\ie 39 different combinations of model settings $\modelsettings$ and TSS methods $(\modelsettings, \text{TSS})$,
on all available sample sets
using the non-parametric two-sided Wilcoxon signed rank test with the Holm correction for the family-wise error.
To better illustrate the differences between individual settings,
we also count the percentage of cases at which one setting
had the error lower than the other.
The pairwise score and the statistical significance
of the pairwise
differences are summarized in Tables~\ref{tab:duelTable_mse} and \ref{tab:duelTable_rde}.

To compare the convenience of individual features as descriptors of areas where
the surrogate model $\model$ with a particular $(\modelsettings, \text{TSS})$ combination
has the best performance,
we use the Kolmogorov-Smirnov test (KS test) testing the equality of the distribution of values of individual features calculated
on the whole testing dataset and on only those sample sets from the testing dataset
for which the $(\modelsettings, \text{TSS})$ combination leads to the lowest error
($\mse$ or $\rde$) among all tested combinations.
The hypothesis of distribution equality is tested
at the family-wise significance level $\alpha=0.05$ after the Holm correction.
The resulting p-values are summarized in Tables~\ref{tab:ksTable_mse_full}--\ref{tab:ksTable_mse_knn}.

\input{tex/defFile}

\input{tex/duelTable_mse}
\input{tex/duelTable_rde}

\input{tex/ksTable_mse_full}
\input{tex/ksTable_rde_full}
\input{tex/ksTable_mse_nearest}
\input{tex/ksTable_rde_nearest}
\clearpage
\input{tex/ksTable_mse_knn}

\clearpage

\small

\bibliographystyle{apalike}
\bibliography{pitra2022ec_supp}

%% file: tex/relTable.tex
\begin{table}
\centering
\caption{Proportion of features for individual TSS with robustness greater or equal to the threshold in the first column. The proportions in brackets represent $\trainset$-based features for given TSS (TSS full has only $\archive$-based). The numbers in bold in the grey row are utilized for the following process.}
\label{table:featProp}
\begin{tabular}{lrrr}
\toprule
threshold &         TSS full &                TSS nearest &                    TSS knn \\
\midrule
      0.5 & \textbf{125}/195 & \textbf{244}/384 (119/189) & \textbf{188}/366 ( 63/171) \\
      0.6 & \textbf{ 82}/195 & \textbf{158}/384 ( 76/189) & \textbf{131}/366 ( 49/171) \\
      0.7 & \textbf{ 54}/195 & \textbf{102}/384 ( 48/189) & \textbf{ 93}/366 ( 39/171) \\
      0.8 & \textbf{ 43}/195 & \textbf{ 80}/384 ( 37/189) & \textbf{ 73}/366 ( 30/171) \\
\rowcolor[gray]{0.8500}
      0.9 & \textbf{ 33}/195 & \textbf{ 60}/384 ( 27/189) & \textbf{ 59}/366 ( 26/171) \\
     0.99 & \textbf{ 28}/195 & \textbf{ 50}/384 ( 22/189) & \textbf{ 30}/366 (  2/171) \\
\bottomrule
\end{tabular}
\end{table}

%% file: tex/corr_dim_full_1.tex
\begin{tabular}{lrr}
\toprule
                                                                        & $N_\nanout$ & rob.(\%) \\
\midrule
\rowcolor[gray]{0.8500}
                                        $\tableFeat{obs}{\archive}{}{}$ &           1 &   100.00 \\
                            $\tableFeat{generation}{}{}{\ftCMA}$ &           0 &   100.00 \\
                                    $\tableFeat{obs}{\archivepred}{}{}$ &          13 &   100.00 \\
\midrule
\rowcolor[gray]{0.8500}
    $\tableFeat{quad\_simple\_cond}{\archive}{\transcma}{\ftMM}$ &           6 &    95.21 \\
\midrule
            $\tableFeat{cma\_lik}{\archive}{\transnone}{\ftCMA}$ &           1 &    99.75 \\
      $\tableFeat{diff\_mean\_10}{\archive}{\transcma}{\ftDisp}$ &          12 &    99.52 \\
      $\tableFeat{diff\_mean\_25}{\archive}{\transcma}{\ftDisp}$ &           6 &    99.55 \\
        $\tableFeat{cma\_lik}{\archivepred}{\transnone}{\ftCMA}$ &          13 &    99.75 \\
  $\tableFeat{diff\_mean\_10}{\archivepred}{\transcma}{\ftDisp}$ &          25 &    99.52 \\
\rowcolor[gray]{0.8500}
  $\tableFeat{diff\_mean\_25}{\archivepred}{\transcma}{\ftDisp}$ &          15 &    99.55 \\
\midrule
      $\tableFeat{diff\_mean\_05}{\archive}{\transcma}{\ftDisp}$ &          27 &    99.49 \\
    $\tableFeat{diff\_median\_05}{\archive}{\transcma}{\ftDisp}$ &          27 &    99.46 \\
\rowcolor[gray]{0.8500}
  $\tableFeat{diff\_mean\_05}{\archivepred}{\transcma}{\ftDisp}$ &          42 &    99.49 \\
$\tableFeat{diff\_median\_05}{\archivepred}{\transcma}{\ftDisp}$ &          42 &    99.48 \\
\midrule
\rowcolor[gray]{0.8500}
    $\tableFeat{diff\_median\_10}{\archive}{\transcma}{\ftDisp}$ &          12 &    99.48 \\
    $\tableFeat{diff\_median\_25}{\archive}{\transcma}{\ftDisp}$ &           6 &    99.49 \\
$\tableFeat{diff\_median\_10}{\archivepred}{\transcma}{\ftDisp}$ &          25 &    99.49 \\
$\tableFeat{diff\_median\_25}{\archivepred}{\transcma}{\ftDisp}$ &          15 &    99.50 \\
\bottomrule
\end{tabular}

%% file: tex/corr_dim_full_2.tex
\begin{tabular}{lrr}
\toprule
                                                                        & $N_\nanout$ & rob.(\%) \\
\midrule
      $\tableFeat{diff\_mean\_02}{\archive}{\transcma}{\ftDisp}$ &          71 &    99.52 \\
\rowcolor[gray]{0.8500}
    $\tableFeat{diff\_median\_02}{\archive}{\transcma}{\ftDisp}$ &          71 &    99.44 \\
  $\tableFeat{diff\_mean\_02}{\archivepred}{\transcma}{\ftDisp}$ &          86 &    99.52 \\
$\tableFeat{diff\_median\_02}{\archivepred}{\transcma}{\ftDisp}$ &          86 &    99.45 \\
\midrule
\rowcolor[gray]{0.8500}
                               $\tableFeat{restart}{}{}{\ftCMA}$ &           0 &   100.00 \\
\midrule
             $\tableFeat{cma\_lik}{\archive}{\transcma}{\ftCMA}$ &           1 &    99.96 \\
\rowcolor[gray]{0.8500}
         $\tableFeat{cma\_lik}{\archivepred}{\transcma}{\ftCMA}$ &          13 &    99.96 \\
\midrule
\rowcolor[gray]{0.8500}
                      $\tableFeat{evopath\_s\_norm}{}{}{\ftCMA}$ &           0 &   100.00 \\
\midrule
\rowcolor[gray]{0.8500}
                      $\tableFeat{evopath\_c\_norm}{}{}{\ftCMA}$ &           0 &   100.00 \\
\midrule
\rowcolor[gray]{0.8500}
  $\tableFeat{lda\_qda\_25}{\archivepred}{\transnone}{\ftLevel}$ &          15 &    94.13 \\
   $\tableFeat{lda\_qda\_25}{\archivepred}{\transcma}{\ftLevel}$ &          15 &    94.12 \\
\midrule
\rowcolor[gray]{0.8500}
                                                $\tableFeat{dim}{}{}{}$ &           0 &   100.00 \\
\midrule
\rowcolor[gray]{0.8500}
  $\tableFeat{lda\_qda\_10}{\archivepred}{\transnone}{\ftLevel}$ &          18 &    92.37 \\
   $\tableFeat{lda\_qda\_10}{\archivepred}{\transcma}{\ftLevel}$ &          18 &    92.38 \\
\midrule
\rowcolor[gray]{0.8500}
                            $\tableFeat{step\_size}{}{}{\ftCMA}$ &           0 &   100.00 \\
\bottomrule
\end{tabular}

%% file: tex/corr_dim_nearest_1.tex
\begin{tabular}{lrr}
\toprule
                                                                         & $N_\nanout$ & rob.(\%) \\
\midrule
\rowcolor[gray]{0.8500}
   $\tableFeat{lda\_qda\_10}{\archivepred}{\transnone}{\ftLevel}$ &          18 &    92.37 \\
    $\tableFeat{lda\_qda\_10}{\archivepred}{\transcma}{\ftLevel}$ &          18 &    92.38 \\
  $\tableFeat{lda\_qda\_10}{\trainpredset}{\transnone}{\ftLevel}$ &          18 &    92.37 \\
   $\tableFeat{lda\_qda\_10}{\trainpredset}{\transcma}{\ftLevel}$ &          18 &    92.38 \\
\midrule
       $\tableFeat{diff\_mean\_05}{\archive}{\transcma}{\ftDisp}$ &          27 &    99.49 \\
     $\tableFeat{diff\_median\_05}{\archive}{\transcma}{\ftDisp}$ &          27 &    99.46 \\
\rowcolor[gray]{0.8500}
   $\tableFeat{diff\_mean\_05}{\archivepred}{\transcma}{\ftDisp}$ &          42 &    99.49 \\
 $\tableFeat{diff\_median\_05}{\archivepred}{\transcma}{\ftDisp}$ &          42 &    99.48 \\
      $\tableFeat{diff\_mean\_05}{\trainset}{\transcma}{\ftDisp}$ &          27 &    99.49 \\
    $\tableFeat{diff\_median\_05}{\trainset}{\transcma}{\ftDisp}$ &          27 &    99.46 \\
  $\tableFeat{diff\_mean\_05}{\trainpredset}{\transcma}{\ftDisp}$ &          42 &    99.49 \\
$\tableFeat{diff\_median\_05}{\trainpredset}{\transcma}{\ftDisp}$ &          42 &    99.48 \\
\midrule
             $\tableFeat{cma\_lik}{\archive}{\transnone}{\ftCMA}$ &           1 &    99.75 \\
       $\tableFeat{diff\_mean\_10}{\archive}{\transcma}{\ftDisp}$ &          12 &    99.52 \\
       $\tableFeat{diff\_mean\_25}{\archive}{\transcma}{\ftDisp}$ &           6 &    99.55 \\
         $\tableFeat{cma\_lik}{\archivepred}{\transnone}{\ftCMA}$ &          13 &    99.75 \\
   $\tableFeat{diff\_mean\_10}{\archivepred}{\transcma}{\ftDisp}$ &          25 &    99.52 \\
\rowcolor[gray]{0.8500}
   $\tableFeat{diff\_mean\_25}{\archivepred}{\transcma}{\ftDisp}$ &          15 &    99.55 \\
            $\tableFeat{cma\_lik}{\trainset}{\transnone}{\ftCMA}$ &           1 &    99.75 \\
      $\tableFeat{diff\_mean\_10}{\trainset}{\transcma}{\ftDisp}$ &          12 &    99.52 \\
      $\tableFeat{diff\_mean\_25}{\trainset}{\transcma}{\ftDisp}$ &           6 &    99.55 \\
        $\tableFeat{cma\_lik}{\trainpredset}{\transnone}{\ftCMA}$ &          13 &    99.75 \\
  $\tableFeat{diff\_mean\_10}{\trainpredset}{\transcma}{\ftDisp}$ &          25 &    99.52 \\
  $\tableFeat{diff\_mean\_25}{\trainpredset}{\transcma}{\ftDisp}$ &          15 &    99.55 \\
\bottomrule
\end{tabular}

%% file: tex/corr_dim_nearest_2.tex
\begin{tabular}{lrr}
\toprule
                                                                         & $N_\nanout$ & rob.(\%) \\
\midrule
\rowcolor[gray]{0.8500}
     $\tableFeat{diff\_median\_10}{\archive}{\transcma}{\ftDisp}$ &          12 &    99.48 \\
     $\tableFeat{diff\_median\_25}{\archive}{\transcma}{\ftDisp}$ &           6 &    99.49 \\
 $\tableFeat{diff\_median\_10}{\archivepred}{\transcma}{\ftDisp}$ &          25 &    99.49 \\
 $\tableFeat{diff\_median\_25}{\archivepred}{\transcma}{\ftDisp}$ &          15 &    99.50 \\
    $\tableFeat{diff\_median\_10}{\trainset}{\transcma}{\ftDisp}$ &          12 &    99.48 \\
    $\tableFeat{diff\_median\_25}{\trainset}{\transcma}{\ftDisp}$ &           6 &    99.49 \\
$\tableFeat{diff\_median\_10}{\trainpredset}{\transcma}{\ftDisp}$ &          25 &    99.49 \\
$\tableFeat{diff\_median\_25}{\trainpredset}{\transcma}{\ftDisp}$ &          15 &    99.50 \\
\midrule
\rowcolor[gray]{0.8500}
                       $\tableFeat{evopath\_c\_norm}{}{}{\ftCMA}$ &           0 &   100.00 \\
\midrule
                                         $\tableFeat{obs}{\archive}{}{}$ &           1 &   100.00 \\
                             $\tableFeat{generation}{}{}{\ftCMA}$ &           0 &   100.00 \\
                                     $\tableFeat{obs}{\archivepred}{}{}$ &          13 &   100.00 \\
\rowcolor[gray]{0.8500}
                                        $\tableFeat{obs}{\trainset}{}{}$ &           1 &   100.00 \\
                                    $\tableFeat{obs}{\trainpredset}{}{}$ &          13 &   100.00 \\
\midrule
              $\tableFeat{cma\_lik}{\archive}{\transcma}{\ftCMA}$ &           1 &    99.96 \\
\rowcolor[gray]{0.8500}
          $\tableFeat{cma\_lik}{\archivepred}{\transcma}{\ftCMA}$ &          13 &    99.96 \\
             $\tableFeat{cma\_lik}{\trainset}{\transcma}{\ftCMA}$ &           1 &    99.96 \\
         $\tableFeat{cma\_lik}{\trainpredset}{\transcma}{\ftCMA}$ &          13 &    99.96 \\
\midrule
\rowcolor[gray]{0.8500}
                       $\tableFeat{evopath\_s\_norm}{}{}{\ftCMA}$ &           0 &   100.00 \\
\midrule
\rowcolor[gray]{0.8500}
                                                 $\tableFeat{dim}{}{}{}$ &           0 &   100.00 \\
\midrule
\rowcolor[gray]{0.8500}
                             $\tableFeat{step\_size}{}{}{\ftCMA}$ &           0 &   100.00 \\
\bottomrule
\end{tabular}

%% file: tex/corr_dim_nearest_3.tex
\begin{tabular}{lrr}
\toprule
                                                                       & $N_\nanout$ & rob.(\%) \\
\midrule
   $\tableFeat{quad\_simple\_cond}{\archive}{\transcma}{\ftMM}$ &           6 &    95.21 \\
\rowcolor[gray]{0.8500}
  $\tableFeat{quad\_simple\_cond}{\trainset}{\transcma}{\ftMM}$ &           6 &    95.21 \\
\midrule
 $\tableFeat{lda\_qda\_25}{\archivepred}{\transnone}{\ftLevel}$ &          15 &    94.13 \\
  $\tableFeat{lda\_qda\_25}{\archivepred}{\transcma}{\ftLevel}$ &          15 &    94.12 \\
\rowcolor[gray]{0.8500}
$\tableFeat{lda\_qda\_25}{\trainpredset}{\transnone}{\ftLevel}$ &          15 &    94.13 \\
 $\tableFeat{lda\_qda\_25}{\trainpredset}{\transcma}{\ftLevel}$ &          15 &    94.12 \\
\midrule
\rowcolor[gray]{0.8500}
                              $\tableFeat{restart}{}{}{\ftCMA}$ &           0 &   100.00 \\
\bottomrule
\end{tabular}

%% file: tex/corr_dim_nearest_4.tex
\begin{tabular}{lrr}
\toprule
                                                                         & $N_\nanout$ & rob.(\%) \\
\midrule
       $\tableFeat{diff\_mean\_02}{\archive}{\transcma}{\ftDisp}$ &          71 &    99.52 \\
     $\tableFeat{diff\_median\_02}{\archive}{\transcma}{\ftDisp}$ &          71 &    99.44 \\
   $\tableFeat{diff\_mean\_02}{\archivepred}{\transcma}{\ftDisp}$ &          86 &    99.52 \\
 $\tableFeat{diff\_median\_02}{\archivepred}{\transcma}{\ftDisp}$ &          86 &    99.45 \\
      $\tableFeat{diff\_mean\_02}{\trainset}{\transcma}{\ftDisp}$ &          71 &    99.52 \\
\rowcolor[gray]{0.8500}
    $\tableFeat{diff\_median\_02}{\trainset}{\transcma}{\ftDisp}$ &          71 &    99.44 \\
  $\tableFeat{diff\_mean\_02}{\trainpredset}{\transcma}{\ftDisp}$ &          86 &    99.52 \\
$\tableFeat{diff\_median\_02}{\trainpredset}{\transcma}{\ftDisp}$ &          86 &    99.45 \\
\bottomrule
\end{tabular}

%% file: tex/corr_dim_knn_1.tex
\begin{tabular}{lrr}
\toprule
                                                                          & $N_\nanout$ & rob.(\%) \\
\midrule
                              $\tableFeat{step\_size}{}{}{\ftCMA}$ &           0 &   100.00 \\
              $\tableFeat{cma\_lik}{\archive}{\transnone}{\ftCMA}$ &           1 &    99.75 \\
        $\tableFeat{diff\_mean\_05}{\archive}{\transcma}{\ftDisp}$ &          27 &    99.49 \\
      $\tableFeat{diff\_median\_05}{\archive}{\transcma}{\ftDisp}$ &          27 &    99.46 \\
\rowcolor[gray]{0.8500}
        $\tableFeat{diff\_mean\_10}{\archive}{\transcma}{\ftDisp}$ &          12 &    99.52 \\
      $\tableFeat{diff\_median\_10}{\archive}{\transcma}{\ftDisp}$ &          12 &    99.48 \\
        $\tableFeat{diff\_mean\_25}{\archive}{\transcma}{\ftDisp}$ &           6 &    99.55 \\
      $\tableFeat{diff\_median\_25}{\archive}{\transcma}{\ftDisp}$ &           6 &    99.49 \\
          $\tableFeat{cma\_lik}{\archivepred}{\transnone}{\ftCMA}$ &          13 &    99.75 \\
    $\tableFeat{diff\_mean\_05}{\archivepred}{\transcma}{\ftDisp}$ &          42 &    99.49 \\
  $\tableFeat{diff\_median\_05}{\archivepred}{\transcma}{\ftDisp}$ &          42 &    99.48 \\
    $\tableFeat{diff\_mean\_10}{\archivepred}{\transcma}{\ftDisp}$ &          25 &    99.52 \\
  $\tableFeat{diff\_median\_10}{\archivepred}{\transcma}{\ftDisp}$ &          25 &    99.49 \\
    $\tableFeat{diff\_mean\_25}{\archivepred}{\transcma}{\ftDisp}$ &          15 &    99.55 \\
  $\tableFeat{diff\_median\_25}{\archivepred}{\transcma}{\ftDisp}$ &          15 &    99.50 \\
$\tableFeat{lin\_simple\_coef\_max}{\trainset}{\transnone}{\ftMM}$ &           6 &    98.12 \\
       $\tableFeat{cma\_mean\_dist}{\trainset}{\transcma}{\ftCMA}$ &           6 &    98.40 \\
   $\tableFeat{cma\_mean\_dist}{\trainpredset}{\transcma}{\ftCMA}$ &          13 &    98.71 \\
\midrule
               $\tableFeat{cma\_lik}{\archive}{\transcma}{\ftCMA}$ &           1 &    99.96 \\
\rowcolor[gray]{0.8500}
           $\tableFeat{cma\_lik}{\archivepred}{\transcma}{\ftCMA}$ &          13 &    99.96 \\
              $\tableFeat{cma\_lik}{\trainset}{\transcma}{\ftCMA}$ &           1 &    99.95 \\
          $\tableFeat{cma\_lik}{\trainpredset}{\transcma}{\ftCMA}$ &          13 &    99.96 \\
\midrule
             $\tableFeat{cma\_lik}{\trainset}{\transnone}{\ftCMA}$ &           1 &    98.81 \\
\rowcolor[gray]{0.8500}
         $\tableFeat{cma\_lik}{\trainpredset}{\transnone}{\ftCMA}$ &          13 &    98.80 \\
\midrule
\rowcolor[gray]{0.8500}
                                                  $\tableFeat{dim}{}{}{}$ &           0 &   100.00 \\
                        $\tableFeat{evopath\_c\_norm}{}{}{\ftCMA}$ &           0 &   100.00 \\
  $\tableFeat{cma\_mean\_dist}{\trainpredset}{\transnone}{\ftCMA}$ &          13 &    93.96 \\
\midrule
\rowcolor[gray]{0.8500}
       $\tableFeat{diff\_mean\_05}{\trainset}{\transcma}{\ftDisp}$ &          30 &    96.80 \\
     $\tableFeat{diff\_median\_05}{\trainset}{\transcma}{\ftDisp}$ &          30 &    95.80 \\
   $\tableFeat{diff\_mean\_05}{\trainpredset}{\transcma}{\ftDisp}$ &          53 &    96.85 \\
 $\tableFeat{diff\_median\_05}{\trainpredset}{\transcma}{\ftDisp}$ &          53 &    95.82 \\
\bottomrule
\end{tabular}

%% file: tex/corr_dim_knn_2.tex
\begin{tabular}{lrr}
\toprule
                                                                          & $N_\nanout$ & rob.(\%) \\
\midrule
\rowcolor[gray]{0.8500}
                        $\tableFeat{evopath\_s\_norm}{}{}{\ftCMA}$ &           0 &   100.00 \\
\midrule
\rowcolor[gray]{0.8500}
    $\tableFeat{lda\_qda\_25}{\archivepred}{\transnone}{\ftLevel}$ &          15 &    94.13 \\
     $\tableFeat{lda\_qda\_25}{\archivepred}{\transcma}{\ftLevel}$ &          15 &    94.12 \\
\midrule
$\tableFeat{lin\_simple\_coef\_min}{\trainset}{\transnone}{\ftMM}$ &           6 &    97.40 \\
\rowcolor[gray]{0.8500}
            $\tableFeat{eps\_max}{\trainset}{\transnone}{\ftInfo}$ &           6 &    97.84 \\
        $\tableFeat{eps\_max}{\trainpredset}{\transnone}{\ftInfo}$ &          13 &    97.70 \\
\midrule
\rowcolor[gray]{0.8500}
    $\tableFeat{quad\_simple\_cond}{\trainset}{\transnone}{\ftMM}$ &           6 &    92.55 \\
\midrule
\rowcolor[gray]{0.8500}
                                 $\tableFeat{restart}{}{}{\ftCMA}$ &           0 &   100.00 \\
\midrule
       $\tableFeat{diff\_mean\_10}{\trainset}{\transcma}{\ftDisp}$ &          15 &    96.96 \\
     $\tableFeat{diff\_median\_10}{\trainset}{\transcma}{\ftDisp}$ &          15 &    96.09 \\
       $\tableFeat{diff\_mean\_25}{\trainset}{\transcma}{\ftDisp}$ &           6 &    97.06 \\
     $\tableFeat{diff\_median\_25}{\trainset}{\transcma}{\ftDisp}$ &           6 &    96.27 \\
\rowcolor[gray]{0.8500}
   $\tableFeat{diff\_mean\_10}{\trainpredset}{\transcma}{\ftDisp}$ &          28 &    97.02 \\
 $\tableFeat{diff\_median\_10}{\trainpredset}{\transcma}{\ftDisp}$ &          28 &    96.16 \\
   $\tableFeat{diff\_mean\_25}{\trainpredset}{\transcma}{\ftDisp}$ &          15 &    97.12 \\
 $\tableFeat{diff\_median\_25}{\trainpredset}{\transcma}{\ftDisp}$ &          15 &    96.36 \\
\midrule
                                          $\tableFeat{obs}{\archive}{}{}$ &           1 &   100.00 \\
                              $\tableFeat{generation}{}{}{\ftCMA}$ &           0 &   100.00 \\
      $\tableFeat{quad\_simple\_cond}{\archive}{\transcma}{\ftMM}$ &           6 &    95.21 \\
\rowcolor[gray]{0.8500}
                                      $\tableFeat{obs}{\archivepred}{}{}$ &          13 &   100.00 \\
                                         $\tableFeat{obs}{\trainset}{}{}$ &           1 &    92.16 \\
                                     $\tableFeat{obs}{\trainpredset}{}{}$ &          13 &    92.26 \\
\midrule
\rowcolor[gray]{0.8500}
    $\tableFeat{lda\_qda\_10}{\archivepred}{\transnone}{\ftLevel}$ &          18 &    92.37 \\
     $\tableFeat{lda\_qda\_10}{\archivepred}{\transcma}{\ftLevel}$ &          18 &    92.38 \\
\midrule
        $\tableFeat{diff\_mean\_02}{\archive}{\transcma}{\ftDisp}$ &          71 &    99.52 \\
\rowcolor[gray]{0.8500}
      $\tableFeat{diff\_median\_02}{\archive}{\transcma}{\ftDisp}$ &          71 &    99.44 \\
    $\tableFeat{diff\_mean\_02}{\archivepred}{\transcma}{\ftDisp}$ &          86 &    99.52 \\
  $\tableFeat{diff\_median\_02}{\archivepred}{\transcma}{\ftDisp}$ &          86 &    99.45 \\
\bottomrule
\end{tabular}

%% file: tex/gpcovTable.tex
\begin{table}
  \caption{Experimental settings of GP covariances:
    $\sigma_0^2$ -- bias constant term,
    $d$ -- metric $d(\xx_p, \xx_q)$,
    $\lscale$ -- characteristic length-scale,
    $\vmat{P}$ -- isotropic distance matrix $\vmat{P} = \lscale^{-2}\vmat{I}$,
    $\tilde{\xx}_p, \tilde{\xx}_q$ -- inputs augmented by a bias unit,
    $\lscale(\xx)$ -- linear function of $\xx$.
  }
  \label{tab:gpcov}
  \centering
  \begin{tabular}{ p{3.3cm}l }
    \toprule
    name & covariance function \\
    \cmidrule(r){1-1}
    \cmidrule(l){2-2}
    linear &
      $\covLIN(\xx_p, \xx_q) = \sigma_0^2 + \xx_p^\top \xx_q$ \\
    quadratic &
      $\covQUAD(\xx_p, \xx_q) = (\sigma_0^2 + \xx_p^\top \xx_q)^2$ \\
    squared-exponential & 
      $\covSE(d; \sigma_f, \lscale) = \sigma_f^2\exp\left(-\frac{d^2}{2\lscale^2}\right)$ \\
    Mat\'{e}rn $\frac{5}{2}$ & 
      $\covMat{5}(d; \sigma_f, \lscale) = \sigma_f^2\left(1 + \frac{\sqrt{5}d}{\lscale} + \frac{5d^2}{3\lscale^2}\right)\exp\left(-\frac{\sqrt{5}d}{\lscale}\right)$ \\
    rational quadratic &
      $\covRQ(d; \sigma_f, \lscale) = \sigma_f^2\left(1 + \frac{d^2}{2\lscale^2\alpha}\right)^{-\alpha}, \quad \alpha > 0$ \\
    neural network &
      $\covNN(\xx_p, \xx_q) =
        \sigma_f^2\,
        \mathrm{arcsin}\left(\frac{2\tilde{\xx}_p^\top \vmat{P} \tilde{\xx}_q}
                         {\sqrt{(1+2\tilde{\xx}_p^\top \vmat{P} \tilde{\xx}_p)
                                (1+2\tilde{\xx}_q^\top \vmat{P} \tilde{\xx}_q)}}\right)$ \\
    spatially varying length-scale \citep{gibbs1997bayesian} &
      $\covGIBBS(\xx_p, \xx_q) = \sigma_f^2 \left(\frac{2 \lscale(\xx_p) \lscale(\xx_q)}{\lscale(\xx_p)^2 + \lscale(\xx_q)^2}\right)^{\frac{\dm}{2}}
                               \exp \left( - \frac{(\xx_p - \xx_q)^\top (\xx_p - \xx_q)}{\lscale(\xx_p)^2 + \lscale(\xx_q)^2}\right)$ \\
    composite &
      $\covSEQUAD = \covSE + \covQUAD$ \\
    \bottomrule
  \end{tabular}
\end{table}

%% file: tex/rfSetTable.tex
\begin{table}
  \caption{Experimental settings of RF for 5 splitting methods found using latin-hypercube design
    on 100 out of 400 combinations of
    the number of trees in RF
    $n_\text{tree} \in \{2^6, 2^7, 2^8, 2^9, 2^{10}\}$,
    the number of points bootstrapped out of $\npt$ training points
    $\npt_t \in \lceil\{\sfrac{1}{4}, \sfrac{1}{2}, \sfrac{3}{4}, 1\}\cdot \npt \rceil$,
    and the number of randomly subsampled dimensions used for training the individual tree
    $n_\dm \in \lceil\{\sfrac{1}{4}, \sfrac{1}{2}, \sfrac{3}{4}, 1\}\cdot \dm \rceil$.
    The maximum tree depth was set to
    $8$, in accordance with~\citep{chen2016xgboost}.
    The remaining decision tree parameters have been taken
    identical to settings from~\citep{pitra2018boosted}.
    The winning settings
    on the validation dataset $\set{D}_\text{val}$
    are shown in the format $[n_\text{tree}, \npt_t, n_\dm]$ (in case of $\npt_t$ and $n_\dm$ are shown only the multipliers of $\npt$ and $\dm$ respectively).
    Preliminary testing showed that RFs
    performance in connection with DTS-CMA-ES is higher
    when the input data are not transformed to the $\sigma^2\CC$ basis.
    Thus, the transformation step is omitted during model training
    for both TSS.
  }
  \label{tab:rfSet}
  \centering
  
  \begin{tabular}{ p{0.8cm}p{1.4cm}p{1.2cm}p{1.8cm}p{1.6cm}p{1.6cm}p{1.9cm} }
    \toprule
    TSS & error measure & 
      CART \citep{breiman1984classification} & 
      SCRT\qquad\citep{dobra2002secret}           & 
      OC1  \citep{murthy1994system}          & 
      PAIR \citep{chaudhuri1994piecewise}    & 
      SUPP\qquad\citep{hinton1996using}          \\ 
    \cmidrule(l){1-1}
    \cmidrule(lr){2-2}
    \cmidrule(lr){3-7}
    \multirow{2}{1cm}{\hyperlink{tss:full}{full}}    & $\mse$ &
    $[2^{10}, 1, \sfrac{3}{4}]$ & $[2^8, \sfrac{1}{4}, \sfrac{1}{2}]$ & $[2^7, 1, 1]$ & $[2^7, 1, \sfrac{3}{4}]$ & $[2^{7}, 1, \sfrac{1}{4}]$ \\
    {} & $\rde$ & $[2^6, \sfrac{3}{4}, 1]$ & $[2^8, \sfrac{3}{4}, 1]$ & $[2^7, 1, 1]$ & $[2^{10}, \sfrac{3}{4}, 1]$ & $[2^6, \sfrac{3}{4}, 1]$ \\
    \multirow{2}{1cm}{\hyperlink{tss:nearest}{nearest}} & $\mse$ &
    $[2^8, 1, \sfrac{3}{4}]$ & $[2^9, \sfrac{1}{2}, 1]$ & $[2^7, 1, 1]$ & $[2^7, 1, \sfrac{3}{4}]$ & $[2^{10}, \sfrac{1}{4}, 1]$ \\
    {} & $\rde$  & $[2^{10}, \sfrac{3}{4}, 1]$ & $[2^8, \sfrac{3}{4}, 1]$ & $[2^7, 1, 1]$ & $[2^{10}, \sfrac{3}{4}, 1]$ & $[2^{6}, \sfrac{3}{4}, 1]$ \\
    \bottomrule
  \end{tabular}
\end{table}

%% file: tex/defFile.tex
\newlength{\dueltabcolw}
\newlength{\savetabcolsep}
\newlength{\savecmidrulekern}
\newlength{\headcolw}
\newlength{\astwidth}
\newcolumntype{R}{>{\raggedleft\arraybackslash}m{\dueltabcolw}}
\newcolumntype{L}{>{\raggedright\arraybackslash}m{\dueltabcolw}}
\newcolumntype{H}{>{\raggedright\arraybackslash}m{\headcolw}}

%% file: tex/modelNanTable.tex
\begin{table}

\setlength{\savetabcolsep}{\tabcolsep}
\setlength{\savecmidrulekern}{\cmidrulekern}

\setlength{\headcolw}{1.0cm}
\setlength{\tabcolsep}{0pt}
\setlength{\cmidrulekern}{2pt}
\setlength{\dueltabcolw}{1.0\textwidth-\headcolw}
\setlength{\dueltabcolw}{\dueltabcolw/19}

\centering
\caption{
  Percentages of cases when the model did not provide usable
prediction (model not trained, its prediction failed, or prediction is constant).}

\label{tab:modelNanTable}
\small
\begin{tabular}{HRRRRRRRRRRRRRRRRRRR}
\toprule
 &\multicolumn{8}{l}{\parbox{8\dueltabcolw}{\centering GP}} & \multicolumn{9}{l}{\parbox{9\dueltabcolw}{\centering RF}} & \multicolumn{1}{l}{\parbox{1\dueltabcolw}{\centering lmm}} & \multicolumn{1}{l}{\parbox{1\dueltabcolw}{\centering lq}}\\
\cmidrule(lr){2-9}
\cmidrule(lr){10-18}
\cmidrule(lr){19-19}
\cmidrule(lr){20-20}
TSS & ${{{}}}^\text{Gibbs}$ & ${{{}}}^\text{LIN}$ & ${{{}}}^\text{Mat}$ & ${{{}}}^\text{NN}$ & ${{{}}}^\text{Q}$ & ${{{}}}^\text{RQ}$ & ${{{}}}^\text{SE}$ & ${{{}}}^\text{SE+Q}$ & ${{}}^\text{CART}_\text{MSE}$ & ${{}}^\text{CART}_\text{RDE}$ & ${{}}^\text{SCRT}_\text{MSE}$ & ${{}}^\text{SCRT}_\text{RDE}$ & ${{}}^\text{OC1}$ & ${{}}^\text{PAIR}_\text{MSE}$ & ${{}}^\text{PAIR}_\text{RDE}$ & ${{}}^\text{SUPP}_\text{MSE}$ & ${{}}^\text{SUPP}_\text{RDE}$ & ${}$ & ${}$\\
\midrule
\hyperlink{tss:full}{full}       & 50.3 & 8.4 & 26.7 & 7.6 & 22.1 & 5.3 & 8.5 & 3.3 & 5.2 & 9.6 & 2.3 & 20.4 & 17.9 & 18.8 & 2.3 & 7.0 & 23.6 & 8.8 & 2.5\\
\hyperlink{tss:nearest}{nearest} & 40.9 & 29.5 & 27.8 & 24.1 & 19.8 & 4.5 & 7.6 & 3.1 & 23.0 & 22.7 & 23.4 & 23.3 & 23.4 & 22.9 & 23.0 & 23.7 & 23.1 & 10.3 & 2.4\\
\hyperlink{tss:knn}{knn}         & --- & --- & --- & --- & --- & --- & --- & --- & --- & --- & --- & --- & --- & --- & --- & --- & --- & 5.5 & ---\\
\end{tabular}
\setlength{\tabcolsep}{\savetabcolsep}
\setlength{\cmidrulekern}{\savecmidrulekern}
\normalsize
\end{table}

%% file: tex/duelTable_rde.tex
\begin{sidewaystable}
\setlength{\savetabcolsep}{\tabcolsep}
\setlength{\savecmidrulekern}{\cmidrulekern}

\setlength{\headcolw}{0.8cm}
\setlength{\tabcolsep}{0pt}
\setlength{\cmidrulekern}{2pt}
\setlength{\dueltabcolw}{1.3\textwidth-\headcolw-1560\tabcolsep}
\setlength{\dueltabcolw}{\dueltabcolw/39}

\settowidth{\astwidth}{${}^{\ast}$}
\centering
\caption{
  A pairwise comparison of the model settings RDE in different TSS.
The percentage of wins of $i$-th model setting against $j$-th model setting over all available data is given in the $i$-th row and $j$-th column.
  The numbers in bold mark the row model setting being significantly better than the column model setting according to the two-sided Wilcoxon signed rank test with the Holm correction at family-wise significance level $\alpha=0.05$.
}

\label{tab:duelTable_rde}
\resizebox{\textwidth}{!}{%
\footnotesize
\begin{tabular}{ HHHLLLLLLLLLLLLLLLLLLLLLLLLLLLLLLLLLLLLLLL }
\toprule
\multicolumn{2}{c}{\multirow{2}{*}{\textbf{RDE}}} & TSS & \multicolumn{19}{l}{\parbox{19\dueltabcolw}{\centering full}} & \multicolumn{19}{l}{\parbox{19\dueltabcolw}{\centering nearest}} & \multicolumn{1}{l}{\parbox{1\dueltabcolw}{\centering knn}}\\
\cmidrule(lr){4-22}
\cmidrule(lr){23-41}
\cmidrule(lr){42-42}
\multicolumn{2}{c}{} & model & \multicolumn{8}{l}{\parbox{8\dueltabcolw}{\centering GP}} & \multicolumn{9}{l}{\parbox{9\dueltabcolw}{\centering RF}} & \multicolumn{1}{l}{\parbox{1\dueltabcolw}{\centering lmm}} & \multicolumn{1}{l}{\parbox{1\dueltabcolw}{\centering lq}} & \multicolumn{8}{l}{\parbox{8\dueltabcolw}{\centering GP}} & \multicolumn{9}{l}{\parbox{9\dueltabcolw}{\centering RF}} & \multicolumn{1}{l}{\parbox{1\dueltabcolw}{\centering lmm}} & \multicolumn{1}{l}{\parbox{1\dueltabcolw}{\centering lq}} & \multicolumn{1}{l}{\parbox{1\dueltabcolw}{\centering lmm}}\\
\bottomrule
\end{tabular}
}
\setlength{\tabcolsep}{\savetabcolsep}
\setlength{\cmidrulekern}{\savecmidrulekern}
\end{sidewaystable}

%% file: tex/tree_full_postproc.tex
\begin{tikzpicture}[>=latex,line join=bevel,]
\node (24) at (465.0bp,8.5bp) [draw,fill=LeafColor,rectangle] {$\text{GP}_\text{Gibbs}$};
  \node (25) at (537.0bp,8.5bp) [draw,fill=LeafColor,rectangle] {$\text{RF}^\text{OC1}$};
  \node (26) at (629.0bp,8.5bp) [draw,fill=LeafColor,rectangle] {$\text{RF}^\text{OC1}$};
  \node (27) at (704.0bp,8.5bp) [draw,fill=LeafColor,rectangle] {$\text{GP}_\text{Gibbs}$};
  \node (20) at (653.0bp,66.0bp) [draw,fill=InnerNodeColor,ellipse] {$\tableFeat{diff\_mean\_25}{\archivepred}{\transcma}{\ftDisp}$};
  \node (21) at (779.0bp,66.0bp) [draw,fill=InnerNodeColor,ellipse] {$\tableFeat{obs}{\archive}{}{}$};
  \node (22) at (213.0bp,8.5bp) [draw,fill=LeafColor,rectangle] {$\text{GP}_\text{RQ}$};
  \node (23) at (277.0bp,8.5bp) [draw,fill=LeafColor,rectangle] {$\text{lq}$};
  \node (28) at (779.0bp,8.5bp) [draw,fill=LeafColor,rectangle] {$\text{GP}_\text{Mat}$};
  \node (29) at (867.0bp,8.5bp) [draw,fill=LeafColor,rectangle] {$\text{RF}^\text{OC1}$};
  \node (1) at (341.0bp,301.0bp) [draw,fill=InnerNodeColor,ellipse] {$\tableFeat{diff\_mean\_25}{\archivepred}{\transcma}{\ftDisp}$};
  \node (3) at (426.0bp,244.0bp) [draw,fill=InnerNodeColor,ellipse] {$\tableFeat{lda\_qda\_10}{\archivepred}{\transnone}{\ftLevel}$};
  \node (2) at (273.0bp,244.0bp) [draw,fill=InnerNodeColor,ellipse] {$\tableFeat{diff\_mean\_25}{\archivepred}{\transcma}{\ftDisp}$};
  \node (5) at (273.0bp,185.0bp) [draw,fill=InnerNodeColor,ellipse] {$\tableFeat{quad\_simple\_cond}{\archive}{\transcma}{\ftMM}$};
  \node (4) at (121.0bp,185.0bp) [draw,fill=InnerNodeColor,ellipse] {$\tableFeat{cma\_lik}{\archivepred}{\transcma}{\ftCMA}$};
  \node (7) at (548.0bp,185.0bp) [draw,fill=InnerNodeColor,ellipse] {$\tableFeat{lda\_qda\_25}{\archivepred}{\transnone}{\ftLevel}$};
  \node (6) at (426.0bp,185.0bp) [draw,fill=InnerNodeColor,ellipse] {$\tableFeat{evopath\_c\_norm}{}{}{\ftCMA}$};
  \node (9) at (121.0bp,126.0bp) [draw,fill=LeafColor,rectangle] {$\text{lmm}$};
  \node (8) at (21.0bp,126.0bp) [draw,fill=LeafColor,rectangle] {$\text{RF}^\text{SUPP}_\text{RDE}$};
  \node (11) at (304.0bp,126.0bp) [draw,fill=InnerNodeColor,ellipse] {$\tableFeat{diff\_mean\_25}{\archivepred}{\transcma}{\ftDisp}$};
  \node (10) at (215.0bp,126.0bp) [draw,fill=LeafColor,rectangle] {$\text{GP}_\text{Mat}$};
  \node (13) at (460.0bp,126.0bp) [draw,fill=LeafColor,rectangle] {$\text{GP}_\text{NN}$};
  \node (12) at (393.0bp,126.0bp) [draw,fill=LeafColor,rectangle] {$\text{GP}_\text{Mat}$};
  \node (15) at (653.0bp,126.0bp) [draw,fill=InnerNodeColor,ellipse] {$\tableFeat{dim}{}{}{}$};
  \node (14) at (548.0bp,126.0bp) [draw,fill=InnerNodeColor,ellipse] {$\tableFeat{diff\_mean\_25}{\archivepred}{\transcma}{\ftDisp}$};
  \node (17) at (341.0bp,66.0bp) [draw,fill=LeafColor,rectangle] {$\text{lq}$};
  \node (16) at (245.0bp,66.0bp) [draw,fill=InnerNodeColor,ellipse] {$\tableFeat{quad\_simple\_cond}{\archive}{\transcma}{\ftMM}$};
  \node (19) at (537.0bp,66.0bp) [draw,fill=InnerNodeColor,ellipse] {$\tableFeat{dim}{}{}{}$};
  \node (18) at (473.0bp,66.0bp) [draw,fill=LeafColor,rectangle] {$\text{RF}^\text{OC1}$};
  \draw [] (5) -- node[right, outer sep=2bp] {$< 8.84$} (10);
  \draw [] (16) -- node[right, outer sep=3bp] {$\geq 357$} (23);
  \draw [] (7) -- node[right, outer sep=12bp] {$\geq 1.06$} (15);
  \draw [] (20) -- node[left, outer sep=0bp] {$-2.96>$} (26);
  \draw [] (1) -- node[right, outer sep=2bp] {$< -114$} (2);
  \draw [] (16) -- node[right, outer sep=0bp] {$< 357$} (22);
  \draw [] (15) -- node[right, outer sep=1bp] {$\in \{2,3,5\}$} (20);
  \draw [] (7) -- node[right, outer sep=2bp] {$< 1.06$} (14);
  \draw [] (11) -- node[right, outer sep=4bp] {$\geq -2.01\cdot10^{3}$} (17);
  \draw [] (20) -- node[right, outer sep=3bp] {$\geq -2.96$} (27);
  \draw [] (2) -- node[right, outer sep=5bp] {$< -6.02\cdot10^{4}$} (4);
  \draw [] (14) -- node[left, outer sep=2bp] {$-20.9>$} (18);
  \draw [] (3) -- node[right, outer sep=12bp] {$\geq 0.12$} (7);
  \draw [] (15) -- node[right, outer sep=10bp] {$\in \{10,20\}$} (21);
  \draw [] (19) -- node[left, outer sep=4bp] {$\{2,3\} \ni$} (24);
  \draw [] (21) -- node[right, outer sep=12bp] {$\geq 1.08\cdot10^{3}$} (29);
  \draw [] (11) -- node[left, outer sep=3bp] {$ -2.01\cdot10^{3}>$} (16);
  \draw [] (6) -- node[right, outer sep=0bp] {$<\!1.27$} (12);
  \draw [] (2) -- node[right, outer sep=2bp] {$\geq -6.02\cdot10^{4}$} (5);
  \draw [] (14) -- node[right, outer sep=2bp] {$\geq -20.9$} (19);
  \draw [] (3) -- node[right, outer sep=2bp] {$< 0.12$} (6);
  \draw [] (4) -- node[left, outer sep=6bp] {$-3.76\cdot10^{26}>$} (8);
  \draw [] (19) -- node[right, outer sep=2bp] {$\in \{5,10,20\}$} (25);
  \draw [] (21) -- node[left, outer sep=1bp] {$1.08\!\cdot\!10^{3}\!>$} (28);
  \draw [] (6) -- node[right, outer sep=2bp] {$\geq 1.27$} (13);
  \draw [] (5) -- node[right, outer sep=2bp] {$\geq 8.84$} (11);
  \draw [] (1) -- node[right, outer sep=8bp] {$\geq -114$} (3);
  \draw [] (4) -- node[right, outer sep=2bp] {$\geq -3.76\cdot10^{26}$} (9);
\end{tikzpicture}

%% file: tex/tree_nearest_postproc.tex
\begin{tikzpicture}[>=latex,line join=bevel,]
\node (20) at (346.0bp,62.0bp) [draw,fill=LeafColor,rectangle] {$\text{GP}_\text{NN}$};
  \node (21) at (423.0bp,62.0bp) [draw,fill=InnerNodeColor,ellipse] {$\tableFeat{lda\_qda\_25}{\trainpredset}{\transnone}{\ftLevel}$};
  \node (22) at (390.0bp,8.0bp) [draw,fill=LeafColor,rectangle] {$\text{RF}^\text{OC1}$};
  \node (23) at (457.0bp,8.0bp) [draw,fill=LeafColor,rectangle] {$\text{GP}_\text{Mat}$};
  \node (1) at (196.0bp,351.0bp) [draw,fill=InnerNodeColor,ellipse] {$\tableFeat{diff\_mean\_25}{\archivepred}{\transcma}{\ftDisp}$};
  \node (3) at (268.0bp,294.0bp) [draw,fill=InnerNodeColor,ellipse] {$\tableFeat{lda\_qda\_25}{\trainpredset}{\transnone}{\ftLevel}$};
  \node (2) at (121.0bp,294.0bp) [draw,fill=InnerNodeColor,ellipse] {$\tableFeat{cma\_lik}{\archivepred}{\transcma}{\ftCMA}$};
  \node (5) at (100.0bp,236.0bp) [draw,fill=InnerNodeColor,ellipse] {$\tableFeat{dim}{}{}{}$};
  \node (4) at (21.0bp,236.0bp) [draw,fill=LeafColor,rectangle] {$\text{RF}^\text{OC1}$};
  \node (7) at (360.0bp,236.0bp) [draw,fill=InnerNodeColor,ellipse] {$\tableFeat{obs}{\trainset}{}{}$};
  \node (6) at (268.0bp,236.0bp) [draw,fill=InnerNodeColor,ellipse] {$\tableFeat{obs}{\trainset}{}{}$};
  \node (9) at (98.0bp,177.0bp) [draw,fill=InnerNodeColor,ellipse] {$\tableFeat{lda\_qda\_25}{\trainpredset}{\transnone}{\ftLevel}$};
  \node (8) at (18.0bp,177.0bp) [draw,fill=LeafColor,rectangle] {$\text{GP}_\text{Mat}$};
  \node (11) at (299.0bp,177.0bp) [draw,fill=InnerNodeColor,ellipse] {$\tableFeat{dim}{}{}{}$};
  \node (10) at (222.0bp,177.0bp) [draw,fill=InnerNodeColor,ellipse] {$\tableFeat{lda\_qda\_25}{\trainpredset}{\transnone}{\ftLevel}$};
  \node (13) at (424.0bp,177.0bp) [draw,fill=LeafColor,rectangle] {$\text{GP}_\text{Mat}$};
  \node (12) at (360.0bp,177.0bp) [draw,fill=LeafColor,rectangle] {$\text{GP}_\text{Gibbs}$};
  \node (15) at (98.0bp,118.0bp) [draw,fill=LeafColor,rectangle] {$\text{GP}_\text{Mat}$};
  \node (14) at (31.0bp,118.0bp) [draw,fill=LeafColor,rectangle] {$\text{RF}^\text{SUPP}_\text{RDE}$};
  \node (17) at (232.0bp,118.0bp) [draw,fill=LeafColor,rectangle] {$\text{GP}_\text{Gibbs}$};
  \node (16) at (165.0bp,118.0bp) [draw,fill=LeafColor,rectangle] {$\text{GP}_\text{Mat}$};
  \node (19) at (381.0bp,118.0bp) [draw,fill=InnerNodeColor,ellipse] {$\tableFeat{obs}{\trainset}{}{}$};
  \node (18) at (299.0bp,118.0bp) [draw,fill=LeafColor,rectangle] {$\text{GP}_\text{Gibbs}$};
  \draw [] (7) -- node[left, outer sep=2bp] {$107 >$} (12);
  \draw [] (5) -- node[left, outer sep=8bp] {$\{2,3,5\} \ni$} (8);
  \draw [] (9) -- node[right, outer sep=2bp] {$\geq 1.14$} (15);
  \draw [] (6) -- node[right, outer sep=2bp] {$< 40$} (10);
  \draw [] (1) -- node[right, outer sep=2bp] {$< -779$} (2);
  \draw [] (11) -- node[right, outer sep=2bp] {$\in \{10,20\}$} (19);
  \draw [] (2) -- node[right, outer sep=2bp] {$\geq -5.31\cdot10^{24}$} (5);
  \draw [] (21) -- node[right, outer sep=4bp] {$\geq 0.97$} (23);
  \draw [] (9) -- node[left, outer sep=2bp] {$1.14 >$} (14);
  \draw [] (6) -- node[right, outer sep=4bp] {$\geq 40$} (11);
  \draw [] (10) -- node[left, outer sep=2bp] {$0.39 >$} (16);
  \draw [] (11) -- node[left, inner sep=1bp] {$\{2,3,5\}\!\!\ni$} (18);
  \draw [] (19) -- node[right, outer sep=2bp] {$< 136$} (20);
  \draw [] (3) -- node[right, outer sep=10bp] {$\geq 1.24$} (7);
  \draw [] (21) -- node[left, outer sep=2bp] {$0.97 >$} (22);
  \draw [] (10) -- node[right, outer sep=0bp] {$\geq\!0.39$} (17);
  \draw [] (19) -- node[right, outer sep=5bp] {$\geq 136$} (21);
  \draw [] (3) -- node[right, outer sep=2bp] {$< 1.24$} (6);
  \draw [] (7) -- node[right, outer sep=8bp] {$\geq 107$} (13);
  \draw [] (5) -- node[right, outer sep=2bp] {$\in \{10,20\}$} (9);
  \draw [] (2) -- node[left, outer sep=7bp] {$-5.31\cdot10^{24}>$} (4);
  \draw [] (1) -- node[right, outer sep=7bp] {$\geq -779$} (3);
\end{tikzpicture}

%% file: tex/feat_full_1.tex
\newcommand{\PreserveBackslash}[1]{\let\temp=\\#1\let\\=\temp}
\newcolumntype{C}[1]{>{\PreserveBackslash\centering}p{#1}}
\newcolumntype{R}[1]{>{\PreserveBackslash\raggedleft}p{#1}}
\newcolumntype{L}[1]{>{\PreserveBackslash\raggedright}p{#1}}
\begin{table}
\centering
\caption{
  \tss{full} features ($\archive=\trainset$ for \tss{full}) from feature sets $\featBasic$, $\featCMA$, $\featDisp$, and $\featyDis$.
  Features are grouped according to their feature sets (separated by horizontal lines). Features with less than 25\% of values equal to $\nanout$ and robustness greater than 0.9, are in gray. $N_\nanout$ denotes the lowest measured number of points from which at most 1\% of feature calculations resulted in $\nanout$ ($N_\nanout = 0$ for sample set independent $\feat{}$). The $(\dm_i, \dm_j)$ column shows the pairs of feature dimensions for which the two-sided Wilcoxon signed rank test with the Bonferroni-Holm correction does not reject the hypothesis of equality of median feature values, at the family-wise level 0.05 for each individual feature.}
  \label{tab:featProp_full_1}
\resizebox{\textwidth}{!}{%
\footnotesize
\begin{tabular}{L{25mm}R{5.5mm}R{6.5mm}L{7mm}R{5.5mm}R{6.5mm}L{7mm}R{5.5mm}R{6.5mm}L{7mm}R{5.5mm}R{6.5mm}L{7mm}}
\toprule
{} & \multicolumn{3}{c}{$\archive$} & \multicolumn{3}{c}{$\trans{\archive}$} & \multicolumn{3}{c}{$\archivepred$} & \multicolumn{3}{c}{$\trans{\archivepred}$} \\
\cmidrule(lr){2-4}
\cmidrule(lr){5-7}
\cmidrule(lr){8-10}
\cmidrule(lr){11-13}
                                                                          &                $N_\nanout$ &                        rob.(\%) &            $(\dm_i, \dm_j)$ &                 $N_\nanout$ &                        rob.(\%) &          $(\dm_i, \dm_j)$ &                 $N_\nanout$ &                        rob.(\%) &          $(\dm_i, \dm_j)$ &                 $N_\nanout$ &                        rob.(\%) &          $(\dm_i, \dm_j)$ \\
\midrule
                                                  $\tableFeat{dim}{}{}{}$ & \cellcolor[gray]{0.8500} 0 & \cellcolor[gray]{0.8500} 100.00 &   \cellcolor[gray]{0.8500}  &  \cellcolor[gray]{0.8500} 0 & \cellcolor[gray]{0.8500} 100.00 & \cellcolor[gray]{0.8500}  &  \cellcolor[gray]{0.8500} 0 & \cellcolor[gray]{0.8500} 100.00 & \cellcolor[gray]{0.8500}  &  \cellcolor[gray]{0.8500} 0 & \cellcolor[gray]{0.8500} 100.00 & \cellcolor[gray]{0.8500}  \\
                                                  $\tableFeat{obs}{}{}{}$ & \cellcolor[gray]{0.8500} 1 & \cellcolor[gray]{0.8500} 100.00 &   \cellcolor[gray]{0.8500}  &  \cellcolor[gray]{0.8500} 1 & \cellcolor[gray]{0.8500} 100.00 & \cellcolor[gray]{0.8500}  & \cellcolor[gray]{0.8500} 13 & \cellcolor[gray]{0.8500} 100.00 & \cellcolor[gray]{0.8500}  & \cellcolor[gray]{0.8500} 13 & \cellcolor[gray]{0.8500} 100.00 & \cellcolor[gray]{0.8500}  \\
\midrule
                              $\tableFeat{generation}{}{}{\ftCMA}$ & \cellcolor[gray]{0.8500} 0 & \cellcolor[gray]{0.8500} 100.00 &   \cellcolor[gray]{0.8500}  &  \cellcolor[gray]{0.8500} 0 & \cellcolor[gray]{0.8500} 100.00 & \cellcolor[gray]{0.8500}  &  \cellcolor[gray]{0.8500} 0 & \cellcolor[gray]{0.8500} 100.00 & \cellcolor[gray]{0.8500}  &  \cellcolor[gray]{0.8500} 0 & \cellcolor[gray]{0.8500} 100.00 & \cellcolor[gray]{0.8500}  \\
                              $\tableFeat{step\_size}{}{}{\ftCMA}$ & \cellcolor[gray]{0.8500} 0 & \cellcolor[gray]{0.8500} 100.00 &   \cellcolor[gray]{0.8500}  &  \cellcolor[gray]{0.8500} 0 & \cellcolor[gray]{0.8500} 100.00 & \cellcolor[gray]{0.8500}  &  \cellcolor[gray]{0.8500} 0 & \cellcolor[gray]{0.8500} 100.00 & \cellcolor[gray]{0.8500}  &  \cellcolor[gray]{0.8500} 0 & \cellcolor[gray]{0.8500} 100.00 & \cellcolor[gray]{0.8500}  \\
                                 $\tableFeat{restart}{}{}{\ftCMA}$ & \cellcolor[gray]{0.8500} 0 & \cellcolor[gray]{0.8500} 100.00 &   \cellcolor[gray]{0.8500}  &  \cellcolor[gray]{0.8500} 0 & \cellcolor[gray]{0.8500} 100.00 & \cellcolor[gray]{0.8500}  &  \cellcolor[gray]{0.8500} 0 & \cellcolor[gray]{0.8500} 100.00 & \cellcolor[gray]{0.8500}  &  \cellcolor[gray]{0.8500} 0 & \cellcolor[gray]{0.8500} 100.00 & \cellcolor[gray]{0.8500}  \\
               $\tableFeat{cma\_mean\_dist}{}{\transnone}{\ftCMA}$ &                          6 &                           70.11 &                             &                           6 &                           56.83 &                           &                          13 &                           63.61 &                           &                          13 &                           54.12 &                           \\
                        $\tableFeat{evopath\_c\_norm}{}{}{\ftCMA}$ & \cellcolor[gray]{0.8500} 0 & \cellcolor[gray]{0.8500} 100.00 &   \cellcolor[gray]{0.8500}  &  \cellcolor[gray]{0.8500} 0 & \cellcolor[gray]{0.8500} 100.00 & \cellcolor[gray]{0.8500}  &  \cellcolor[gray]{0.8500} 0 & \cellcolor[gray]{0.8500} 100.00 & \cellcolor[gray]{0.8500}  &  \cellcolor[gray]{0.8500} 0 & \cellcolor[gray]{0.8500} 100.00 & \cellcolor[gray]{0.8500}  \\
                        $\tableFeat{evopath\_s\_norm}{}{}{\ftCMA}$ & \cellcolor[gray]{0.8500} 0 & \cellcolor[gray]{0.8500} 100.00 &   \cellcolor[gray]{0.8500}  &  \cellcolor[gray]{0.8500} 0 & \cellcolor[gray]{0.8500} 100.00 & \cellcolor[gray]{0.8500}  &  \cellcolor[gray]{0.8500} 0 & \cellcolor[gray]{0.8500} 100.00 & \cellcolor[gray]{0.8500}  &  \cellcolor[gray]{0.8500} 0 & \cellcolor[gray]{0.8500} 100.00 & \cellcolor[gray]{0.8500}  \\
                      $\tableFeat{cma\_lik}{}{\transnone}{\ftCMA}$ & \cellcolor[gray]{0.8500} 1 &  \cellcolor[gray]{0.8500} 99.75 &   \cellcolor[gray]{0.8500}  &  \cellcolor[gray]{0.8500} 1 &  \cellcolor[gray]{0.8500} 99.96 & \cellcolor[gray]{0.8500}  & \cellcolor[gray]{0.8500} 13 &  \cellcolor[gray]{0.8500} 99.75 & \cellcolor[gray]{0.8500}  & \cellcolor[gray]{0.8500} 13 &  \cellcolor[gray]{0.8500} 99.96 & \cellcolor[gray]{0.8500}  \\
\midrule
              $\tableFeat{ratio\_mean\_02}{}{\transnone}{\ftDisp}$ &                         71 &                           57.69 & (2,\,5), (3,\,10), (3,\,20) &                          71 &                           62.33 &         (2,\,5), (5,\,20) &                          86 &                           57.36 &         (2,\,5), (3,\,20) &                          86 &                           62.16 &                           \\
            $\tableFeat{ratio\_median\_02}{}{\transnone}{\ftDisp}$ &                         71 &                           65.95 &                     (3,\,5) &                          71 &                           69.28 &                           &                          86 &                           65.41 &                           &                          86 &                           69.60 &                           \\
               $\tableFeat{diff\_mean\_02}{}{\transnone}{\ftDisp}$ &                         71 &                           47.75 &                             & \cellcolor[gray]{0.8500} 71 &  \cellcolor[gray]{0.8500} 99.52 & \cellcolor[gray]{0.8500}  &                          86 &                           48.77 &                           & \cellcolor[gray]{0.8500} 86 &  \cellcolor[gray]{0.8500} 99.52 & \cellcolor[gray]{0.8500}  \\
             $\tableFeat{diff\_median\_02}{}{\transnone}{\ftDisp}$ &                         71 &                           50.21 &                             & \cellcolor[gray]{0.8500} 71 &  \cellcolor[gray]{0.8500} 99.44 & \cellcolor[gray]{0.8500}  &                          86 &                           51.44 &                           & \cellcolor[gray]{0.8500} 86 &  \cellcolor[gray]{0.8500} 99.45 & \cellcolor[gray]{0.8500}  \\
              $\tableFeat{ratio\_mean\_05}{}{\transnone}{\ftDisp}$ &                         27 &                           55.22 &                    (5,\,20) &                          27 &                           60.59 &                           &                          42 &                           54.76 &                  (5,\,20) &                          42 &                           60.32 &                           \\
            $\tableFeat{ratio\_median\_05}{}{\transnone}{\ftDisp}$ &                         27 &                           59.80 &                             &                          27 &                           64.43 &                           &                          42 &                           59.66 &                  (5,\,20) &                          42 &                           64.57 &                           \\
               $\tableFeat{diff\_mean\_05}{}{\transnone}{\ftDisp}$ &                         27 &                           47.99 &                             & \cellcolor[gray]{0.8500} 27 &  \cellcolor[gray]{0.8500} 99.49 & \cellcolor[gray]{0.8500}  &                          42 &                           49.16 &                           & \cellcolor[gray]{0.8500} 42 &  \cellcolor[gray]{0.8500} 99.49 & \cellcolor[gray]{0.8500}  \\
             $\tableFeat{diff\_median\_05}{}{\transnone}{\ftDisp}$ &                         27 &                           51.16 &                             & \cellcolor[gray]{0.8500} 27 &  \cellcolor[gray]{0.8500} 99.46 & \cellcolor[gray]{0.8500}  &                          42 &                           52.88 &                           & \cellcolor[gray]{0.8500} 42 &  \cellcolor[gray]{0.8500} 99.48 & \cellcolor[gray]{0.8500}  \\
              $\tableFeat{ratio\_mean\_10}{}{\transnone}{\ftDisp}$ &                         12 &                           54.63 &                    (5,\,20) &                          12 &                           59.33 &                           &                          25 &                           54.06 &                  (5,\,20) &                          25 &                           58.71 &                           \\
            $\tableFeat{ratio\_median\_10}{}{\transnone}{\ftDisp}$ &                         12 &                           58.29 &                             &                          12 &                           62.38 &                   (2,\,3) &                          25 &                           57.80 &                           &                          25 &                           61.82 &                           \\
               $\tableFeat{diff\_mean\_10}{}{\transnone}{\ftDisp}$ &                         12 &                           49.70 &                             & \cellcolor[gray]{0.8500} 12 &  \cellcolor[gray]{0.8500} 99.52 & \cellcolor[gray]{0.8500}  &                          25 &                           50.88 &                           & \cellcolor[gray]{0.8500} 25 &  \cellcolor[gray]{0.8500} 99.52 & \cellcolor[gray]{0.8500}  \\
             $\tableFeat{diff\_median\_10}{}{\transnone}{\ftDisp}$ &                         12 &                           53.51 &                             & \cellcolor[gray]{0.8500} 12 &  \cellcolor[gray]{0.8500} 99.48 & \cellcolor[gray]{0.8500}  &                          25 &                           55.53 &                           & \cellcolor[gray]{0.8500} 25 &  \cellcolor[gray]{0.8500} 99.49 & \cellcolor[gray]{0.8500}  \\
              $\tableFeat{ratio\_mean\_25}{}{\transnone}{\ftDisp}$ &                          6 &                           54.80 &                             &                           6 &                           58.86 &                           &                          15 &                           53.87 &                           &                          15 &                           58.06 &                           \\
            $\tableFeat{ratio\_median\_25}{}{\transnone}{\ftDisp}$ &                          6 &                           56.21 &                             &                           6 &                           59.74 &                           &                          15 &                           54.91 &                           &                          15 &                           58.26 &                           \\
               $\tableFeat{diff\_mean\_25}{}{\transnone}{\ftDisp}$ &                          6 &                           53.78 &                             &  \cellcolor[gray]{0.8500} 6 &  \cellcolor[gray]{0.8500} 99.55 & \cellcolor[gray]{0.8500}  &                          15 &                           55.29 &                           & \cellcolor[gray]{0.8500} 15 &  \cellcolor[gray]{0.8500} 99.55 & \cellcolor[gray]{0.8500}  \\
             $\tableFeat{diff\_median\_25}{}{\transnone}{\ftDisp}$ &                          6 &                           57.40 &                             &  \cellcolor[gray]{0.8500} 6 &  \cellcolor[gray]{0.8500} 99.49 & \cellcolor[gray]{0.8500}  &                          15 &                           60.24 &                           & \cellcolor[gray]{0.8500} 15 &  \cellcolor[gray]{0.8500} 99.50 & \cellcolor[gray]{0.8500}  \\
\midrule
                     $\tableFeat{skewness}{}{\transnone}{\ftyDis}$ &                          6 &                           36.71 &                             &                         $-$ &                             $-$ &                       $-$ &                         $-$ &                             $-$ &                       $-$ &                         $-$ &                             $-$ &                       $-$ \\
                     $\tableFeat{kurtosis}{}{\transnone}{\ftyDis}$ &                          6 &                           63.06 &                             &                         $-$ &                             $-$ &                       $-$ &                         $-$ &                             $-$ &                       $-$ &                         $-$ &                             $-$ &                       $-$ \\
            $\tableFeat{number\_of\_peaks}{}{\transnone}{\ftyDis}$ &                          6 &                           32.34 &                             &                         $-$ &                             $-$ &                       $-$ &                         $-$ &                             $-$ &                       $-$ &                         $-$ &                             $-$ &                       $-$ \\
\bottomrule
\end{tabular}
\normalsize
}
\end{table}

\begin{table}
\catcode`\-=12 
\centering
\caption{
  \tss{full} features ($\archive=\trainset$ for \tss{full}) from feature sets \featLevel, \featMM, \featInfo, and \featNBC.
  Features are grouped according to their feature sets (separated by horizontal lines). Features with less than 25\% of values equal to $\nanout$ and robustness greater than 0.9, are in gray. $N_\nanout$ denotes the lowest measured number of points from which at most 1\% of feature calculations resulted in $\nanout$ ($N_\nanout = 0$ for sample set independent $\feat{}$). The $(\dm_i, \dm_j)$ column shows the pairs of feature dimensions for which the two-sided Wilcoxon signed rank test with the Bonferroni-Holm correction does not reject the hypothesis of equality of median feature values, at the family-wise level 0.05 for each individual feature.}
\label{tab:featProp_full_2}
\resizebox{\textwidth}{!}{%
\footnotesize
\begin{tabular}{L{25mm}R{5.5mm}R{6.5mm}L{7mm}R{5.5mm}R{6.5mm}L{7mm}R{5.5mm}R{6.5mm}L{7mm}R{5.5mm}R{6.5mm}L{7mm}}
\toprule
{} & \multicolumn{3}{c}{$\archive$} & \multicolumn{3}{c}{$\trans{\archive}$} & \multicolumn{3}{c}{$\archivepred$} & \multicolumn{3}{c}{$\trans{\archivepred}$} \\
\cmidrule(lr){2-4}
\cmidrule(lr){5-7}
\cmidrule(lr){8-10}
\cmidrule(lr){11-13}
                                                                          &                $N_\nanout$ &                        rob.(\%) &            $(\dm_i, \dm_j)$ &                 $N_\nanout$ &                        rob.(\%) &          $(\dm_i, \dm_j)$ &                 $N_\nanout$ &                        rob.(\%) &          $(\dm_i, \dm_j)$ &                 $N_\nanout$ &                        rob.(\%) &          $(\dm_i, \dm_j)$ \\
\midrule
               $\tableFeat{mmce\_lda\_10}{}{\transnone}{\ftLevel}$ &                          6 &                           39.47 &                             &                           6 &                           39.47 &                           &                          13 &                           53.01 &                           &                          13 &                           53.01 &                           \\
               $\tableFeat{mmce\_qda\_10}{}{\transnone}{\ftLevel}$ &                          6 &                           69.30 &                             &                           6 &                           69.88 &                           &                          13 &                           65.30 &                           &                          13 &                           65.91 &                           \\
               $\tableFeat{mmce\_mda\_10}{}{\transnone}{\ftLevel}$ &                          6 &                           28.07 &                             &                           6 &                           28.14 &                           &                          13 &                           27.38 &                           &                          13 &                           27.64 &                           \\
                $\tableFeat{lda\_qda\_10}{}{\transnone}{\ftLevel}$ &                          6 &                           80.09 &                             &                           6 &                           80.12 &                           & \cellcolor[gray]{0.8500} 18 &  \cellcolor[gray]{0.8500} 92.37 & \cellcolor[gray]{0.8500}  & \cellcolor[gray]{0.8500} 18 &  \cellcolor[gray]{0.8500} 92.38 & \cellcolor[gray]{0.8500}  \\
                $\tableFeat{lda\_mda\_10}{}{\transnone}{\ftLevel}$ &                          6 &                           44.32 &                             &                           6 &                           42.71 &                           &                          13 &                           46.42 &                           &                          13 &                           48.00 &                           \\
                $\tableFeat{qda\_mda\_10}{}{\transnone}{\ftLevel}$ &                          6 &                           81.14 &                             &                           6 &                           80.51 &                           &                          13 &                           78.27 &                           &                          13 &                           77.33 &                           \\
               $\tableFeat{mmce\_lda\_25}{}{\transnone}{\ftLevel}$ &                          6 &                           32.64 &                             &                           6 &                           32.64 &                           &                          13 &                           54.19 &                           &                          13 &                           54.19 &                           \\
               $\tableFeat{mmce\_qda\_25}{}{\transnone}{\ftLevel}$ &                          6 &                           62.01 &                             &                           6 &                           62.41 &                           &                          13 &                           56.49 &                           &                          13 &                           56.69 &                           \\
               $\tableFeat{mmce\_mda\_25}{}{\transnone}{\ftLevel}$ &                          6 &                           22.86 &                             &                           6 &                           22.63 &                           &                          13 &                           27.03 &                           &                          13 &                           26.99 &                           \\
                $\tableFeat{lda\_qda\_25}{}{\transnone}{\ftLevel}$ &                          6 &                           76.25 &                             &                           6 &                           76.29 &                   (3,\,5) & \cellcolor[gray]{0.8500} 15 &  \cellcolor[gray]{0.8500} 94.13 & \cellcolor[gray]{0.8500}  & \cellcolor[gray]{0.8500} 15 &  \cellcolor[gray]{0.8500} 94.12 & \cellcolor[gray]{0.8500}  \\
                $\tableFeat{lda\_mda\_25}{}{\transnone}{\ftLevel}$ &                          6 &                           32.07 &                    (3,\,20) &                           6 &                           27.68 &       (3,\,10), (10,\,20) &                          13 &                           22.86 &                           &                          13 &                           20.39 &                           \\
                $\tableFeat{qda\_mda\_25}{}{\transnone}{\ftLevel}$ &                          6 &                           77.24 &                             &                           6 &                           77.68 &                           &                          13 &                           63.08 &                           &                          13 &                           63.11 &                           \\
               $\tableFeat{mmce\_lda\_50}{}{\transnone}{\ftLevel}$ &                          6 &                           22.40 &                             &                           6 &                           22.39 &                           &                          13 &                           18.46 &                           &                          13 &                           18.46 &                           \\
               $\tableFeat{mmce\_qda\_50}{}{\transnone}{\ftLevel}$ &                          6 &                           48.24 &                             &                           6 &                           49.46 &                           &                          13 &                           32.88 &                           &                          13 &                           33.15 &                           \\
               $\tableFeat{mmce\_mda\_50}{}{\transnone}{\ftLevel}$ &                          6 &                           20.13 &                             &                           6 &                           19.60 &                           &                          13 &                           38.52 &                           &                          13 &                           37.58 &                           \\
                $\tableFeat{lda\_qda\_50}{}{\transnone}{\ftLevel}$ &                          6 &                           69.22 &                   (10,\,20) &                           6 &                           69.24 &                 (10,\,20) &                          15 &                           84.45 &                           &                          15 &                           84.41 &                           \\
                $\tableFeat{lda\_mda\_50}{}{\transnone}{\ftLevel}$ &                          6 &                           36.05 &                             &                           6 &                           30.90 &                           &                          13 &                            6.80 &                  (5,\,20) &                          13 &                            4.51 &                           \\
                $\tableFeat{qda\_mda\_50}{}{\transnone}{\ftLevel}$ &                          6 &                           42.65 &                    (3,\,20) &                           6 &                           40.68 &                           &                          13 &                           21.96 &                  (5,\,20) &                          13 &                           22.47 &         (3,\,5), (5,\,20) \\
\midrule
           $\tableFeat{lin\_simple\_adj\_r2}{}{\transnone}{\ftMM}$ &                          6 &                           19.64 &                             &                           6 &                           19.62 &                           &                         $-$ &                             $-$ &                       $-$ &                         $-$ &                             $-$ &                       $-$ \\
         $\tableFeat{lin\_simple\_coef\_min}{}{\transnone}{\ftMM}$ &                          6 &                           51.95 &                             &                           6 &                           85.41 &                           &                         $-$ &                             $-$ &                       $-$ &                         $-$ &                             $-$ &                       $-$ \\
         $\tableFeat{lin\_simple\_coef\_max}{}{\transnone}{\ftMM}$ &                          6 &                           29.73 &                             &                           6 &                           77.68 &                           &                         $-$ &                             $-$ &                       $-$ &                         $-$ &                             $-$ &                       $-$ \\
$\tableFeat{lin\_simple\_coef\_max\_by\_min}{}{\transnone}{\ftMM}$ &                          6 &                           27.78 &                             &                           6 &                           69.78 &                           &                         $-$ &                             $-$ &                       $-$ &                         $-$ &                             $-$ &                       $-$ \\
      $\tableFeat{lin\_w\_interact\_adj\_r2}{}{\transnone}{\ftMM}$ &                        100 &                           44.77 &                             &                         100 &                           43.89 &                           &                         $-$ &                             $-$ &                       $-$ &                         $-$ &                             $-$ &                       $-$ \\
          $\tableFeat{quad\_simple\_adj\_r2}{}{\transnone}{\ftMM}$ &                          6 &                           50.23 &                             &                           6 &                           52.16 &                           &                         $-$ &                             $-$ &                       $-$ &                         $-$ &                             $-$ &                       $-$ \\
             $\tableFeat{quad\_simple\_cond}{}{\transnone}{\ftMM}$ &                          6 &                           46.45 &                             &  \cellcolor[gray]{0.8500} 6 &  \cellcolor[gray]{0.8500} 95.21 & \cellcolor[gray]{0.8500}  &                         $-$ &                             $-$ &                       $-$ &                         $-$ &                             $-$ &                       $-$ \\
     $\tableFeat{quad\_w\_interact\_adj\_r2}{}{\transnone}{\ftMM}$ &                        629 &                           82.50 &                             &                         531 &                           77.81 &                           &                         $-$ &                             $-$ &                       $-$ &                         $-$ &                             $-$ &                       $-$ \\
\midrule
                       $\tableFeat{h\_max}{}{\transnone}{\ftInfo}$ &                          6 &                           14.46 &                     (3,\,5) &                           6 &                           33.83 &                           &                          13 &                           35.74 &                           &                          13 &                           35.42 &                           \\
                       $\tableFeat{eps\_s}{}{\transnone}{\ftInfo}$ &                          6 &                            6.59 &                             &                           6 &                           34.65 &                           &                        3056 &                           26.05 &                           &                        3196 &                           54.56 &                           \\
                     $\tableFeat{eps\_max}{}{\transnone}{\ftInfo}$ &                          6 &                           64.93 &                             &                           6 &                           84.99 &                           &                          13 &                           65.17 &                           &                          13 &                           84.18 &                           \\
                           $\tableFeat{m0}{}{\transnone}{\ftInfo}$ &                          6 &                            1.30 &                             &                           6 &                            3.88 &                           &                         $-$ &                             $-$ &                       $-$ &                         $-$ &                             $-$ &                       $-$ \\
                   $\tableFeat{eps\_ratio}{}{\transnone}{\ftInfo}$ &                          6 &                           27.67 &                             &                           6 &                           44.91 &                           &                         $-$ &                             $-$ &                       $-$ &                         $-$ &                             $-$ &                       $-$ \\
\midrule
                $\tableFeat{nb\_std\_ratio}{}{\transnone}{\ftNBC}$ &                          6 &                           31.11 &                             &                           6 &                           18.27 &                           &                         $-$ &                             $-$ &                       $-$ &                         $-$ &                             $-$ &                       $-$ \\
               $\tableFeat{nb\_mean\_ratio}{}{\transnone}{\ftNBC}$ &                          6 &                           38.98 &                             &                           6 &                           32.96 &                           &                         $-$ &                             $-$ &                       $-$ &                         $-$ &                             $-$ &                       $-$ \\
                       $\tableFeat{nb\_cor}{}{\transnone}{\ftNBC}$ &                          6 &                           45.94 &                             &                           6 &                           32.39 &                           &                         $-$ &                             $-$ &                       $-$ &                         $-$ &                             $-$ &                       $-$ \\
                   $\tableFeat{dist\_ratio}{}{\transnone}{\ftNBC}$ &                          6 &                           55.23 &                             &                           6 &                           50.78 &                           &                         $-$ &                             $-$ &                       $-$ &                         $-$ &                             $-$ &                       $-$ \\
              $\tableFeat{nb\_fitness\_cor}{}{\transnone}{\ftNBC}$ &                          6 &                           77.17 &                             &                           6 &                           77.13 &                           &                         $-$ &                             $-$ &                       $-$ &                         $-$ &                             $-$ &                       $-$ \\
\bottomrule
\end{tabular}
\normalsize
}
\end{table}

%% file: tex/feat_nearest_1.tex
\newcolumntype{C}[1]{>{\PreserveBackslash\centering}p{#1}}
\newcolumntype{R}[1]{>{\PreserveBackslash\raggedleft}p{#1}}
\newcolumntype{L}[1]{>{\PreserveBackslash\raggedright}p{#1}}
\begin{table}
\centering
\caption{
  \tss{nearest} features from feature sets \featBasic, \featCMA, \featDisp, and \featyDis{}
  (only $\trainset$-based and sample set indepent, $\archive$-based are identical to \tss{full} in Table~\ref{tab:featProp_full_1}). Features are grouped according to their feature sets (separated by horizontal lines). Features with less than 25\% of values equal to $\nanout$ and robustness greater than 0.9, are in gray. $N_\nanout$ denotes the lowest measured number of points from which at most 1\% of feature calculations resulted in $\nanout$ ($N_\nanout = 0$ for sample set independent $\feat{}$). The $(\dm_i, \dm_j)$ column shows the pairs of feature dimensions for which the two-sided Wilcoxon signed rank test with the Bonferroni-Holm correction does not reject the hypothesis of equality of median feature values, at the family-wise level 0.05 for each individual feature.}
\label{tab:featProp_nearest_1}
\resizebox{\textwidth}{!}{%
\footnotesize
\begin{tabular}{L{25mm}R{5.5mm}R{6.5mm}L{7mm}R{5.5mm}R{6.5mm}L{7mm}R{5.5mm}R{6.5mm}L{7mm}R{5.5mm}R{6.5mm}L{7mm}}
\toprule
{} & \multicolumn{3}{c}{$\trainset$} & \multicolumn{3}{c}{$\trans{\trainset}$} & \multicolumn{3}{c}{$\trainpredset$} & \multicolumn{3}{c}{$\trans{\trainpredset}$} \\
\cmidrule(lr){2-4}
\cmidrule(lr){5-7}
\cmidrule(lr){8-10}
\cmidrule(lr){11-13}
                                                                          &                $N_\nanout$ &                        rob.(\%) &            $(\dm_i, \dm_j)$ &                 $N_\nanout$ &                        rob.(\%) &          $(\dm_i, \dm_j)$ &                 $N_\nanout$ &                        rob.(\%) &          $(\dm_i, \dm_j)$ &                 $N_\nanout$ &                        rob.(\%) &          $(\dm_i, \dm_j)$ \\
\midrule
                                                  $\tableFeat{dim}{}{}{}$ & \cellcolor[gray]{0.8500} 0 & \cellcolor[gray]{0.8500} 100.00 &   \cellcolor[gray]{0.8500}  &  \cellcolor[gray]{0.8500} 0 & \cellcolor[gray]{0.8500} 100.00 & \cellcolor[gray]{0.8500}  &  \cellcolor[gray]{0.8500} 0 & \cellcolor[gray]{0.8500} 100.00 & \cellcolor[gray]{0.8500}  &  \cellcolor[gray]{0.8500} 0 & \cellcolor[gray]{0.8500} 100.00 & \cellcolor[gray]{0.8500}  \\
                                                  $\tableFeat{obs}{}{}{}$ & \cellcolor[gray]{0.8500} 1 & \cellcolor[gray]{0.8500} 100.00 &   \cellcolor[gray]{0.8500}  &  \cellcolor[gray]{0.8500} 1 & \cellcolor[gray]{0.8500} 100.00 & \cellcolor[gray]{0.8500}  & \cellcolor[gray]{0.8500} 13 & \cellcolor[gray]{0.8500} 100.00 & \cellcolor[gray]{0.8500}  & \cellcolor[gray]{0.8500} 13 & \cellcolor[gray]{0.8500} 100.00 & \cellcolor[gray]{0.8500}  \\
\midrule
                              $\tableFeat{generation}{}{}{\ftCMA}$ & \cellcolor[gray]{0.8500} 0 & \cellcolor[gray]{0.8500} 100.00 &   \cellcolor[gray]{0.8500}  &  \cellcolor[gray]{0.8500} 0 & \cellcolor[gray]{0.8500} 100.00 & \cellcolor[gray]{0.8500}  &  \cellcolor[gray]{0.8500} 0 & \cellcolor[gray]{0.8500} 100.00 & \cellcolor[gray]{0.8500}  &  \cellcolor[gray]{0.8500} 0 & \cellcolor[gray]{0.8500} 100.00 & \cellcolor[gray]{0.8500}  \\
                              $\tableFeat{step\_size}{}{}{\ftCMA}$ & \cellcolor[gray]{0.8500} 0 & \cellcolor[gray]{0.8500} 100.00 &   \cellcolor[gray]{0.8500}  &  \cellcolor[gray]{0.8500} 0 & \cellcolor[gray]{0.8500} 100.00 & \cellcolor[gray]{0.8500}  &  \cellcolor[gray]{0.8500} 0 & \cellcolor[gray]{0.8500} 100.00 & \cellcolor[gray]{0.8500}  &  \cellcolor[gray]{0.8500} 0 & \cellcolor[gray]{0.8500} 100.00 & \cellcolor[gray]{0.8500}  \\
                                 $\tableFeat{restart}{}{}{\ftCMA}$ & \cellcolor[gray]{0.8500} 0 & \cellcolor[gray]{0.8500} 100.00 &   \cellcolor[gray]{0.8500}  &  \cellcolor[gray]{0.8500} 0 & \cellcolor[gray]{0.8500} 100.00 & \cellcolor[gray]{0.8500}  &  \cellcolor[gray]{0.8500} 0 & \cellcolor[gray]{0.8500} 100.00 & \cellcolor[gray]{0.8500}  &  \cellcolor[gray]{0.8500} 0 & \cellcolor[gray]{0.8500} 100.00 & \cellcolor[gray]{0.8500}  \\
               $\tableFeat{cma\_mean\_dist}{}{\transnone}{\ftCMA}$ &                          6 &                           70.11 &                             &                           6 &                           56.83 &                           &                          13 &                           63.61 &                           &                          13 &                           54.12 &                           \\
                        $\tableFeat{evopath\_c\_norm}{}{}{\ftCMA}$ & \cellcolor[gray]{0.8500} 0 & \cellcolor[gray]{0.8500} 100.00 &   \cellcolor[gray]{0.8500}  &  \cellcolor[gray]{0.8500} 0 & \cellcolor[gray]{0.8500} 100.00 & \cellcolor[gray]{0.8500}  &  \cellcolor[gray]{0.8500} 0 & \cellcolor[gray]{0.8500} 100.00 & \cellcolor[gray]{0.8500}  &  \cellcolor[gray]{0.8500} 0 & \cellcolor[gray]{0.8500} 100.00 & \cellcolor[gray]{0.8500}  \\
                        $\tableFeat{evopath\_s\_norm}{}{}{\ftCMA}$ & \cellcolor[gray]{0.8500} 0 & \cellcolor[gray]{0.8500} 100.00 &   \cellcolor[gray]{0.8500}  &  \cellcolor[gray]{0.8500} 0 & \cellcolor[gray]{0.8500} 100.00 & \cellcolor[gray]{0.8500}  &  \cellcolor[gray]{0.8500} 0 & \cellcolor[gray]{0.8500} 100.00 & \cellcolor[gray]{0.8500}  &  \cellcolor[gray]{0.8500} 0 & \cellcolor[gray]{0.8500} 100.00 & \cellcolor[gray]{0.8500}  \\
                      $\tableFeat{cma\_lik}{}{\transnone}{\ftCMA}$ & \cellcolor[gray]{0.8500} 1 &  \cellcolor[gray]{0.8500} 99.75 &   \cellcolor[gray]{0.8500}  &  \cellcolor[gray]{0.8500} 1 &  \cellcolor[gray]{0.8500} 99.96 & \cellcolor[gray]{0.8500}  & \cellcolor[gray]{0.8500} 13 &  \cellcolor[gray]{0.8500} 99.75 & \cellcolor[gray]{0.8500}  & \cellcolor[gray]{0.8500} 13 &  \cellcolor[gray]{0.8500} 99.96 & \cellcolor[gray]{0.8500}  \\
\midrule
              $\tableFeat{ratio\_mean\_02}{}{\transnone}{\ftDisp}$ &                         71 &                           57.69 & (2,\,5), (3,\,10), (3,\,20) &                          71 &                           62.33 &         (2,\,5), (5,\,20) &                          86 &                           57.36 &         (2,\,5), (3,\,20) &                          86 &                           62.16 &                           \\
            $\tableFeat{ratio\_median\_02}{}{\transnone}{\ftDisp}$ &                         71 &                           65.95 &                     (3,\,5) &                          71 &                           69.28 &                           &                          86 &                           65.41 &                           &                          86 &                           69.60 &                           \\
               $\tableFeat{diff\_mean\_02}{}{\transnone}{\ftDisp}$ &                         71 &                           47.75 &                             & \cellcolor[gray]{0.8500} 71 &  \cellcolor[gray]{0.8500} 99.52 & \cellcolor[gray]{0.8500}  &                          86 &                           48.77 &                           & \cellcolor[gray]{0.8500} 86 &  \cellcolor[gray]{0.8500} 99.52 & \cellcolor[gray]{0.8500}  \\
             $\tableFeat{diff\_median\_02}{}{\transnone}{\ftDisp}$ &                         71 &                           50.21 &                             & \cellcolor[gray]{0.8500} 71 &  \cellcolor[gray]{0.8500} 99.44 & \cellcolor[gray]{0.8500}  &                          86 &                           51.44 &                           & \cellcolor[gray]{0.8500} 86 &  \cellcolor[gray]{0.8500} 99.45 & \cellcolor[gray]{0.8500}  \\
              $\tableFeat{ratio\_mean\_05}{}{\transnone}{\ftDisp}$ &                         27 &                           55.22 &                    (5,\,20) &                          27 &                           60.59 &                           &                          42 &                           54.76 &                  (5,\,20) &                          42 &                           60.32 &                           \\
            $\tableFeat{ratio\_median\_05}{}{\transnone}{\ftDisp}$ &                         27 &                           59.80 &                             &                          27 &                           64.43 &                           &                          42 &                           59.66 &                  (5,\,20) &                          42 &                           64.57 &                           \\
               $\tableFeat{diff\_mean\_05}{}{\transnone}{\ftDisp}$ &                         27 &                           47.99 &                             & \cellcolor[gray]{0.8500} 27 &  \cellcolor[gray]{0.8500} 99.49 & \cellcolor[gray]{0.8500}  &                          42 &                           49.16 &                           & \cellcolor[gray]{0.8500} 42 &  \cellcolor[gray]{0.8500} 99.49 & \cellcolor[gray]{0.8500}  \\
             $\tableFeat{diff\_median\_05}{}{\transnone}{\ftDisp}$ &                         27 &                           51.16 &                             & \cellcolor[gray]{0.8500} 27 &  \cellcolor[gray]{0.8500} 99.46 & \cellcolor[gray]{0.8500}  &                          42 &                           52.88 &                           & \cellcolor[gray]{0.8500} 42 &  \cellcolor[gray]{0.8500} 99.48 & \cellcolor[gray]{0.8500}  \\
              $\tableFeat{ratio\_mean\_10}{}{\transnone}{\ftDisp}$ &                         12 &                           54.63 &                    (5,\,20) &                          12 &                           59.33 &                           &                          25 &                           54.06 &                  (5,\,20) &                          25 &                           58.71 &                           \\
            $\tableFeat{ratio\_median\_10}{}{\transnone}{\ftDisp}$ &                         12 &                           58.29 &                             &                          12 &                           62.38 &                   (2,\,3) &                          25 &                           57.80 &                           &                          25 &                           61.82 &                           \\
               $\tableFeat{diff\_mean\_10}{}{\transnone}{\ftDisp}$ &                         12 &                           49.70 &                             & \cellcolor[gray]{0.8500} 12 &  \cellcolor[gray]{0.8500} 99.52 & \cellcolor[gray]{0.8500}  &                          25 &                           50.88 &                           & \cellcolor[gray]{0.8500} 25 &  \cellcolor[gray]{0.8500} 99.52 & \cellcolor[gray]{0.8500}  \\
             $\tableFeat{diff\_median\_10}{}{\transnone}{\ftDisp}$ &                         12 &                           53.51 &                             & \cellcolor[gray]{0.8500} 12 &  \cellcolor[gray]{0.8500} 99.48 & \cellcolor[gray]{0.8500}  &                          25 &                           55.53 &                           & \cellcolor[gray]{0.8500} 25 &  \cellcolor[gray]{0.8500} 99.49 & \cellcolor[gray]{0.8500}  \\
              $\tableFeat{ratio\_mean\_25}{}{\transnone}{\ftDisp}$ &                          6 &                           54.80 &                             &                           6 &                           58.86 &                           &                          15 &                           53.87 &                           &                          15 &                           58.06 &                           \\
            $\tableFeat{ratio\_median\_25}{}{\transnone}{\ftDisp}$ &                          6 &                           56.21 &                             &                           6 &                           59.74 &                           &                          15 &                           54.91 &                           &                          15 &                           58.26 &                           \\
               $\tableFeat{diff\_mean\_25}{}{\transnone}{\ftDisp}$ &                          6 &                           53.78 &                             &  \cellcolor[gray]{0.8500} 6 &  \cellcolor[gray]{0.8500} 99.55 & \cellcolor[gray]{0.8500}  &                          15 &                           55.29 &                           & \cellcolor[gray]{0.8500} 15 &  \cellcolor[gray]{0.8500} 99.55 & \cellcolor[gray]{0.8500}  \\
             $\tableFeat{diff\_median\_25}{}{\transnone}{\ftDisp}$ &                          6 &                           57.40 &                             &  \cellcolor[gray]{0.8500} 6 &  \cellcolor[gray]{0.8500} 99.49 & \cellcolor[gray]{0.8500}  &                          15 &                           60.24 &                           & \cellcolor[gray]{0.8500} 15 &  \cellcolor[gray]{0.8500} 99.50 & \cellcolor[gray]{0.8500}  \\
\midrule
                     $\tableFeat{skewness}{}{\transnone}{\ftyDis}$ &                          6 &                           36.71 &                             &                         $-$ &                             $-$ &                       $-$ &                         $-$ &                             $-$ &                       $-$ &                         $-$ &                             $-$ &                       $-$ \\
                     $\tableFeat{kurtosis}{}{\transnone}{\ftyDis}$ &                          6 &                           63.06 &                             &                         $-$ &                             $-$ &                       $-$ &                         $-$ &                             $-$ &                       $-$ &                         $-$ &                             $-$ &                       $-$ \\
            $\tableFeat{number\_of\_peaks}{}{\transnone}{\ftyDis}$ &                          6 &                           32.34 &                             &                         $-$ &                             $-$ &                       $-$ &                         $-$ &                             $-$ &                       $-$ &                         $-$ &                             $-$ &                       $-$ \\
\bottomrule
\end{tabular}
\normalsize
}
\end{table}
            
\begin{table}
\catcode`\-=12 
\centering
\caption{
  \tss{nearest} features from feature sets \featLevel, \featMM, \featInfo, and \featNBC{}
  (only $\trainset$-based and sample set indepent, $\archive$-based are identical to \tss{full} in Table~\ref{tab:featProp_full_2}). Features are grouped according to their feature sets (separated by horizontal lines). Features with less than 25\% of values equal to $\nanout$ and robustness greater than 0.9, are in gray. $N_\nanout$ denotes the lowest measured number of points from which at most 1\% of feature calculations resulted in $\nanout$ ($N_\nanout = 0$ for sample set independent $\feat{}$). The $(\dm_i, \dm_j)$ column shows the pairs of feature dimensions for which the two-sided Wilcoxon signed rank test with the Bonferroni-Holm correction does not reject the hypothesis of equality of median feature values, at the family-wise level 0.05 for each individual feature.}
\label{tab:featProp_nearest_2}
\resizebox{\textwidth}{!}{%
\footnotesize
\begin{tabular}{L{25mm}R{5.5mm}R{6.5mm}L{7mm}R{5.5mm}R{6.5mm}L{7mm}R{5.5mm}R{6.5mm}L{7mm}R{5.5mm}R{6.5mm}L{7mm}}
\toprule
{} & \multicolumn{3}{c}{$\trainset$} & \multicolumn{3}{c}{$\trans{\trainset}$} & \multicolumn{3}{c}{$\trainpredset$} & \multicolumn{3}{c}{$\trans{\trainpredset}$} \\
\cmidrule(lr){2-4}
\cmidrule(lr){5-7}
\cmidrule(lr){8-10}
\cmidrule(lr){11-13}
                                                                          &                $N_\nanout$ &                        rob.(\%) &            $(\dm_i, \dm_j)$ &                 $N_\nanout$ &                        rob.(\%) &          $(\dm_i, \dm_j)$ &                 $N_\nanout$ &                        rob.(\%) &          $(\dm_i, \dm_j)$ &                 $N_\nanout$ &                        rob.(\%) &          $(\dm_i, \dm_j)$ \\
\midrule
               $\tableFeat{mmce\_lda\_10}{}{\transnone}{\ftLevel}$ &                          6 &                           39.47 &                             &                           6 &                           39.47 &                           &                          13 &                           53.01 &                           &                          13 &                           53.01 &                           \\
               $\tableFeat{mmce\_qda\_10}{}{\transnone}{\ftLevel}$ &                          6 &                           69.30 &                             &                           6 &                           69.88 &                           &                          13 &                           65.30 &                           &                          13 &                           65.91 &                           \\
               $\tableFeat{mmce\_mda\_10}{}{\transnone}{\ftLevel}$ &                          6 &                           28.07 &                             &                           6 &                           28.14 &                           &                          13 &                           27.38 &                           &                          13 &                           27.64 &                           \\
                $\tableFeat{lda\_qda\_10}{}{\transnone}{\ftLevel}$ &                          6 &                           80.09 &                             &                           6 &                           80.12 &                           & \cellcolor[gray]{0.8500} 18 &  \cellcolor[gray]{0.8500} 92.37 & \cellcolor[gray]{0.8500}  & \cellcolor[gray]{0.8500} 18 &  \cellcolor[gray]{0.8500} 92.38 & \cellcolor[gray]{0.8500}  \\
                $\tableFeat{lda\_mda\_10}{}{\transnone}{\ftLevel}$ &                          6 &                           44.32 &                             &                           6 &                           42.71 &                           &                          13 &                           46.42 &                           &                          13 &                           48.00 &                           \\
                $\tableFeat{qda\_mda\_10}{}{\transnone}{\ftLevel}$ &                          6 &                           81.14 &                             &                           6 &                           80.51 &                           &                          13 &                           78.27 &                           &                          13 &                           77.33 &                           \\
               $\tableFeat{mmce\_lda\_25}{}{\transnone}{\ftLevel}$ &                          6 &                           32.64 &                             &                           6 &                           32.64 &                           &                          13 &                           54.19 &                           &                          13 &                           54.19 &                           \\
               $\tableFeat{mmce\_qda\_25}{}{\transnone}{\ftLevel}$ &                          6 &                           62.01 &                             &                           6 &                           62.41 &                           &                          13 &                           56.49 &                           &                          13 &                           56.69 &                           \\
               $\tableFeat{mmce\_mda\_25}{}{\transnone}{\ftLevel}$ &                          6 &                           22.86 &                             &                           6 &                           22.63 &                           &                          13 &                           27.03 &                           &                          13 &                           26.99 &                           \\
                $\tableFeat{lda\_qda\_25}{}{\transnone}{\ftLevel}$ &                          6 &                           76.25 &                             &                           6 &                           76.29 &                   (3,\,5) & \cellcolor[gray]{0.8500} 15 &  \cellcolor[gray]{0.8500} 94.13 & \cellcolor[gray]{0.8500}  & \cellcolor[gray]{0.8500} 15 &  \cellcolor[gray]{0.8500} 94.12 & \cellcolor[gray]{0.8500}  \\
                $\tableFeat{lda\_mda\_25}{}{\transnone}{\ftLevel}$ &                          6 &                           32.07 &                    (3,\,20) &                           6 &                           27.68 &       (3,\,10), (10,\,20) &                          13 &                           22.86 &                           &                          13 &                           20.39 &                           \\
                $\tableFeat{qda\_mda\_25}{}{\transnone}{\ftLevel}$ &                          6 &                           77.24 &                             &                           6 &                           77.68 &                           &                          13 &                           63.08 &                           &                          13 &                           63.11 &                           \\
               $\tableFeat{mmce\_lda\_50}{}{\transnone}{\ftLevel}$ &                          6 &                           22.40 &                             &                           6 &                           22.39 &                           &                          13 &                           18.46 &                           &                          13 &                           18.46 &                           \\
               $\tableFeat{mmce\_qda\_50}{}{\transnone}{\ftLevel}$ &                          6 &                           48.24 &                             &                           6 &                           49.46 &                           &                          13 &                           32.88 &                           &                          13 &                           33.15 &                           \\
               $\tableFeat{mmce\_mda\_50}{}{\transnone}{\ftLevel}$ &                          6 &                           20.13 &                             &                           6 &                           19.60 &                           &                          13 &                           38.52 &                           &                          13 &                           37.58 &                           \\
                $\tableFeat{lda\_qda\_50}{}{\transnone}{\ftLevel}$ &                          6 &                           69.22 &                   (10,\,20) &                           6 &                           69.24 &                 (10,\,20) &                          15 &                           84.45 &                           &                          15 &                           84.41 &                           \\
                $\tableFeat{lda\_mda\_50}{}{\transnone}{\ftLevel}$ &                          6 &                           36.05 &                             &                           6 &                           30.90 &                           &                          13 &                            6.80 &                  (5,\,20) &                          13 &                            4.51 &                           \\
                $\tableFeat{qda\_mda\_50}{}{\transnone}{\ftLevel}$ &                          6 &                           42.65 &                    (3,\,20) &                           6 &                           40.68 &                           &                          13 &                           21.96 &                  (5,\,20) &                          13 &                           22.47 &         (3,\,5), (5,\,20) \\
\midrule
           $\tableFeat{lin\_simple\_adj\_r2}{}{\transnone}{\ftMM}$ &                          6 &                           19.64 &                             &                           6 &                           19.62 &                           &                         $-$ &                             $-$ &                       $-$ &                         $-$ &                             $-$ &                       $-$ \\
         $\tableFeat{lin\_simple\_coef\_min}{}{\transnone}{\ftMM}$ &                          6 &                           51.95 &                             &                           6 &                           85.41 &                           &                         $-$ &                             $-$ &                       $-$ &                         $-$ &                             $-$ &                       $-$ \\
         $\tableFeat{lin\_simple\_coef\_max}{}{\transnone}{\ftMM}$ &                          6 &                           29.73 &                             &                           6 &                           77.68 &                           &                         $-$ &                             $-$ &                       $-$ &                         $-$ &                             $-$ &                       $-$ \\
$\tableFeat{lin\_simple\_coef\_max\_by\_min}{}{\transnone}{\ftMM}$ &                          6 &                           27.78 &                             &                           6 &                           69.78 &                           &                         $-$ &                             $-$ &                       $-$ &                         $-$ &                             $-$ &                       $-$ \\
      $\tableFeat{lin\_w\_interact\_adj\_r2}{}{\transnone}{\ftMM}$ &                        100 &                           44.77 &                             &                         100 &                           43.89 &                           &                         $-$ &                             $-$ &                       $-$ &                         $-$ &                             $-$ &                       $-$ \\
          $\tableFeat{quad\_simple\_adj\_r2}{}{\transnone}{\ftMM}$ &                          6 &                           50.23 &                             &                           6 &                           52.16 &                           &                         $-$ &                             $-$ &                       $-$ &                         $-$ &                             $-$ &                       $-$ \\
             $\tableFeat{quad\_simple\_cond}{}{\transnone}{\ftMM}$ &                          6 &                           46.45 &                             &  \cellcolor[gray]{0.8500} 6 &  \cellcolor[gray]{0.8500} 95.21 & \cellcolor[gray]{0.8500}  &                         $-$ &                             $-$ &                       $-$ &                         $-$ &                             $-$ &                       $-$ \\
     $\tableFeat{quad\_w\_interact\_adj\_r2}{}{\transnone}{\ftMM}$ &                        629 &                           82.50 &                             &                         531 &                           77.81 &                           &                         $-$ &                             $-$ &                       $-$ &                         $-$ &                             $-$ &                       $-$ \\
\midrule
                       $\tableFeat{h\_max}{}{\transnone}{\ftInfo}$ &                          6 &                           14.46 &                     (3,\,5) &                           6 &                           33.83 &                           &                          13 &                           35.74 &                           &                          13 &                           35.42 &                           \\
                       $\tableFeat{eps\_s}{}{\transnone}{\ftInfo}$ &                          6 &                            6.59 &                             &                           6 &                           34.65 &                           &                        3056 &                           26.05 &                           &                        3196 &                           54.56 &                           \\
                     $\tableFeat{eps\_max}{}{\transnone}{\ftInfo}$ &                          6 &                           64.93 &                             &                           6 &                           84.99 &                           &                          13 &                           65.17 &                           &                          13 &                           84.18 &                           \\
                           $\tableFeat{m0}{}{\transnone}{\ftInfo}$ &                          6 &                            1.30 &                             &                           6 &                            3.88 &                           &                         $-$ &                             $-$ &                       $-$ &                         $-$ &                             $-$ &                       $-$ \\
                   $\tableFeat{eps\_ratio}{}{\transnone}{\ftInfo}$ &                          6 &                           27.67 &                             &                           6 &                           44.91 &                           &                         $-$ &                             $-$ &                       $-$ &                         $-$ &                             $-$ &                       $-$ \\
\midrule
                $\tableFeat{nb\_std\_ratio}{}{\transnone}{\ftNBC}$ &                          6 &                           31.11 &                             &                           6 &                           18.27 &                           &                         $-$ &                             $-$ &                       $-$ &                         $-$ &                             $-$ &                       $-$ \\
               $\tableFeat{nb\_mean\_ratio}{}{\transnone}{\ftNBC}$ &                          6 &                           38.98 &                             &                           6 &                           32.96 &                           &                         $-$ &                             $-$ &                       $-$ &                         $-$ &                             $-$ &                       $-$ \\
                       $\tableFeat{nb\_cor}{}{\transnone}{\ftNBC}$ &                          6 &                           45.94 &                             &                           6 &                           32.39 &                           &                         $-$ &                             $-$ &                       $-$ &                         $-$ &                             $-$ &                       $-$ \\
                   $\tableFeat{dist\_ratio}{}{\transnone}{\ftNBC}$ &                          6 &                           55.23 &                             &                           6 &                           50.78 &                           &                         $-$ &                             $-$ &                       $-$ &                         $-$ &                             $-$ &                       $-$ \\
              $\tableFeat{nb\_fitness\_cor}{}{\transnone}{\ftNBC}$ &                          6 &                           77.17 &                             &                           6 &                           77.13 &                           &                         $-$ &                             $-$ &                       $-$ &                         $-$ &                             $-$ &                       $-$ \\
\bottomrule
\end{tabular}
\normalsize
}
\end{table}

%% file: tex/feat_knn_1.tex
\newcolumntype{C}[1]{>{\PreserveBackslash\centering}p{#1}}
\newcolumntype{R}[1]{>{\PreserveBackslash\raggedleft}p{#1}}
\newcolumntype{L}[1]{>{\PreserveBackslash\raggedright}p{#1}}
\begin{table}
\centering
\caption{
  \tss{knn} features from feature sets \featBasic, \featCMA, \featDisp, and \featyDis{}
  (only $\trainset$-based and sample set indepent, $\archive$-based are identical to \tss{full} in Table~\ref{tab:featProp_full_1}). Features are grouped according to their feature sets (separated by horizontal lines). Features with less than 25\% of values equal to $\nanout$ and robustness greater than 0.9, are in gray. $N_\nanout$ denotes the lowest measured number of points from which at most 1\% of feature calculations resulted in $\nanout$ ($N_\nanout = 0$ for sample set independent $\feat{}$). The $(\dm_i, \dm_j)$ column shows the pairs of feature dimensions for which the two-sided Wilcoxon signed rank test with the Bonferroni-Holm correction does not reject the hypothesis of equality of median feature values, at the family-wise level 0.05 for each individual feature.}
\label{tab:featProp_knn_1}
\resizebox{\textwidth}{!}{%
\footnotesize
\begin{tabular}{L{25mm}R{5.5mm}R{6.5mm}L{7mm}R{5.5mm}R{6.5mm}L{7mm}R{5.5mm}R{6.5mm}L{7mm}R{5.5mm}R{6.5mm}L{7mm}}
\toprule
{} & \multicolumn{3}{c}{$\trainset$} & \multicolumn{3}{c}{$\trans{\trainset}$} & \multicolumn{3}{c}{$\trainpredset$} & \multicolumn{3}{c}{$\trans{\trainpredset}$} \\
\cmidrule(lr){2-4}
\cmidrule(lr){5-7}
\cmidrule(lr){8-10}
\cmidrule(lr){11-13}
                                                                          &                $N_\nanout$ &                        rob.(\%) &                      $(\dm_i, \dm_j)$ &                 $N_\nanout$ &                        rob.(\%) &          $(\dm_i, \dm_j)$ &                 $N_\nanout$ &                        rob.(\%) &          $(\dm_i, \dm_j)$ &                 $N_\nanout$ &                        rob.(\%) &          $(\dm_i, \dm_j)$ \\
\midrule
                                                  $\tableFeat{dim}{}{}{}$ & \cellcolor[gray]{0.8500} 0 & \cellcolor[gray]{0.8500} 100.00 &             \cellcolor[gray]{0.8500}  &  \cellcolor[gray]{0.8500} 0 & \cellcolor[gray]{0.8500} 100.00 & \cellcolor[gray]{0.8500}  &  \cellcolor[gray]{0.8500} 0 & \cellcolor[gray]{0.8500} 100.00 & \cellcolor[gray]{0.8500}  &  \cellcolor[gray]{0.8500} 0 & \cellcolor[gray]{0.8500} 100.00 & \cellcolor[gray]{0.8500}  \\
                                                  $\tableFeat{obs}{}{}{}$ & \cellcolor[gray]{0.8500} 1 &  \cellcolor[gray]{0.8500} 92.16 &             \cellcolor[gray]{0.8500}  &  \cellcolor[gray]{0.8500} 1 &  \cellcolor[gray]{0.8500} 92.16 & \cellcolor[gray]{0.8500}  & \cellcolor[gray]{0.8500} 13 &  \cellcolor[gray]{0.8500} 92.26 & \cellcolor[gray]{0.8500}  & \cellcolor[gray]{0.8500} 13 &  \cellcolor[gray]{0.8500} 92.26 & \cellcolor[gray]{0.8500}  \\
\midrule
                              $\tableFeat{generation}{}{}{\ftCMA}$ & \cellcolor[gray]{0.8500} 0 & \cellcolor[gray]{0.8500} 100.00 &             \cellcolor[gray]{0.8500}  &  \cellcolor[gray]{0.8500} 0 & \cellcolor[gray]{0.8500} 100.00 & \cellcolor[gray]{0.8500}  &  \cellcolor[gray]{0.8500} 0 & \cellcolor[gray]{0.8500} 100.00 & \cellcolor[gray]{0.8500}  &  \cellcolor[gray]{0.8500} 0 & \cellcolor[gray]{0.8500} 100.00 & \cellcolor[gray]{0.8500}  \\
                              $\tableFeat{step\_size}{}{}{\ftCMA}$ & \cellcolor[gray]{0.8500} 0 & \cellcolor[gray]{0.8500} 100.00 &             \cellcolor[gray]{0.8500}  &  \cellcolor[gray]{0.8500} 0 & \cellcolor[gray]{0.8500} 100.00 & \cellcolor[gray]{0.8500}  &  \cellcolor[gray]{0.8500} 0 & \cellcolor[gray]{0.8500} 100.00 & \cellcolor[gray]{0.8500}  &  \cellcolor[gray]{0.8500} 0 & \cellcolor[gray]{0.8500} 100.00 & \cellcolor[gray]{0.8500}  \\
                                 $\tableFeat{restart}{}{}{\ftCMA}$ & \cellcolor[gray]{0.8500} 0 & \cellcolor[gray]{0.8500} 100.00 &             \cellcolor[gray]{0.8500}  &  \cellcolor[gray]{0.8500} 0 & \cellcolor[gray]{0.8500} 100.00 & \cellcolor[gray]{0.8500}  &  \cellcolor[gray]{0.8500} 0 & \cellcolor[gray]{0.8500} 100.00 & \cellcolor[gray]{0.8500}  &  \cellcolor[gray]{0.8500} 0 & \cellcolor[gray]{0.8500} 100.00 & \cellcolor[gray]{0.8500}  \\
               $\tableFeat{cma\_mean\_dist}{}{\transnone}{\ftCMA}$ &                          6 &                           78.64 &                                       &  \cellcolor[gray]{0.8500} 6 &  \cellcolor[gray]{0.8500} 98.40 & \cellcolor[gray]{0.8500}  & \cellcolor[gray]{0.8500} 13 &  \cellcolor[gray]{0.8500} 93.96 & \cellcolor[gray]{0.8500}  & \cellcolor[gray]{0.8500} 13 &  \cellcolor[gray]{0.8500} 98.71 & \cellcolor[gray]{0.8500}  \\
                        $\tableFeat{evopath\_c\_norm}{}{}{\ftCMA}$ & \cellcolor[gray]{0.8500} 0 & \cellcolor[gray]{0.8500} 100.00 &             \cellcolor[gray]{0.8500}  &  \cellcolor[gray]{0.8500} 0 & \cellcolor[gray]{0.8500} 100.00 & \cellcolor[gray]{0.8500}  &  \cellcolor[gray]{0.8500} 0 & \cellcolor[gray]{0.8500} 100.00 & \cellcolor[gray]{0.8500}  &  \cellcolor[gray]{0.8500} 0 & \cellcolor[gray]{0.8500} 100.00 & \cellcolor[gray]{0.8500}  \\
                        $\tableFeat{evopath\_s\_norm}{}{}{\ftCMA}$ & \cellcolor[gray]{0.8500} 0 & \cellcolor[gray]{0.8500} 100.00 &             \cellcolor[gray]{0.8500}  &  \cellcolor[gray]{0.8500} 0 & \cellcolor[gray]{0.8500} 100.00 & \cellcolor[gray]{0.8500}  &  \cellcolor[gray]{0.8500} 0 & \cellcolor[gray]{0.8500} 100.00 & \cellcolor[gray]{0.8500}  &  \cellcolor[gray]{0.8500} 0 & \cellcolor[gray]{0.8500} 100.00 & \cellcolor[gray]{0.8500}  \\
                      $\tableFeat{cma\_lik}{}{\transnone}{\ftCMA}$ & \cellcolor[gray]{0.8500} 1 &  \cellcolor[gray]{0.8500} 98.81 &             \cellcolor[gray]{0.8500}  &  \cellcolor[gray]{0.8500} 1 &  \cellcolor[gray]{0.8500} 99.95 & \cellcolor[gray]{0.8500}  & \cellcolor[gray]{0.8500} 13 &  \cellcolor[gray]{0.8500} 98.80 & \cellcolor[gray]{0.8500}  & \cellcolor[gray]{0.8500} 13 &  \cellcolor[gray]{0.8500} 99.96 & \cellcolor[gray]{0.8500}  \\
\midrule
              $\tableFeat{ratio\_mean\_02}{}{\transnone}{\ftDisp}$ &                         74 &                           30.53 &                                       &                          74 &                           23.95 &                           &                         109 &                           29.18 &                           &                         109 &                           23.33 &                           \\
            $\tableFeat{ratio\_median\_02}{}{\transnone}{\ftDisp}$ &                         74 &                           34.49 &                              (5,\,10) &                          74 &                           25.97 &                           &                         109 &                           32.08 &                           &                         109 &                           24.94 &                           \\
               $\tableFeat{diff\_mean\_02}{}{\transnone}{\ftDisp}$ &                         74 &                           50.17 &                                       &                          74 &                           96.75 &                           &                         109 &                           50.99 &                           &                         109 &                           96.76 &                           \\
             $\tableFeat{diff\_median\_02}{}{\transnone}{\ftDisp}$ &                         74 &                           49.55 &                                       &                          74 &                           95.54 &                           &                         109 &                           50.37 &                           &                         109 &                           95.47 &                           \\
              $\tableFeat{ratio\_mean\_05}{}{\transnone}{\ftDisp}$ &                         30 &                           28.43 &                                       &                          30 &                           21.63 &                           &                          53 &                           26.56 &                           &                          53 &                           21.19 &                           \\
            $\tableFeat{ratio\_median\_05}{}{\transnone}{\ftDisp}$ &                         30 &                           35.85 &                                       &                          30 &                           24.73 &                  (3,\,10) &                          53 &                           31.45 &                  (2,\,20) &                          53 &                           23.58 &                           \\
               $\tableFeat{diff\_mean\_05}{}{\transnone}{\ftDisp}$ &                         30 &                           55.94 &                                       & \cellcolor[gray]{0.8500} 30 &  \cellcolor[gray]{0.8500} 96.80 & \cellcolor[gray]{0.8500}  &                          53 &                           56.94 &                           & \cellcolor[gray]{0.8500} 53 &  \cellcolor[gray]{0.8500} 96.85 & \cellcolor[gray]{0.8500}  \\
             $\tableFeat{diff\_median\_05}{}{\transnone}{\ftDisp}$ &                         30 &                           55.11 &                                       & \cellcolor[gray]{0.8500} 30 &  \cellcolor[gray]{0.8500} 95.80 & \cellcolor[gray]{0.8500}  &                          53 &                           56.23 &                           & \cellcolor[gray]{0.8500} 53 &  \cellcolor[gray]{0.8500} 95.82 & \cellcolor[gray]{0.8500}  \\
              $\tableFeat{ratio\_mean\_10}{}{\transnone}{\ftDisp}$ &                         15 &                           25.57 & (2,\,3), (2,\,10), (3,\,10), (5,\,10) &                          15 &                           19.13 &                           &                          28 &                           24.33 &                  (5,\,10) &                          28 &                           18.97 &                           \\
            $\tableFeat{ratio\_median\_10}{}{\transnone}{\ftDisp}$ &                         15 &                           32.86 &                                       &                          15 &                           21.96 &                           &                          28 &                           29.68 &                           &                          28 &                           21.49 &                           \\
               $\tableFeat{diff\_mean\_10}{}{\transnone}{\ftDisp}$ &                         15 &                           58.33 &                                       & \cellcolor[gray]{0.8500} 15 &  \cellcolor[gray]{0.8500} 96.96 & \cellcolor[gray]{0.8500}  &                          28 &                           59.42 &                           & \cellcolor[gray]{0.8500} 28 &  \cellcolor[gray]{0.8500} 97.02 & \cellcolor[gray]{0.8500}  \\
             $\tableFeat{diff\_median\_10}{}{\transnone}{\ftDisp}$ &                         15 &                           57.49 &                                       & \cellcolor[gray]{0.8500} 15 &  \cellcolor[gray]{0.8500} 96.09 & \cellcolor[gray]{0.8500}  &                          28 &                           58.85 &                           & \cellcolor[gray]{0.8500} 28 &  \cellcolor[gray]{0.8500} 96.16 & \cellcolor[gray]{0.8500}  \\
              $\tableFeat{ratio\_mean\_25}{}{\transnone}{\ftDisp}$ &                          6 &                           19.39 &                                       &                           6 &                           14.96 &                           &                          15 &                           19.01 &                           &                          15 &                           14.16 &                           \\
            $\tableFeat{ratio\_median\_25}{}{\transnone}{\ftDisp}$ &                          6 &                           23.38 &                                       &                           6 &                           15.76 &                   (3,\,5) &                          15 &                           24.08 &                           &                          15 &                           16.25 &                           \\
               $\tableFeat{diff\_mean\_25}{}{\transnone}{\ftDisp}$ &                          6 &                           62.31 &                                       &  \cellcolor[gray]{0.8500} 6 &  \cellcolor[gray]{0.8500} 97.06 & \cellcolor[gray]{0.8500}  &                          15 &                           62.72 &                           & \cellcolor[gray]{0.8500} 15 &  \cellcolor[gray]{0.8500} 97.12 & \cellcolor[gray]{0.8500}  \\
             $\tableFeat{diff\_median\_25}{}{\transnone}{\ftDisp}$ &                          6 &                           61.67 &                                       &  \cellcolor[gray]{0.8500} 6 &  \cellcolor[gray]{0.8500} 96.27 & \cellcolor[gray]{0.8500}  &                          15 &                           62.60 &                           & \cellcolor[gray]{0.8500} 15 &  \cellcolor[gray]{0.8500} 96.36 & \cellcolor[gray]{0.8500}  \\
\midrule
                     $\tableFeat{skewness}{}{\transnone}{\ftyDis}$ &                          6 &                           30.49 &                                       &                         $-$ &                             $-$ &                       $-$ &                         $-$ &                             $-$ &                       $-$ &                         $-$ &                             $-$ &                       $-$ \\
                     $\tableFeat{kurtosis}{}{\transnone}{\ftyDis}$ &                          6 &                           72.61 &                                       &                         $-$ &                             $-$ &                       $-$ &                         $-$ &                             $-$ &                       $-$ &                         $-$ &                             $-$ &                       $-$ \\
            $\tableFeat{number\_of\_peaks}{}{\transnone}{\ftyDis}$ &                          6 &                           26.63 &                                       &                         $-$ &                             $-$ &                       $-$ &                         $-$ &                             $-$ &                       $-$ &                         $-$ &                             $-$ &                       $-$ \\
\bottomrule
\end{tabular}
\normalsize
}
\end{table}
            
\begin{table}
\catcode`\-=12 
\centering
\caption{
  \tss{knn} features from feature sets \featLevel, \featMM, \featInfo, and \featNBC{}
  (only $\trainset$-based, $\archive$-based are identical to \tss{full} in Table~\ref{tab:featProp_full_2}). Features are grouped according to their feature sets (separated by horizontal lines). Features with less than 25\% of values equal to $\nanout$ and robustness greater than 0.9, are in gray. $N_\nanout$ denotes the lowest measured number of points from which at most 1\% of feature calculations resulted in $\nanout$ ($N_\nanout = 0$ for sample set independent $\feat{}$). The $(\dm_i, \dm_j)$ column shows the pairs of feature dimensions for which the two-sided Wilcoxon signed rank test with the Bonferroni-Holm correction does not reject the hypothesis of equality of median feature values, at the family-wise level 0.05 for each individual feature.}
\label{tab:featProp_knn_2}
\resizebox{\textwidth}{!}{%
\footnotesize
\begin{tabular}{L{25mm}R{5.5mm}R{6.5mm}L{7mm}R{5.5mm}R{6.5mm}L{7mm}R{5.5mm}R{6.5mm}L{7mm}R{5.5mm}R{6.5mm}L{7mm}}
\toprule
{} & \multicolumn{3}{c}{$\trainset$} & \multicolumn{3}{c}{$\trans{\trainset}$} & \multicolumn{3}{c}{$\trainpredset$} & \multicolumn{3}{c}{$\trans{\trainpredset}$} \\
\cmidrule(lr){2-4}
\cmidrule(lr){5-7}
\cmidrule(lr){8-10}
\cmidrule(lr){11-13}
                                                                          &                $N_\nanout$ &                        rob.(\%) &                      $(\dm_i, \dm_j)$ &                 $N_\nanout$ &                        rob.(\%) &          $(\dm_i, \dm_j)$ &                 $N_\nanout$ &                        rob.(\%) &          $(\dm_i, \dm_j)$ &                 $N_\nanout$ &                        rob.(\%) &          $(\dm_i, \dm_j)$ \\
\midrule
               $\tableFeat{mmce\_lda\_10}{}{\transnone}{\ftLevel}$ &                          6 &                           11.39 &                                       &                           6 &                           11.39 &                           &                          13 &                            9.63 &                           &                          13 &                            9.93 &                           \\
               $\tableFeat{mmce\_qda\_10}{}{\transnone}{\ftLevel}$ &                          6 &                           67.83 &                                       &                           6 &                           69.40 &                           &                          13 &                           69.58 &                           &                          13 &                           70.79 &                   (3,\,5) \\
               $\tableFeat{mmce\_mda\_10}{}{\transnone}{\ftLevel}$ &                          6 &                           21.15 &                                       &                           6 &                           20.32 &                           &                          13 &                           19.93 &                           &                          13 &                           19.25 &                           \\
                $\tableFeat{lda\_qda\_10}{}{\transnone}{\ftLevel}$ &                         23 &                           67.38 &                                       &                          23 &                           67.38 &                           &                          37 &                           70.31 &                           &                          37 &                           70.32 &                           \\
                $\tableFeat{lda\_mda\_10}{}{\transnone}{\ftLevel}$ &                          6 &                           15.74 &                                       &                           6 &                           13.69 &                           &                          13 &                           22.17 &                           &                          13 &                           20.80 &                           \\
                $\tableFeat{qda\_mda\_10}{}{\transnone}{\ftLevel}$ &                          6 &                           57.54 &                                       &                           6 &                           56.51 &                           &                          13 &                           55.10 &                           &                          13 &                           55.06 &                           \\
               $\tableFeat{mmce\_lda\_25}{}{\transnone}{\ftLevel}$ &                          6 &                            9.73 &                                       &                           6 &                            9.63 &                           &                          13 &                           15.42 &                  (3,\,20) &                          13 &                           15.65 &                  (3,\,20) \\
               $\tableFeat{mmce\_qda\_25}{}{\transnone}{\ftLevel}$ &                          6 &                           37.27 &                                       &                           6 &                           38.00 &                           &                          13 &                           36.70 &                           &                          13 &                           37.25 &                           \\
               $\tableFeat{mmce\_mda\_25}{}{\transnone}{\ftLevel}$ &                          6 &                           17.79 &                                       &                           6 &                           16.60 &                           &                          13 &                           15.60 &                           &                          13 &                           14.90 &                           \\
                $\tableFeat{lda\_qda\_25}{}{\transnone}{\ftLevel}$ &                          6 &                           73.93 &                              (5,\,10) &                           6 &                           73.87 &                  (5,\,10) &                          18 &                           89.97 &                           &                          18 &                           89.95 &                           \\
                $\tableFeat{lda\_mda\_25}{}{\transnone}{\ftLevel}$ &                          6 &                            8.97 &                                       &                           6 &                            6.33 &                           &                          13 &                            4.46 &                           &                          13 &                            2.95 &                           \\
                $\tableFeat{qda\_mda\_25}{}{\transnone}{\ftLevel}$ &                          6 &                           47.47 &                                       &                           6 &                           49.72 &                           &                          13 &                           35.90 &                           &                          13 &                           39.50 &         (2,\,10), (3,\,5) \\
               $\tableFeat{mmce\_lda\_50}{}{\transnone}{\ftLevel}$ &                          6 &                           17.46 &                                       &                           6 &                           16.62 &                           &                          13 &                            7.86 &                           &                          13 &                            7.88 &                           \\
               $\tableFeat{mmce\_qda\_50}{}{\transnone}{\ftLevel}$ &                          6 &                           34.21 &                                       &                           6 &                           34.32 &                           &                          13 &                           23.54 &                           &                          13 &                           27.64 &                           \\
               $\tableFeat{mmce\_mda\_50}{}{\transnone}{\ftLevel}$ &                          6 &                           24.08 &                                       &                           6 &                           20.56 &                           &                          13 &                           20.74 &                           &                          13 &                           19.07 &                   (2,\,3) \\
                $\tableFeat{lda\_qda\_50}{}{\transnone}{\ftLevel}$ &                          6 &                           41.38 &                                       &                           6 &                           41.37 &                           &                          18 &                           73.19 &                           &                          18 &                           73.01 &                           \\
                $\tableFeat{lda\_mda\_50}{}{\transnone}{\ftLevel}$ &                          6 &                           26.79 &                                       &                           6 &                           32.00 &                           &                          13 &                            2.70 &                  (5,\,10) &                          13 &                            2.30 &                           \\
                $\tableFeat{qda\_mda\_50}{}{\transnone}{\ftLevel}$ &                          6 &                           17.30 &                                       &                           6 &                           24.22 &                           &                          13 &                            5.95 &                           &                          13 &                            7.69 &                  (5,\,20) \\
\midrule
           $\tableFeat{lin\_simple\_adj\_r2}{}{\transnone}{\ftMM}$ &                          6 &                           15.67 &                                       &                           6 &                           15.63 &                           &                         $-$ &                             $-$ &                       $-$ &                         $-$ &                             $-$ &                       $-$ \\
         $\tableFeat{lin\_simple\_coef\_min}{}{\transnone}{\ftMM}$ & \cellcolor[gray]{0.8500} 6 &  \cellcolor[gray]{0.8500} 97.40 &             \cellcolor[gray]{0.8500}  &                           6 &                           63.93 &                           &                         $-$ &                             $-$ &                       $-$ &                         $-$ &                             $-$ &                       $-$ \\
         $\tableFeat{lin\_simple\_coef\_max}{}{\transnone}{\ftMM}$ & \cellcolor[gray]{0.8500} 6 &  \cellcolor[gray]{0.8500} 98.12 &      \cellcolor[gray]{0.8500} (2,\,3) &                           6 &                           87.04 &                           &                         $-$ &                             $-$ &                       $-$ &                         $-$ &                             $-$ &                       $-$ \\
$\tableFeat{lin\_simple\_coef\_max\_by\_min}{}{\transnone}{\ftMM}$ &                          6 &                           34.41 &                                       &                           6 &                           45.65 &                           &                         $-$ &                             $-$ &                       $-$ &                         $-$ &                             $-$ &                       $-$ \\
      $\tableFeat{lin\_w\_interact\_adj\_r2}{}{\transnone}{\ftMM}$ &                        100 &                           37.50 &                                       &                         100 &                           30.25 &                           &                         $-$ &                             $-$ &                       $-$ &                         $-$ &                             $-$ &                       $-$ \\
          $\tableFeat{quad\_simple\_adj\_r2}{}{\transnone}{\ftMM}$ &                          6 &                           41.50 &                                       &                           6 &                           37.91 &                           &                         $-$ &                             $-$ &                       $-$ &                         $-$ &                             $-$ &                       $-$ \\
             $\tableFeat{quad\_simple\_cond}{}{\transnone}{\ftMM}$ & \cellcolor[gray]{0.8500} 6 &  \cellcolor[gray]{0.8500} 92.55 &             \cellcolor[gray]{0.8500}  &                           6 &                           87.21 &                           &                         $-$ &                             $-$ &                       $-$ &                         $-$ &                             $-$ &                       $-$ \\
     $\tableFeat{quad\_w\_interact\_adj\_r2}{}{\transnone}{\ftMM}$ &                        688 &                           69.48 &                                       &                         632 &                           60.02 &                           &                         $-$ &                             $-$ &                       $-$ &                         $-$ &                             $-$ &                       $-$ \\
\midrule
                       $\tableFeat{h\_max}{}{\transnone}{\ftInfo}$ &                          6 &                            1.94 &                                       &                           6 &                            1.57 &                   (3,\,5) &                          13 &                           31.85 &                           &                          13 &                           38.67 &                           \\
                       $\tableFeat{eps\_s}{}{\transnone}{\ftInfo}$ &                          6 &                           38.77 &                                       &                           6 &                            3.58 &                           &                        2424 &                           66.99 &                           &                        2485 &                           70.18 &                  (5,\,10) \\
                     $\tableFeat{eps\_max}{}{\transnone}{\ftInfo}$ & \cellcolor[gray]{0.8500} 6 &  \cellcolor[gray]{0.8500} 97.84 &             \cellcolor[gray]{0.8500}  &                           6 &                           59.04 &                           & \cellcolor[gray]{0.8500} 13 &  \cellcolor[gray]{0.8500} 97.70 & \cellcolor[gray]{0.8500}  &                          13 &                           58.95 &                           \\
                           $\tableFeat{m0}{}{\transnone}{\ftInfo}$ &                          6 &                            0.84 &                                       &                           6 &                            0.44 &                           &                         $-$ &                             $-$ &                       $-$ &                         $-$ &                             $-$ &                       $-$ \\
                   $\tableFeat{eps\_ratio}{}{\transnone}{\ftInfo}$ &                          6 &                           17.19 &                                       &                           6 &                            5.59 &                           &                         $-$ &                             $-$ &                       $-$ &                         $-$ &                             $-$ &                       $-$ \\
\midrule
                $\tableFeat{nb\_std\_ratio}{}{\transnone}{\ftNBC}$ &                          6 &                           20.10 &                                       &                           6 &                           14.20 &                           &                         $-$ &                             $-$ &                       $-$ &                         $-$ &                             $-$ &                       $-$ \\
               $\tableFeat{nb\_mean\_ratio}{}{\transnone}{\ftNBC}$ &                          6 &                           36.38 &                                       &                           6 &                           30.42 &                           &                         $-$ &                             $-$ &                       $-$ &                         $-$ &                             $-$ &                       $-$ \\
                       $\tableFeat{nb\_cor}{}{\transnone}{\ftNBC}$ &                          6 &                           28.49 &                                       &                           6 &                           26.50 &                           &                         $-$ &                             $-$ &                       $-$ &                         $-$ &                             $-$ &                       $-$ \\
                   $\tableFeat{dist\_ratio}{}{\transnone}{\ftNBC}$ &                          6 &                           41.24 &                                       &                           6 &                           26.53 &                           &                         $-$ &                             $-$ &                       $-$ &                         $-$ &                             $-$ &                       $-$ \\
              $\tableFeat{nb\_fitness\_cor}{}{\transnone}{\ftNBC}$ &                          6 &                           45.12 &                                       &                           6 &                           45.14 &                           &                         $-$ &                             $-$ &                       $-$ &                         $-$ &                             $-$ &                       $-$ \\
\bottomrule
\end{tabular}
\normalsize
}
\end{table}

%% file: tex/duelTable_mse.tex
\begin{sidewaystable}
\setlength{\savetabcolsep}{\tabcolsep}
\setlength{\savecmidrulekern}{\cmidrulekern}

\setlength{\headcolw}{0.8cm}
\setlength{\tabcolsep}{0pt}
\setlength{\cmidrulekern}{2pt}
\setlength{\dueltabcolw}{1.3\textwidth-\headcolw-1560\tabcolsep}
\setlength{\dueltabcolw}{\dueltabcolw/39}

\settowidth{\astwidth}{${}^{\ast}$}
\centering
\caption{
  A pairwise comparison of the model settings MSE in different TSS.
The percentage of wins of $i$-th model setting against $j$-th model setting over all available data is given in the $i$-th row and $j$-th column.
  The numbers in bold mark the row model setting being significantly better than the column model setting according to the two-sided Wilcoxon signed rank test with the Holm correction at family-wise significance level $\alpha=0.05$.
}

\label{tab:duelTable_mse}
\resizebox{\textwidth}{!}{%
\footnotesize
\begin{tabular}{ HHHLLLLLLLLLLLLLLLLLLLLLLLLLLLLLLLLLLLLLLL }
\toprule
\multicolumn{2}{c}{\multirow{2}{*}{\textbf{MSE}}} & TSS & \multicolumn{19}{l}{\parbox{19\dueltabcolw}{\centering full}} & \multicolumn{19}{l}{\parbox{19\dueltabcolw}{\centering nearest}} & \multicolumn{1}{l}{\parbox{1\dueltabcolw}{\centering knn}}\\
\cmidrule(lr){4-22}
\cmidrule(lr){23-41}
\cmidrule(lr){42-42}
\multicolumn{2}{c}{} & model & \multicolumn{8}{l}{\parbox{8\dueltabcolw}{\centering GP}} & \multicolumn{9}{l}{\parbox{9\dueltabcolw}{\centering RF}} & \multicolumn{1}{l}{\parbox{1\dueltabcolw}{\centering lmm}} & \multicolumn{1}{l}{\parbox{1\dueltabcolw}{\centering lq}} & \multicolumn{8}{l}{\parbox{8\dueltabcolw}{\centering GP}} & \multicolumn{9}{l}{\parbox{9\dueltabcolw}{\centering RF}} & \multicolumn{1}{l}{\parbox{1\dueltabcolw}{\centering lmm}} & \multicolumn{1}{l}{\parbox{1\dueltabcolw}{\centering lq}} & \multicolumn{1}{l}{\parbox{1\dueltabcolw}{\centering lmm}}\\
\bottomrule
\end{tabular}
}
\setlength{\tabcolsep}{\savetabcolsep}
\setlength{\cmidrulekern}{\savecmidrulekern}
\end{sidewaystable}

%% file: tex/ksTable_mse_full.tex
\begin{table*}[t]
\catcode`\-=12 
\caption{The p-values of the Kolmogorov-Smirnov (KS) test comparing the equality of probability distributions of individual \tss{full} feature representatives on all data and on those data on which a particular model setting scored best in MSE. The p-values are after the Holm correction and they are shown only if the KS test rejects the equality of both distributions at the family-wise significance level $\alpha=0.05$, non-rejecting the equality hypothesis is indicated with ---. Zeros indicate p-values below the smallest double precision number.
}
\ifx\tabcolw\undefined
  \newlength{\tabcolw}
\fi
\ifx\savetabcolsep\undefined
  \newlength{\savetabcolsep}
\fi
\ifx\savecmidrulekern\undefined
  \newlength{\savecmidrulekern}
\fi
\ifx\headcolw\undefined
  \newlength{\headcolw}
\fi
\ifx\groupheadcolw\undefined
  \newlength{\groupheadcolw}
\fi
\setlength{\savetabcolsep}{\tabcolsep}
\setlength{\savecmidrulekern}{\cmidrulekern}

\setlength{\headcolw}{2.2cm}
\setlength{\groupheadcolw}{0cm}
\setlength{\tabcolsep}{0pt}
\setlength{\cmidrulekern}{2pt}
\setlength{\tabcolw}{\textwidth-\groupheadcolw-\headcolw-40\tabcolsep}
\setlength{\tabcolw}{\tabcolw/19}

\ifx\astwidth\undefined
  \newlength{\astwidth}
\fi
\settowidth{\astwidth}{${}^{\ast}$}
\centering
\newcolumntype{R}{>{\raggedleft\arraybackslash}m{\tabcolw}}
\newcolumntype{H}{>{\raggedright\arraybackslash}m{\headcolw}}
\newcolumntype{G}{>{\raggedright\arraybackslash}m{\groupheadcolw}}
\small
\begin{tabular}{ GHRRRRRRRRRRRRRRRRRRRRRRRRRRRRRRRRRRRRRR }
\toprule
{} & $\model$ & \multicolumn{8}{l}{\parbox{8\tabcolw}{\centering GP}} & \multicolumn{9}{l}{\parbox{9\tabcolw}{\centering RF}} & \multicolumn{1}{l}{\parbox{1\tabcolw}{\centering lmm}} & \multicolumn{1}{l}{\parbox{1\tabcolw}{\centering lq}}\\
\cmidrule(lr){2-2}
\cmidrule(lr){3-10}
\cmidrule(lr){11-19}
\cmidrule(lr){20-20}
\cmidrule(lr){21-21}
{} & {settings} & {${{{}}}^\text{Gibbs}$} & {${{{}}}^\text{LIN}$} & {${{{}}}^\text{Mat}$} & {${{{}}}^\text{NN}$} & {${{{}}}^\text{Q}$} & {${{{}}}^\text{RQ}$} & {${{{}}}^\text{SE+Q}$} & {${{{}}}^\text{SE}$} & {${{}}^\text{CART}_\text{MSE}$} & {${{}}^\text{CART}_\text{RDE}$} & {${{}}^\text{OC1}$} & {${{}}^\text{PAIR}_\text{MSE}$} & {${{}}^\text{PAIR}_\text{RDE}$} & {${{}}^\text{SCRT}_\text{MSE}$} & {${{}}^\text{SCRT}_\text{RDE}$} & {${{}}^\text{SUPP}_\text{MSE}$} & {${{}}^\text{SUPP}_\text{RDE}$} & {${}$} & {${}$}\\
\midrule
\multirow{14}{*}{} & \scriptsize{$\tableFeat{dim}{}{}{}$} &  \tiny{2e-101} &  \tiny{\makebox{---}} &  \tiny{3.2e-2} &  \tiny{1.3e-3} &  \tiny{2.9e-5} &  \tiny{1.7e-23} &  \tiny{2.8e-13} &  \tiny{\makebox{---}} &  \tiny{1.3e-59} &  \tiny{1.9e-16} &  \tiny{\makebox{---}} &  \tiny{\makebox{---}} &  \tiny{2.4e-2} &  \tiny{3.9e-2} &  \tiny{1.5e-3} &  \tiny{1.0e-16} &  \tiny{7.4e-3} &  \tiny{4e-115} &  \tiny{7.8e-46}\\[-2pt]
 & \scriptsize{$\tableFeat{obs}{\archive}{}{}$} &  \tiny{1.8e-40} &  \tiny{\makebox{---}} &  \tiny{1.0e-7} &  \tiny{1.7e-7} &  \tiny{7.0e-29} &  \tiny{5.9e-36} &  \tiny{1.6e-21} &  \tiny{\makebox{---}} &  \tiny{2.6e-43} &  \tiny{1.7e-7} &  \tiny{8.4e-11} &  \tiny{\makebox{---}} &  \tiny{2.3e-10} &  \tiny{1.1e-6} &  \tiny{6.0e-15} &  \tiny{1.4e-10} &  \tiny{\makebox{---}} &  \tiny{3.2e-14} &  \tiny{3e-259}\\[-2pt]
 & \scriptsize{$\tableFeat{evopath\_c\_norm}{}{}{\ftCMA}$} &  \tiny{3.5e-7} &  \tiny{3.5e-8} &  \tiny{7.9e-37} &  \tiny{\makebox{---}} &  \tiny{2.6e-69} &  \tiny{7.6e-62} &  \tiny{9.1e-30} &  \tiny{2.1e-10} &  \tiny{9.9e-20} &  \tiny{3.7e-18} &  \tiny{4.5e-4} &  \tiny{\makebox{---}} &  \tiny{1.2e-8} &  \tiny{9.3e-8} &  \tiny{7.0e-18} &  \tiny{2.2e-10} &  \tiny{4.3e-24} &  \tiny{1e-216} &  \tiny{5e-307}\\[-2pt]
 & \scriptsize{$\tableFeat{evopath\_s\_norm}{}{}{\ftCMA}$} &  \tiny{1.3e-88} &  \tiny{2.1e-3} &  \tiny{9.7e-10} &  \tiny{1.5e-25} &  \tiny{3.3e-48} &  \tiny{6.0e-33} &  \tiny{9.2e-17} &  \tiny{7.4e-15} &  \tiny{5.1e-18} &  \tiny{7.0e-13} &  \tiny{6.9e-14} &  \tiny{2.7e-6} &  \tiny{2.6e-5} &  \tiny{3.9e-5} &  \tiny{1.3e-10} &  \tiny{2.1e-14} &  \tiny{6.3e-11} &  \tiny{1.3e-16} &  \tiny{1.1e-13}\\[-2pt]
 & \scriptsize{$\tableFeat{restart}{}{}{\ftCMA}$} &  \tiny{7e-105} &  \tiny{1.5e-7} &  \tiny{9.7e-20} &  \tiny{2.3e-61} &  \tiny{7e-273} &  \tiny{5e-193} &  \tiny{3e-137} &  \tiny{3.5e-46} &  \tiny{3.3e-21} &  \tiny{8.2e-18} &  \tiny{1.7e-60} &  \tiny{\makebox{---}} &  \tiny{2.0e-4} &  \tiny{4.1e-9} &  \tiny{8.7e-98} &  \tiny{5.5e-14} &  \tiny{5.1e-37} &  \tiny{1.9e-12} &  \tiny{2.2e-15}\\[-2pt]
 & \scriptsize{$\tableFeat{step\_size}{}{}{\ftCMA}$} &  \tiny{0} &  \tiny{5.2e-19} &  \tiny{1.5e-62} &  \tiny{2e-123} &  \tiny{1e-267} &  \tiny{9e-183} &  \tiny{6.1e-90} &  \tiny{5.4e-79} &  \tiny{5.0e-53} &  \tiny{1.5e-57} &  \tiny{1.1e-58} &  \tiny{2.0e-20} &  \tiny{3.4e-48} &  \tiny{2.2e-47} &  \tiny{2.5e-79} &  \tiny{5.6e-46} &  \tiny{8.4e-85} &  \tiny{1.3e-28} &  \tiny{9e-194}\\[-2pt]
 & \scriptsize{$\tableFeat{cma\_lik}{\archivepred}{\transcma}{\ftCMA}$} &  \tiny{1.5e-37} &  \tiny{1.8e-2} &  \tiny{8.2e-7} &  \tiny{2.7e-14} &  \tiny{2.9e-30} &  \tiny{2.8e-33} &  \tiny{3.8e-8} &  \tiny{6.6e-14} &  \tiny{1.7e-9} &  \tiny{7.8e-9} &  \tiny{4.3e-8} &  \tiny{2.0e-3} &  \tiny{2.2e-6} &  \tiny{1.6e-5} &  \tiny{9.3e-14} &  \tiny{9.7e-8} &  \tiny{4.5e-12} &  \tiny{6.0e-34} &  \tiny{1.5e-22}\\[-2pt]
 & \scriptsize{$\tableFeat{diff\_median\_02}{\archive}{\transcma}{\ftDisp}$} &  \tiny{0} &  \tiny{2.7e-30} &  \tiny{3e-103} &  \tiny{3e-159} &  \tiny{0} &  \tiny{0} &  \tiny{4e-160} &  \tiny{2e-112} &  \tiny{3.9e-94} &  \tiny{7e-110} &  \tiny{1.1e-91} &  \tiny{6.1e-7} &  \tiny{8.3e-50} &  \tiny{2.3e-38} &  \tiny{3.8e-89} &  \tiny{5.9e-62} &  \tiny{6e-107} &  \tiny{3e-103} &  \tiny{6e-250}\\[-2pt]
 & \scriptsize{$\tableFeat{diff\_median\_10}{\archive}{\transcma}{\ftDisp}$} &  \tiny{0} &  \tiny{1.9e-35} &  \tiny{1e-126} &  \tiny{9e-173} &  \tiny{0} &  \tiny{0} &  \tiny{3e-166} &  \tiny{2e-130} &  \tiny{2.8e-89} &  \tiny{4e-118} &  \tiny{5.0e-93} &  \tiny{4.7e-11} &  \tiny{1.8e-62} &  \tiny{5.6e-53} &  \tiny{2.4e-90} &  \tiny{1.4e-61} &  \tiny{4e-125} &  \tiny{2.5e-82} &  \tiny{0}\\[-2pt]
 & \scriptsize{$\tableFeat{diff\_mean\_05}{\archivepred}{\transcma}{\ftDisp}$} &  \tiny{0} &  \tiny{1.6e-35} &  \tiny{5e-132} &  \tiny{7e-188} &  \tiny{0} &  \tiny{0} &  \tiny{1e-185} &  \tiny{7e-138} &  \tiny{6e-115} &  \tiny{2e-133} &  \tiny{1e-104} &  \tiny{3.1e-9} &  \tiny{3.0e-68} &  \tiny{5.8e-55} &  \tiny{7e-102} &  \tiny{6.5e-81} &  \tiny{5e-129} &  \tiny{2.0e-81} &  \tiny{0}\\[-2pt]
 & \scriptsize{$\tableFeat{diff\_mean\_25}{\archivepred}{\transcma}{\ftDisp}$} &  \tiny{0} &  \tiny{6.2e-40} &  \tiny{3e-157} &  \tiny{6e-186} &  \tiny{0} &  \tiny{0} &  \tiny{2e-181} &  \tiny{4e-142} &  \tiny{3e-102} &  \tiny{8e-130} &  \tiny{7e-102} &  \tiny{3.9e-10} &  \tiny{4.9e-77} &  \tiny{7.2e-60} &  \tiny{1.1e-99} &  \tiny{1.7e-79} &  \tiny{1e-128} &  \tiny{1e-103} &  \tiny{0}\\[-2pt]
 & \scriptsize{$\tableFeat{lda\_qda\_10}{\archivepred}{\transnone}{\ftLevel}$} &  \tiny{9.7e-29} &  \tiny{1.7e-5} &  \tiny{8.2e-18} &  \tiny{3.4e-19} &  \tiny{1.6e-22} &  \tiny{4.4e-5} &  \tiny{7.8e-25} &  \tiny{1.5e-4} &  \tiny{5.4e-4} &  \tiny{5.4e-12} &  \tiny{1.7e-10} &  \tiny{\makebox{---}} &  \tiny{1.0e-11} &  \tiny{1.1e-7} &  \tiny{1.7e-35} &  \tiny{2.5e-3} &  \tiny{1.7e-9} &  \tiny{2.0e-56} &  \tiny{0}\\[-2pt]
 & \scriptsize{$\tableFeat{lda\_qda\_25}{\archivepred}{\transnone}{\ftLevel}$} &  \tiny{6.9e-30} &  \tiny{5.7e-33} &  \tiny{6.2e-31} &  \tiny{6.1e-8} &  \tiny{3.8e-9} &  \tiny{1.2e-12} &  \tiny{2.6e-16} &  \tiny{2.0e-29} &  \tiny{7.2e-14} &  \tiny{1.2e-23} &  \tiny{3.1e-13} &  \tiny{1.5e-6} &  \tiny{1.6e-44} &  \tiny{9.8e-33} &  \tiny{3.9e-27} &  \tiny{1.3e-14} &  \tiny{1.2e-17} &  \tiny{3.8e-14} &  \tiny{3e-181}\\[-2pt]
 & \scriptsize{$\tableFeat{quad\_simple\_cond}{\archive}{\transcma}{\ftMM}$} &  \tiny{7.6e-83} &  \tiny{1.4e-9} &  \tiny{3.5e-25} &  \tiny{8.7e-13} &  \tiny{2e-145} &  \tiny{1.7e-96} &  \tiny{2.9e-46} &  \tiny{2.3e-34} &  \tiny{8.3e-11} &  \tiny{1.1e-11} &  \tiny{3.9e-16} &  \tiny{\makebox{---}} &  \tiny{5.2e-5} &  \tiny{5.5e-3} &  \tiny{7.3e-8} &  \tiny{3.0e-7} &  \tiny{1.9e-13} &  \tiny{7e-235} &  \tiny{1e-127}\\[-2pt]
\bottomrule
\end{tabular}

\setlength{\tabcolsep}{\savetabcolsep}
\setlength{\cmidrulekern}{\savecmidrulekern}
\label{tab:ksTable_mse_full}
\end{table*}

%% file: tex/ksTable_rde_full.tex
\begin{table*}[t]
\catcode`\-=12 
\caption{
  The p-values of the Kolmogorov-Smirnov (KS) test comparing the equality of probability distributions of individual
  \tss{full} feature representatives on all data and on those data on which a particular model setting scored best in RDE.
  The p-values are after the Holm correction and they are shown only if the KS test rejects the equality of both distributions at the family-wise significance level $\alpha=0.05$, non-rejecting the equality hypothesis is indicated with ---.
  Zeros indicate p-values below the smallest double precision number.
}
\ifx\tabcolw\undefined
  \newlength{\tabcolw}
\fi
\ifx\savetabcolsep\undefined
  \newlength{\savetabcolsep}
\fi
\ifx\savecmidrulekern\undefined
  \newlength{\savecmidrulekern}
\fi
\ifx\headcolw\undefined
  \newlength{\headcolw}
\fi
\ifx\groupheadcolw\undefined
  \newlength{\groupheadcolw}
\fi
\setlength{\savetabcolsep}{\tabcolsep}
\setlength{\savecmidrulekern}{\cmidrulekern}

\setlength{\headcolw}{2.2cm}
\setlength{\groupheadcolw}{0cm}
\setlength{\tabcolsep}{0pt}
\setlength{\cmidrulekern}{2pt}
\setlength{\tabcolw}{\textwidth-\groupheadcolw-\headcolw-40\tabcolsep}
\setlength{\tabcolw}{\tabcolw/19}

\ifx\astwidth\undefined
  \newlength{\astwidth}
\fi
\settowidth{\astwidth}{${}^{\ast}$}
\centering
\newcolumntype{R}{>{\raggedleft\arraybackslash}m{\tabcolw}}
\newcolumntype{H}{>{\raggedright\arraybackslash}m{\headcolw}}
\newcolumntype{G}{>{\raggedright\arraybackslash}m{\groupheadcolw}}
\small
\begin{tabular}{ GHRRRRRRRRRRRRRRRRRRRRRRRRRRRRRRRRRRRRRR }
\toprule
{} & $\model$ & \multicolumn{8}{l}{\parbox{8\tabcolw}{\centering GP}} & \multicolumn{9}{l}{\parbox{9\tabcolw}{\centering RF}} & \multicolumn{1}{l}{\parbox{1\tabcolw}{\centering lmm}} & \multicolumn{1}{l}{\parbox{1\tabcolw}{\centering lq}}\\
\cmidrule(lr){2-2}
\cmidrule(lr){3-10}
\cmidrule(lr){11-19}
\cmidrule(lr){20-20}
\cmidrule(lr){21-21}
{} & {settings} & {${{{}}}^\text{Gibbs}$} & {${{{}}}^\text{LIN}$} & {${{{}}}^\text{Mat}$} & {${{{}}}^\text{NN}$} & {${{{}}}^\text{Q}$} & {${{{}}}^\text{RQ}$} & {${{{}}}^\text{SE+Q}$} & {${{{}}}^\text{SE}$} & {${{}}^\text{CART}_\text{MSE}$} & {${{}}^\text{CART}_\text{RDE}$} & {${{}}^\text{OC1}$} & {${{}}^\text{PAIR}_\text{MSE}$} & {${{}}^\text{PAIR}_\text{RDE}$} & {${{}}^\text{SCRT}_\text{MSE}$} & {${{}}^\text{SCRT}_\text{RDE}$} & {${{}}^\text{SUPP}_\text{MSE}$} & {${{}}^\text{SUPP}_\text{RDE}$} & {${}$} & {${}$}\\
\midrule
\multirow{14}{*}{} & \scriptsize{$\tableFeat{dim}{}{}{}$} &  \tiny{6e-112} &  \tiny{3e-182} &  \tiny{4.8e-30} &  \tiny{4.8e-18} &  \tiny{3.0e-14} &  \tiny{1.6e-25} &  \tiny{2.5e-14} &  \tiny{2.2e-8} &  \tiny{\makebox{---}} &  \tiny{2.8e-12} &  \tiny{1.2e-10} &  \tiny{1.1e-17} &  \tiny{1.1e-26} &  \tiny{1.9e-15} &  \tiny{1.5e-22} &  \tiny{6.6e-5} &  \tiny{1.3e-25} &  \tiny{1.5e-17} &  \tiny{4.1e-72}\\[-2pt]
 & \scriptsize{$\tableFeat{obs}{\archive}{}{}$} &  \tiny{2e-286} &  \tiny{2e-152} &  \tiny{6e-308} &  \tiny{5e-103} &  \tiny{3e-232} &  \tiny{4e-111} &  \tiny{8e-175} &  \tiny{7e-179} &  \tiny{1.5e-17} &  \tiny{2.5e-8} &  \tiny{1.9e-2} &  \tiny{1.4e-4} &  \tiny{6.1e-26} &  \tiny{4.0e-20} &  \tiny{6.4e-5} &  \tiny{5.5e-5} &  \tiny{7.1e-6} &  \tiny{4e-120} &  \tiny{4e-280}\\[-2pt]
 & \scriptsize{$\tableFeat{evopath\_c\_norm}{}{}{\ftCMA}$} &  \tiny{2.6e-56} &  \tiny{0} &  \tiny{0} &  \tiny{3e-131} &  \tiny{6e-295} &  \tiny{7e-152} &  \tiny{1e-170} &  \tiny{7e-140} &  \tiny{6.9e-4} &  \tiny{9.5e-6} &  \tiny{2.2e-3} &  \tiny{1.2e-4} &  \tiny{\makebox{---}} &  \tiny{\makebox{---}} &  \tiny{3.9e-6} &  \tiny{1.4e-7} &  \tiny{1.1e-3} &  \tiny{1e-172} &  \tiny{8e-309}\\[-2pt]
 & \scriptsize{$\tableFeat{evopath\_s\_norm}{}{}{\ftCMA}$} &  \tiny{5.6e-69} &  \tiny{8.8e-27} &  \tiny{5.5e-4} &  \tiny{3.1e-62} &  \tiny{9.8e-13} &  \tiny{6.1e-25} &  \tiny{3.7e-16} &  \tiny{9.0e-20} &  \tiny{4.3e-11} &  \tiny{7.3e-4} &  \tiny{3.9e-2} &  \tiny{\makebox{---}} &  \tiny{1.4e-11} &  \tiny{1.7e-5} &  \tiny{1.3e-8} &  \tiny{9.6e-4} &  \tiny{3.3e-7} &  \tiny{4.9e-27} &  \tiny{1.7e-21}\\[-2pt]
 & \scriptsize{$\tableFeat{restart}{}{}{\ftCMA}$} &  \tiny{4.3e-2} &  \tiny{2.2e-77} &  \tiny{1.7e-59} &  \tiny{3.5e-6} &  \tiny{3.1e-9} &  \tiny{1.5e-7} &  \tiny{2.2e-25} &  \tiny{6.6e-37} &  \tiny{5.1e-31} &  \tiny{1.1e-4} &  \tiny{3.8e-2} &  \tiny{\makebox{---}} &  \tiny{2.0e-2} &  \tiny{\makebox{---}} &  \tiny{3.3e-8} &  \tiny{1.0e-4} &  \tiny{\makebox{---}} &  \tiny{1.7e-54} &  \tiny{7e-138}\\[-2pt]
 & \scriptsize{$\tableFeat{step\_size}{}{}{\ftCMA}$} &  \tiny{0} &  \tiny{3e-234} &  \tiny{1e-227} &  \tiny{0} &  \tiny{0} &  \tiny{0} &  \tiny{0} &  \tiny{0} &  \tiny{5.3e-75} &  \tiny{2.3e-23} &  \tiny{1.8e-4} &  \tiny{5.5e-6} &  \tiny{1.3e-68} &  \tiny{2.7e-48} &  \tiny{2.4e-23} &  \tiny{5.0e-40} &  \tiny{3.3e-17} &  \tiny{8.7e-36} &  \tiny{4.7e-87}\\[-2pt]
 & \scriptsize{$\tableFeat{cma\_lik}{\archivepred}{\transcma}{\ftCMA}$} &  \tiny{6.3e-65} &  \tiny{9.4e-19} &  \tiny{1.3e-53} &  \tiny{1.4e-36} &  \tiny{2.1e-74} &  \tiny{1.3e-48} &  \tiny{1.3e-46} &  \tiny{5.4e-38} &  \tiny{\makebox{---}} &  \tiny{\makebox{---}} &  \tiny{\makebox{---}} &  \tiny{\makebox{---}} &  \tiny{7.8e-4} &  \tiny{5.3e-3} &  \tiny{4.2e-4} &  \tiny{4.5e-2} &  \tiny{5.8e-9} &  \tiny{2.1e-46} &  \tiny{5.1e-42}\\[-2pt]
 & \scriptsize{$\tableFeat{diff\_median\_02}{\archive}{\transcma}{\ftDisp}$} &  \tiny{0} &  \tiny{9e-234} &  \tiny{0} &  \tiny{0} &  \tiny{0} &  \tiny{0} &  \tiny{0} &  \tiny{0} &  \tiny{1.2e-88} &  \tiny{1.7e-23} &  \tiny{1.0e-7} &  \tiny{\makebox{---}} &  \tiny{2.7e-72} &  \tiny{7.0e-57} &  \tiny{9.6e-15} &  \tiny{1.8e-37} &  \tiny{3.3e-15} &  \tiny{2e-104} &  \tiny{5e-110}\\[-2pt]
 & \scriptsize{$\tableFeat{diff\_median\_10}{\archive}{\transcma}{\ftDisp}$} &  \tiny{0} &  \tiny{0} &  \tiny{0} &  \tiny{0} &  \tiny{0} &  \tiny{0} &  \tiny{0} &  \tiny{0} &  \tiny{4.8e-88} &  \tiny{3.6e-21} &  \tiny{1.3e-9} &  \tiny{7.6e-6} &  \tiny{1.0e-93} &  \tiny{3.3e-79} &  \tiny{1.8e-13} &  \tiny{1.1e-43} &  \tiny{1.6e-12} &  \tiny{3.8e-76} &  \tiny{1e-130}\\[-2pt]
 & \scriptsize{$\tableFeat{diff\_mean\_05}{\archivepred}{\transcma}{\ftDisp}$} &  \tiny{0} &  \tiny{0} &  \tiny{0} &  \tiny{0} &  \tiny{0} &  \tiny{0} &  \tiny{0} &  \tiny{0} &  \tiny{5.2e-98} &  \tiny{4.2e-22} &  \tiny{1.2e-9} &  \tiny{8.6e-4} &  \tiny{1.9e-93} &  \tiny{4.7e-71} &  \tiny{1.7e-8} &  \tiny{3.6e-47} &  \tiny{1.1e-8} &  \tiny{2.9e-66} &  \tiny{1e-128}\\[-2pt]
 & \scriptsize{$\tableFeat{diff\_mean\_25}{\archivepred}{\transcma}{\ftDisp}$} &  \tiny{0} &  \tiny{0} &  \tiny{0} &  \tiny{0} &  \tiny{0} &  \tiny{0} &  \tiny{0} &  \tiny{0} &  \tiny{1.3e-93} &  \tiny{1.3e-20} &  \tiny{5.5e-11} &  \tiny{6.1e-4} &  \tiny{2e-103} &  \tiny{2.8e-86} &  \tiny{5.5e-9} &  \tiny{2.0e-44} &  \tiny{3.6e-8} &  \tiny{1e-106} &  \tiny{5e-201}\\[-2pt]
 & \scriptsize{$\tableFeat{lda\_qda\_10}{\archivepred}{\transnone}{\ftLevel}$} &  \tiny{6e-221} &  \tiny{1e-268} &  \tiny{0} &  \tiny{1e-105} &  \tiny{0} &  \tiny{4e-167} &  \tiny{2e-241} &  \tiny{8e-223} &  \tiny{6.8e-6} &  \tiny{5.2e-4} &  \tiny{1.7e-2} &  \tiny{1.2e-7} &  \tiny{3.5e-5} &  \tiny{2.5e-9} &  \tiny{5.1e-4} &  \tiny{1.7e-2} &  \tiny{\makebox{---}} &  \tiny{4e-129} &  \tiny{0}\\[-2pt]
 & \scriptsize{$\tableFeat{lda\_qda\_25}{\archivepred}{\transnone}{\ftLevel}$} &  \tiny{1.6e-54} &  \tiny{2.0e-40} &  \tiny{3.4e-90} &  \tiny{7.7e-24} &  \tiny{1.3e-74} &  \tiny{1.4e-35} &  \tiny{6.3e-45} &  \tiny{1.0e-51} &  \tiny{5.2e-5} &  \tiny{1.2e-3} &  \tiny{5.2e-10} &  \tiny{1.8e-18} &  \tiny{3.0e-36} &  \tiny{5.1e-33} &  \tiny{3.3e-3} &  \tiny{7.5e-5} &  \tiny{1.1e-7} &  \tiny{5.5e-25} &  \tiny{4.8e-53}\\[-2pt]
 & \scriptsize{$\tableFeat{quad\_simple\_cond}{\archive}{\transcma}{\ftMM}$} &  \tiny{1e-190} &  \tiny{8.5e-68} &  \tiny{6e-220} &  \tiny{2.1e-64} &  \tiny{0} &  \tiny{2e-177} &  \tiny{3e-180} &  \tiny{4e-148} &  \tiny{7.1e-5} &  \tiny{1.6e-3} &  \tiny{\makebox{---}} &  \tiny{8.8e-3} &  \tiny{4.3e-9} &  \tiny{6.0e-6} &  \tiny{2.6e-4} &  \tiny{2.0e-3} &  \tiny{3.5e-2} &  \tiny{0} &  \tiny{0}\\[-2pt]
\bottomrule
\end{tabular}

\setlength{\tabcolsep}{\savetabcolsep}
\setlength{\cmidrulekern}{\savecmidrulekern}
\label{tab:ksTable_rde_full}
\end{table*}

%% file: tex/ksTable_mse_nearest.tex
\begin{table*}[t]
\catcode`\-=12 
\caption{
  The p-values of the Kolmogorov-Smirnov (KS) test comparing the equality of probability distributions of individual
  \tss{nearest} feature representatives on all data and on those data on which a particular model setting scored best
  in MSE.
  The p-values are after the Holm correction and they are shown only if the KS test rejects the equality of both distributions at the family-wise significance level $\alpha=0.05$, non-rejecting the equality hypothesis is indicated with ---.
  Zeros indicate p-values below the smallest double precision number.
}
\ifx\tabcolw\undefined
  \newlength{\tabcolw}
\fi
\ifx\savetabcolsep\undefined
  \newlength{\savetabcolsep}
\fi
\ifx\savecmidrulekern\undefined
  \newlength{\savecmidrulekern}
\fi
\ifx\headcolw\undefined
  \newlength{\headcolw}
\fi
\ifx\groupheadcolw\undefined
  \newlength{\groupheadcolw}
\fi
\setlength{\savetabcolsep}{\tabcolsep}
\setlength{\savecmidrulekern}{\cmidrulekern}

\setlength{\headcolw}{2.2cm}
\setlength{\groupheadcolw}{0cm}
\setlength{\tabcolsep}{0pt}
\setlength{\cmidrulekern}{2pt}
\setlength{\tabcolw}{\textwidth-\groupheadcolw-\headcolw-40\tabcolsep}
\setlength{\tabcolw}{\tabcolw/19}

\ifx\astwidth\undefined
  \newlength{\astwidth}
\fi
\settowidth{\astwidth}{${}^{\ast}$}
\centering
\newcolumntype{R}{>{\raggedleft\arraybackslash}m{\tabcolw}}
\newcolumntype{H}{>{\raggedright\arraybackslash}m{\headcolw}}
\newcolumntype{G}{>{\raggedright\arraybackslash}m{\groupheadcolw}}
\small
\begin{tabular}{ GHRRRRRRRRRRRRRRRRRRRRRRRRRRRRRRRRRRRRRR }
\toprule
{} & $\model$ & \multicolumn{8}{l}{\parbox{8\tabcolw}{\centering GP}} & \multicolumn{9}{l}{\parbox{9\tabcolw}{\centering RF}} & \multicolumn{1}{l}{\parbox{1\tabcolw}{\centering lmm}} & \multicolumn{1}{l}{\parbox{1\tabcolw}{\centering lq}}\\
\cmidrule(lr){2-2}
\cmidrule(lr){3-10}
\cmidrule(lr){11-19}
\cmidrule(lr){20-20}
\cmidrule(lr){21-21}
{} & {settings} & {${{{}}}^\text{Gibbs}$} & {${{{}}}^\text{LIN}$} & {${{{}}}^\text{Mat}$} & {${{{}}}^\text{NN}$} & {${{{}}}^\text{Q}$} & {${{{}}}^\text{RQ}$} & {${{{}}}^\text{SE+Q}$} & {${{{}}}^\text{SE}$} & {${{}}^\text{CART}_\text{MSE}$} & {${{}}^\text{CART}_\text{RDE}$} & {${{}}^\text{OC1}$} & {${{}}^\text{PAIR}_\text{MSE}$} & {${{}}^\text{PAIR}_\text{RDE}$} & {${{}}^\text{SCRT}_\text{MSE}$} & {${{}}^\text{SCRT}_\text{RDE}$} & {${{}}^\text{SUPP}_\text{MSE}$} & {${{}}^\text{SUPP}_\text{RDE}$} & {${}$} & {${}$}\\
\midrule
\multirow{14}{*}{} & \scriptsize{$\tableFeat{dim}{}{}{}$} &  \tiny{0} &  \tiny{\makebox{---}} &  \tiny{1.9e-7} &  \tiny{8.2e-19} &  \tiny{3.8e-3} &  \tiny{1.8e-9} &  \tiny{5.5e-81} &  \tiny{7.9e-3} &  \tiny{9.7e-30} &  \tiny{5.7e-14} &  \tiny{2.9e-3} &  \tiny{7.2e-3} &  \tiny{2.9e-10} &  \tiny{4.4e-7} &  \tiny{2.5e-13} &  \tiny{\makebox{---}} &  \tiny{1.5e-2} &  \tiny{6.7e-36} &  \tiny{1.5e-77}\\[-2pt]
 & \scriptsize{$\tableFeat{obs}{\trainset}{}{}$} &  \tiny{0} &  \tiny{5.0e-3} &  \tiny{6.1e-6} &  \tiny{6.0e-20} &  \tiny{6.2e-6} &  \tiny{2.2e-9} &  \tiny{3.7e-50} &  \tiny{1.5e-6} &  \tiny{4.5e-35} &  \tiny{4.7e-22} &  \tiny{2.1e-9} &  \tiny{2.0e-5} &  \tiny{8.2e-13} &  \tiny{2.2e-6} &  \tiny{2.3e-24} &  \tiny{9.3e-3} &  \tiny{3.4e-11} &  \tiny{2e-124} &  \tiny{2.7e-12}\\[-2pt]
 & \scriptsize{$\tableFeat{evopath\_c\_norm}{}{}{\ftCMA}$} &  \tiny{5e-131} &  \tiny{6.1e-5} &  \tiny{1.9e-46} &  \tiny{1.1e-7} &  \tiny{7.0e-32} &  \tiny{5.2e-18} &  \tiny{1.6e-33} &  \tiny{1.5e-80} &  \tiny{3.5e-26} &  \tiny{7.9e-19} &  \tiny{3.8e-23} &  \tiny{5.5e-12} &  \tiny{2.7e-19} &  \tiny{2.2e-17} &  \tiny{7.4e-19} &  \tiny{4.7e-30} &  \tiny{9.2e-13} &  \tiny{2.6e-61} &  \tiny{1e-183}\\[-2pt]
 & \scriptsize{$\tableFeat{evopath\_s\_norm}{}{}{\ftCMA}$} &  \tiny{3.2e-25} &  \tiny{2.1e-2} &  \tiny{1.7e-3} &  \tiny{7.9e-10} &  \tiny{2.8e-4} &  \tiny{2.5e-20} &  \tiny{6.2e-21} &  \tiny{1.0e-4} &  \tiny{3.2e-9} &  \tiny{1.1e-5} &  \tiny{2.9e-7} &  \tiny{1.4e-5} &  \tiny{1.6e-7} &  \tiny{8.8e-7} &  \tiny{1.6e-6} &  \tiny{4.6e-6} &  \tiny{3.9e-5} &  \tiny{1.7e-56} &  \tiny{8.3e-7}\\[-2pt]
 & \scriptsize{$\tableFeat{restart}{}{}{\ftCMA}$} &  \tiny{2.4e-69} &  \tiny{2.5e-4} &  \tiny{\makebox{---}} &  \tiny{1.7e-2} &  \tiny{1.2e-3} &  \tiny{2.3e-75} &  \tiny{3.8e-42} &  \tiny{1.0e-45} &  \tiny{2.6e-2} &  \tiny{\makebox{---}} &  \tiny{3.2e-2} &  \tiny{\makebox{---}} &  \tiny{3.3e-2} &  \tiny{\makebox{---}} &  \tiny{\makebox{---}} &  \tiny{7.9e-4} &  \tiny{2.5e-4} &  \tiny{1.1e-42} &  \tiny{7.0e-64}\\[-2pt]
 & \scriptsize{$\tableFeat{step\_size}{}{}{\ftCMA}$} &  \tiny{8e-311} &  \tiny{2.5e-5} &  \tiny{2.4e-21} &  \tiny{1.3e-4} &  \tiny{1.6e-27} &  \tiny{0} &  \tiny{0} &  \tiny{6e-178} &  \tiny{6.3e-13} &  \tiny{1.6e-14} &  \tiny{6.6e-29} &  \tiny{3.1e-8} &  \tiny{2.1e-16} &  \tiny{1.9e-9} &  \tiny{9.7e-25} &  \tiny{1.7e-34} &  \tiny{6.3e-29} &  \tiny{2e-136} &  \tiny{2.0e-78}\\[-2pt]
 & \scriptsize{$\tableFeat{cma\_lik}{\archivepred}{\transcma}{\ftCMA}$} &  \tiny{8.2e-60} &  \tiny{\makebox{---}} &  \tiny{6.1e-9} &  \tiny{2.0e-4} &  \tiny{1.1e-28} &  \tiny{3e-182} &  \tiny{1e-202} &  \tiny{6e-180} &  \tiny{1.0e-6} &  \tiny{1.2e-4} &  \tiny{1.2e-15} &  \tiny{2.6e-7} &  \tiny{1.9e-5} &  \tiny{1.1e-6} &  \tiny{1.4e-11} &  \tiny{1.6e-10} &  \tiny{2.8e-10} &  \tiny{5.1e-83} &  \tiny{3.7e-22}\\[-2pt]
 & \scriptsize{$\tableFeat{diff\_median\_10}{\archive}{\transcma}{\ftDisp}$} &  \tiny{3e-146} &  \tiny{1.9e-10} &  \tiny{4.9e-37} &  \tiny{7.4e-16} &  \tiny{5.2e-57} &  \tiny{0} &  \tiny{0} &  \tiny{0} &  \tiny{3.8e-23} &  \tiny{1.6e-22} &  \tiny{3.8e-38} &  \tiny{4.6e-18} &  \tiny{5.0e-26} &  \tiny{7.3e-23} &  \tiny{1.3e-32} &  \tiny{1.5e-60} &  \tiny{5.9e-45} &  \tiny{0} &  \tiny{2.7e-14}\\[-2pt]
 & \scriptsize{$\tableFeat{diff\_mean\_05}{\archivepred}{\transcma}{\ftDisp}$} &  \tiny{5e-192} &  \tiny{2.4e-10} &  \tiny{4.3e-38} &  \tiny{4.6e-15} &  \tiny{1e-100} &  \tiny{0} &  \tiny{0} &  \tiny{0} &  \tiny{6.3e-34} &  \tiny{1.6e-25} &  \tiny{9.3e-52} &  \tiny{9.9e-27} &  \tiny{5.7e-31} &  \tiny{5.8e-35} &  \tiny{5.2e-43} &  \tiny{1.7e-73} &  \tiny{2.2e-51} &  \tiny{0} &  \tiny{2.3e-19}\\[-2pt]
 & \scriptsize{$\tableFeat{diff\_mean\_25}{\archivepred}{\transcma}{\ftDisp}$} &  \tiny{4e-197} &  \tiny{3.7e-11} &  \tiny{6.3e-47} &  \tiny{5.8e-18} &  \tiny{8e-101} &  \tiny{0} &  \tiny{0} &  \tiny{0} &  \tiny{5.8e-31} &  \tiny{2.3e-25} &  \tiny{5.2e-50} &  \tiny{9.3e-26} &  \tiny{2.0e-30} &  \tiny{5.4e-35} &  \tiny{1.3e-40} &  \tiny{1.0e-79} &  \tiny{8.4e-50} &  \tiny{0} &  \tiny{9.6e-43}\\[-2pt]
 & \scriptsize{$\tableFeat{diff\_median\_02}{\trainset}{\transcma}{\ftDisp}$} &  \tiny{1.2e-26} &  \tiny{1.9e-14} &  \tiny{5.6e-7} &  \tiny{2.0e-33} &  \tiny{4.8e-21} &  \tiny{2e-208} &  \tiny{7e-228} &  \tiny{3e-106} &  \tiny{\makebox{---}} &  \tiny{\makebox{---}} &  \tiny{6.3e-5} &  \tiny{2.5e-3} &  \tiny{\makebox{---}} &  \tiny{\makebox{---}} &  \tiny{1.6e-4} &  \tiny{\makebox{---}} &  \tiny{5.0e-9} &  \tiny{2.7e-81} &  \tiny{2.1e-28}\\[-2pt]
 & \scriptsize{$\tableFeat{lda\_qda\_10}{\archivepred}{\transnone}{\ftLevel}$} &  \tiny{4.1e-35} &  \tiny{2.2e-2} &  \tiny{3.8e-16} &  \tiny{1.4e-20} &  \tiny{\makebox{---}} &  \tiny{2.4e-76} &  \tiny{8e-143} &  \tiny{6.8e-66} &  \tiny{1.8e-4} &  \tiny{1.3e-5} &  \tiny{3.9e-5} &  \tiny{2.0e-7} &  \tiny{3.4e-5} &  \tiny{2.5e-6} &  \tiny{1.2e-2} &  \tiny{2.1e-14} &  \tiny{6.2e-4} &  \tiny{1.3e-21} &  \tiny{4e-150}\\[-2pt]
 & \scriptsize{$\tableFeat{lda\_qda\_25}{\trainpredset}{\transnone}{\ftLevel}$} &  \tiny{9.5e-15} &  \tiny{3.2e-11} &  \tiny{\makebox{---}} &  \tiny{8.0e-14} &  \tiny{2.9e-5} &  \tiny{1.4e-15} &  \tiny{5.7e-23} &  \tiny{2.7e-22} &  \tiny{1.1e-4} &  \tiny{2.6e-5} &  \tiny{6.1e-15} &  \tiny{8.9e-15} &  \tiny{1.6e-4} &  \tiny{2.4e-12} &  \tiny{9.7e-13} &  \tiny{4.5e-16} &  \tiny{3.7e-12} &  \tiny{9e-123} &  \tiny{9.8e-57}\\[-2pt]
 & \scriptsize{$\tableFeat{quad\_simple\_cond}{\trainset}{\transcma}{\ftMM}$} &  \tiny{8.8e-96} &  \tiny{\makebox{---}} &  \tiny{6.8e-6} &  \tiny{1.9e-10} &  \tiny{3.3e-38} &  \tiny{4e-111} &  \tiny{2e-118} &  \tiny{3.0e-90} &  \tiny{2.2e-11} &  \tiny{1.7e-8} &  \tiny{5.2e-8} &  \tiny{5.6e-4} &  \tiny{2.7e-9} &  \tiny{7.4e-3} &  \tiny{9.1e-8} &  \tiny{1.3e-4} &  \tiny{4.4e-5} &  \tiny{2.5e-20} &  \tiny{3.6e-3}\\[-2pt]
\bottomrule
\end{tabular}

\setlength{\tabcolsep}{\savetabcolsep}
\setlength{\cmidrulekern}{\savecmidrulekern}
\label{tab:ksTable_mse_nearest}
\end{table*}

%% file: tex/ksTable_rde_nearest.tex
\begin{table*}[t]
\catcode`\-=12 
\caption{
  The p-values of the Kolmogorov-Smirnov (KS) test comparing the equality of probability distributions of individual \tss{nearest} feature representatives on all data and on those data on which a particular model setting scored best
  in RDE.
  The p-values are after the Holm correction and they are shown only if the KS test rejects the equality of both distributions at the family-wise significance level $\alpha=0.05$, non-rejecting the equality hypothesis is indicated with ---.
  Zeros indicate p-values below the smallest double precision number.
}
\ifx\tabcolw\undefined
  \newlength{\tabcolw}
\fi
\ifx\savetabcolsep\undefined
  \newlength{\savetabcolsep}
\fi
\ifx\savecmidrulekern\undefined
  \newlength{\savecmidrulekern}
\fi
\ifx\headcolw\undefined
  \newlength{\headcolw}
\fi
\ifx\groupheadcolw\undefined
  \newlength{\groupheadcolw}
\fi
\setlength{\savetabcolsep}{\tabcolsep}
\setlength{\savecmidrulekern}{\cmidrulekern}

\setlength{\headcolw}{2.2cm}
\setlength{\groupheadcolw}{0cm}
\setlength{\tabcolsep}{0pt}
\setlength{\cmidrulekern}{2pt}
\setlength{\tabcolw}{\textwidth-\groupheadcolw-\headcolw-40\tabcolsep}
\setlength{\tabcolw}{\tabcolw/19}

\ifx\astwidth\undefined
  \newlength{\astwidth}
\fi
\settowidth{\astwidth}{${}^{\ast}$}
\centering
\newcolumntype{R}{>{\raggedleft\arraybackslash}m{\tabcolw}}
\newcolumntype{H}{>{\raggedright\arraybackslash}m{\headcolw}}
\newcolumntype{G}{>{\raggedright\arraybackslash}m{\groupheadcolw}}
\small
\begin{tabular}{ GHRRRRRRRRRRRRRRRRRRRRRRRRRRRRRRRRRRRRRR }
\toprule
{} & $\model$ & \multicolumn{8}{l}{\parbox{8\tabcolw}{\centering GP}} & \multicolumn{9}{l}{\parbox{9\tabcolw}{\centering RF}} & \multicolumn{1}{l}{\parbox{1\tabcolw}{\centering lmm}} & \multicolumn{1}{l}{\parbox{1\tabcolw}{\centering lq}}\\
\cmidrule(lr){2-2}
\cmidrule(lr){3-10}
\cmidrule(lr){11-19}
\cmidrule(lr){20-20}
\cmidrule(lr){21-21}
{} & {settings} & {${{{}}}^\text{Gibbs}$} & {${{{}}}^\text{LIN}$} & {${{{}}}^\text{Mat}$} & {${{{}}}^\text{NN}$} & {${{{}}}^\text{Q}$} & {${{{}}}^\text{RQ}$} & {${{{}}}^\text{SE+Q}$} & {${{{}}}^\text{SE}$} & {${{}}^\text{CART}_\text{MSE}$} & {${{}}^\text{CART}_\text{RDE}$} & {${{}}^\text{OC1}$} & {${{}}^\text{PAIR}_\text{MSE}$} & {${{}}^\text{PAIR}_\text{RDE}$} & {${{}}^\text{SCRT}_\text{MSE}$} & {${{}}^\text{SCRT}_\text{RDE}$} & {${{}}^\text{SUPP}_\text{MSE}$} & {${{}}^\text{SUPP}_\text{RDE}$} & {${}$} & {${}$}\\
\midrule
\multirow{14}{*}{} & \scriptsize{$\tableFeat{dim}{}{}{}$} &  \tiny{0} &  \tiny{2e-189} &  \tiny{1.2e-31} &  \tiny{1.5e-28} &  \tiny{9.2e-32} &  \tiny{2.6e-15} &  \tiny{7.0e-10} &  \tiny{5.6e-44} &  \tiny{9.5e-7} &  \tiny{2.3e-10} &  \tiny{2.2e-34} &  \tiny{2.4e-26} &  \tiny{8.3e-11} &  \tiny{1.5e-24} &  \tiny{1.4e-8} &  \tiny{4.3e-59} &  \tiny{2.0e-29} &  \tiny{7.8e-54} &  \tiny{2.7e-7}\\[-2pt]
 & \scriptsize{$\tableFeat{obs}{\trainset}{}{}$} &  \tiny{0} &  \tiny{9.3e-50} &  \tiny{1.2e-11} &  \tiny{4.7e-16} &  \tiny{7.0e-54} &  \tiny{1.7e-15} &  \tiny{1.5e-12} &  \tiny{3.1e-48} &  \tiny{2.1e-5} &  \tiny{8.1e-5} &  \tiny{7.5e-20} &  \tiny{4.4e-16} &  \tiny{7.3e-5} &  \tiny{5.3e-16} &  \tiny{9.8e-5} &  \tiny{6.6e-37} &  \tiny{1.0e-15} &  \tiny{4.8e-58} &  \tiny{1.2e-28}\\[-2pt]
 & \scriptsize{$\tableFeat{evopath\_c\_norm}{}{}{\ftCMA}$} &  \tiny{2e-171} &  \tiny{0} &  \tiny{0} &  \tiny{1e-132} &  \tiny{6.0e-83} &  \tiny{5e-159} &  \tiny{4e-202} &  \tiny{2e-284} &  \tiny{3.3e-2} &  \tiny{3.3e-2} &  \tiny{2.2e-5} &  \tiny{\makebox{---}} &  \tiny{3.4e-2} &  \tiny{1.5e-2} &  \tiny{1.9e-3} &  \tiny{4.2e-12} &  \tiny{4.9e-5} &  \tiny{3e-202} &  \tiny{4.5e-13}\\[-2pt]
 & \scriptsize{$\tableFeat{evopath\_s\_norm}{}{}{\ftCMA}$} &  \tiny{1.0e-67} &  \tiny{6.7e-34} &  \tiny{1.0e-17} &  \tiny{2.6e-45} &  \tiny{4.1e-12} &  \tiny{1.1e-9} &  \tiny{5.8e-17} &  \tiny{7.7e-10} &  \tiny{5.3e-3} &  \tiny{2.7e-3} &  \tiny{1.4e-3} &  \tiny{1.3e-3} &  \tiny{2.1e-2} &  \tiny{6.0e-6} &  \tiny{\makebox{---}} &  \tiny{7.3e-5} &  \tiny{1.5e-4} &  \tiny{2.2e-71} &  \tiny{1.5e-59}\\[-2pt]
 & \scriptsize{$\tableFeat{restart}{}{}{\ftCMA}$} &  \tiny{4.0e-93} &  \tiny{2.9e-91} &  \tiny{1.9e-83} &  \tiny{7.1e-43} &  \tiny{1e-146} &  \tiny{2e-280} &  \tiny{0} &  \tiny{2e-322} &  \tiny{1.1e-2} &  \tiny{\makebox{---}} &  \tiny{1.6e-3} &  \tiny{3.2e-4} &  \tiny{3.5e-4} &  \tiny{1.9e-4} &  \tiny{4.4e-2} &  \tiny{5.7e-7} &  \tiny{3.4e-2} &  \tiny{2e-196} &  \tiny{3e-270}\\[-2pt]
 & \scriptsize{$\tableFeat{step\_size}{}{}{\ftCMA}$} &  \tiny{0} &  \tiny{1e-236} &  \tiny{0} &  \tiny{2e-285} &  \tiny{7e-234} &  \tiny{8.6e-80} &  \tiny{3e-139} &  \tiny{2.8e-72} &  \tiny{1.0e-22} &  \tiny{5.3e-22} &  \tiny{1.7e-19} &  \tiny{3.5e-20} &  \tiny{2.5e-17} &  \tiny{9.8e-27} &  \tiny{2.5e-16} &  \tiny{3.9e-22} &  \tiny{1.6e-20} &  \tiny{2e-191} &  \tiny{9e-138}\\[-2pt]
 & \scriptsize{$\tableFeat{cma\_lik}{\archivepred}{\transcma}{\ftCMA}$} &  \tiny{6.9e-64} &  \tiny{2.1e-14} &  \tiny{1.4e-50} &  \tiny{2.5e-31} &  \tiny{2.5e-86} &  \tiny{6.3e-71} &  \tiny{5.2e-95} &  \tiny{1.3e-86} &  \tiny{4.3e-8} &  \tiny{1.9e-8} &  \tiny{9.0e-11} &  \tiny{6.4e-11} &  \tiny{2.1e-8} &  \tiny{7.3e-13} &  \tiny{1.0e-9} &  \tiny{2.0e-11} &  \tiny{3.4e-7} &  \tiny{5.2e-72} &  \tiny{3.4e-12}\\[-2pt]
 & \scriptsize{$\tableFeat{diff\_median\_10}{\archive}{\transcma}{\ftDisp}$} &  \tiny{1e-214} &  \tiny{0} &  \tiny{0} &  \tiny{0} &  \tiny{3e-307} &  \tiny{0} &  \tiny{0} &  \tiny{0} &  \tiny{1.2e-23} &  \tiny{4.5e-24} &  \tiny{1.5e-15} &  \tiny{3.9e-27} &  \tiny{1.2e-15} &  \tiny{5.9e-28} &  \tiny{3.8e-16} &  \tiny{1.5e-15} &  \tiny{4.6e-16} &  \tiny{0} &  \tiny{5e-225}\\[-2pt]
 & \scriptsize{$\tableFeat{diff\_mean\_05}{\archivepred}{\transcma}{\ftDisp}$} &  \tiny{1e-191} &  \tiny{0} &  \tiny{0} &  \tiny{0} &  \tiny{0} &  \tiny{0} &  \tiny{0} &  \tiny{0} &  \tiny{2.6e-22} &  \tiny{8.3e-21} &  \tiny{9.9e-20} &  \tiny{1.6e-29} &  \tiny{3.8e-15} &  \tiny{1.9e-29} &  \tiny{3.8e-19} &  \tiny{9.5e-15} &  \tiny{4.9e-14} &  \tiny{0} &  \tiny{0}\\[-2pt]
 & \scriptsize{$\tableFeat{diff\_mean\_25}{\archivepred}{\transcma}{\ftDisp}$} &  \tiny{6e-197} &  \tiny{0} &  \tiny{0} &  \tiny{0} &  \tiny{0} &  \tiny{0} &  \tiny{0} &  \tiny{0} &  \tiny{2.6e-22} &  \tiny{2.7e-21} &  \tiny{1.7e-20} &  \tiny{1.3e-30} &  \tiny{9.8e-15} &  \tiny{9.5e-31} &  \tiny{3.6e-19} &  \tiny{2.7e-15} &  \tiny{5.0e-14} &  \tiny{0} &  \tiny{2e-317}\\[-2pt]
 & \scriptsize{$\tableFeat{diff\_median\_02}{\trainset}{\transcma}{\ftDisp}$} &  \tiny{1.0e-22} &  \tiny{8e-132} &  \tiny{9.6e-62} &  \tiny{2.9e-84} &  \tiny{1e-107} &  \tiny{5e-262} &  \tiny{0} &  \tiny{4e-256} &  \tiny{\makebox{---}} &  \tiny{1.3e-3} &  \tiny{\makebox{---}} &  \tiny{1.2e-3} &  \tiny{1.1e-3} &  \tiny{5.3e-3} &  \tiny{1.7e-2} &  \tiny{3.4e-5} &  \tiny{9.7e-6} &  \tiny{6e-200} &  \tiny{1e-215}\\[-2pt]
 & \scriptsize{$\tableFeat{lda\_qda\_10}{\archivepred}{\transnone}{\ftLevel}$} &  \tiny{8.2e-52} &  \tiny{3e-266} &  \tiny{0} &  \tiny{2e-116} &  \tiny{9e-130} &  \tiny{2.1e-26} &  \tiny{1.4e-19} &  \tiny{6.8e-31} &  \tiny{7.1e-3} &  \tiny{2.4e-5} &  \tiny{3.5e-3} &  \tiny{1.7e-5} &  \tiny{8.8e-6} &  \tiny{2.8e-4} &  \tiny{\makebox{---}} &  \tiny{1.7e-3} &  \tiny{1.3e-2} &  \tiny{1.8e-49} &  \tiny{5e-107}\\[-2pt]
 & \scriptsize{$\tableFeat{lda\_qda\_25}{\trainpredset}{\transnone}{\ftLevel}$} &  \tiny{6.8e-88} &  \tiny{8.2e-12} &  \tiny{1.0e-22} &  \tiny{2.3e-17} &  \tiny{1.2e-42} &  \tiny{2e-268} &  \tiny{9e-231} &  \tiny{1e-186} &  \tiny{4.0e-6} &  \tiny{1.4e-3} &  \tiny{5.1e-7} &  \tiny{1.3e-10} &  \tiny{6.1e-5} &  \tiny{7.5e-10} &  \tiny{2.7e-9} &  \tiny{2.2e-11} &  \tiny{2.5e-8} &  \tiny{7e-146} &  \tiny{8.2e-85}\\[-2pt]
 & \scriptsize{$\tableFeat{quad\_simple\_cond}{\trainset}{\transcma}{\ftMM}$} &  \tiny{3.4e-30} &  \tiny{3.9e-71} &  \tiny{4.2e-27} &  \tiny{1.7e-22} &  \tiny{3.3e-58} &  \tiny{1.1e-46} &  \tiny{9.8e-55} &  \tiny{4.2e-55} &  \tiny{2.4e-4} &  \tiny{6.2e-4} &  \tiny{\makebox{---}} &  \tiny{1.2e-3} &  \tiny{3.5e-3} &  \tiny{6.8e-4} &  \tiny{\makebox{---}} &  \tiny{\makebox{---}} &  \tiny{\makebox{---}} &  \tiny{1.1e-21} &  \tiny{6.0e-16}\\[-2pt]
\bottomrule
\end{tabular}

\setlength{\tabcolsep}{\savetabcolsep}
\setlength{\cmidrulekern}{\savecmidrulekern}
\label{tab:ksTable_rde_nearest}
\end{table*}

%% file: tex/ksTable_mse_knn.tex
\setlength{\tabcolsep}{0pt}
\newcolumntype{R}{>{\raggedleft\arraybackslash}m{0.85cm}}
\begin{table}[t]
\catcode`\-=12 
\centering
\caption{
  The p-values of the Kolmogorov-Smirnov (KS) test comparing the equality of probability distributions of individual
  \tss{knn} feature representatives on all data and on those data on which the lmm model setting scored best
  in MSE and RDE.
  The p-values are after the Holm correction and they are shown only if the KS test rejects the equality of both distributions at the family-wise significance level $\alpha=0.05$.
  Zeros indicate p-values below the smallest double precision number.
}
\label{tab:ksTable_mse_knn}
\small
\begin{tabular}{RRRRRRRRRRRRRRR}
\toprule
    & \rotatebox{90}{\parbox{2.7cm}{\footnotesize$\tableFeat{dim}{}{}{}$}} & \rotatebox{90}{\parbox{2.7cm}{\footnotesize$\tableFeat{obs}{\archivepred}{}{}$}} & \rotatebox{90}{\parbox{2.7cm}{\footnotesize$\tableFeat{evopath\_s\_norm}{}{}{\ftCMA}$}} & \rotatebox{90}{\parbox{2.7cm}{\footnotesize$\tableFeat{restart}{}{}{\ftCMA}$}} & \rotatebox{90}{\parbox{2.7cm}{\footnotesize$\tableFeat{cma\_lik}{\archivepred}{\transcma}{\ftCMA}$}} & \rotatebox{90}{\parbox{2.7cm}{\footnotesize$\tableFeat{cma\_lik}{\trainpredset}{\transnone}{\ftCMA}$}} & \rotatebox{90}{\parbox{2.7cm}{\footnotesize$\tableFeat{diff\_mean\_10}{\archive}{\transcma}{\ftDisp}$}} & \rotatebox{90}{\parbox{2.7cm}{\footnotesize$\tableFeat{diff\_median\_02}{\archive}{\transcma}{\ftDisp}$}} & \rotatebox{90}{\parbox{2.7cm}{\footnotesize$\tableFeat{diff\_mean\_10}{\trainpredset}{\transcma}{\ftDisp}$}} & \rotatebox{90}{\parbox{2.7cm}{\footnotesize$\tableFeat{diff\_mean\_05}{\trainset}{\transcma}{\ftDisp}$}} & \rotatebox{90}{\parbox{2.7cm}{\footnotesize$\tableFeat{eps\_max}{\trainset}{\transnone}{\ftInfo}$}} & \rotatebox{90}{\parbox{2.7cm}{\footnotesize$\tableFeat{lda\_qda\_10}{\archivepred}{\transnone}{\ftLevel}$}} & \rotatebox{90}{\parbox{2.7cm}{\footnotesize$\tableFeat{lda\_qda\_25}{\archivepred}{\transnone}{\ftLevel}$}} & \rotatebox{90}{\parbox{2.7cm}{\footnotesize$\tableFeat{quad\_simple\_cond}{\trainset}{\transnone}{\ftMM}$}} \\
\midrule
MSE &                                          \tiny{5.2e-32} &                                                      \tiny{9.2e-08} &                                                                    \tiny{1.6e-63} &                                                           \tiny{1.9e-75} &                                                                                 \tiny{1.2e-78} &                                                                                    \tiny{3e-267} &                                                                                          \tiny{0} &                                                                                            \tiny{0} &                                                                                        \tiny{8.3e-126} &                                                                                    \tiny{7.4e-161} &                                                                                      \tiny{0} &                                                                                        \tiny{1.1e-23} &                                                                                        \tiny{2.8e-60} &                                                                                          \tiny{2e-25} \\
RDE &                                          \tiny{9.5e-45} &                                                      \tiny{1.2e-28} &                                                                    \tiny{2.8e-67} &                                                          \tiny{1.4e-196} &                                                                                 \tiny{1.6e-46} &                                                                                         \tiny{0} &                                                                                          \tiny{0} &                                                                                            \tiny{0} &                                                                                        \tiny{2.8e-188} &                                                                                    \tiny{4.9e-209} &                                                                                      \tiny{0} &                                                                                        \tiny{9.2e-14} &                                                                                       \tiny{7.4e-119} &                                                                                       \tiny{6.6e-319} \\
\bottomrule
\end{tabular}
\end{table}
\setlength{\tabcolsep}{\savetabcolsep}

%% file: pitra2022ec.bbl
\newcommand\onlysort[1]{}
\begin{thebibliography}{}

\bibitem[Auger et~al., 2013]{auger2013benchmarking}
Auger, A., Brockhoff, D., and Hansen, N. (2013).
\newblock Benchmarking the local metamodel {CMA-ES} on the noiseless
  {BBOB}'2013 test bed.
\newblock GECCO '13, pages 1225--1232.

\bibitem[Auger and Hansen, 2005]{auger2005restart}
Auger, A. and Hansen, N. (2005).
\newblock A restart {CMA} evolution strategy with increasing population size.
\newblock volume~2 of {\em CEC '05}, pages 1769--1776. IEEE.

\bibitem[Baerns and Hole\v{n}a, 2009]{baerns2009combinatorial}
Baerns, M. and Hole\v{n}a, M. (2009).
\newblock {\em Combinatorial Development of Solid Catalytic Materials. Design
  of High-Throughput Experiments, Data Analysis, Data Mining}.
\newblock Imperial College Press~/~World Scientific, London.

\bibitem[Bajer et~al., 2019]{bajer2019gaussian}
Bajer, L., Pitra, Z., Repick\'{y}, J., and Hole\v{n}a, M. (2019).
\newblock {G}aussian process surrogate models for the {CMA} {E}volution
  {S}trategy.
\newblock {\em Evolutionary Computation}, 27(4):665--697.

\bibitem[Breiman, 1984]{breiman1984classification}
Breiman, L. (1984).
\newblock {\em Classification and regression trees}.
\newblock Chapman \& Hall/CRC.

\bibitem[Breiman, 2001]{breiman2001random}
Breiman, L. (2001).
\newblock Random forests.
\newblock {\em Machine learning}, 45(1):5--32.

\bibitem[B\"{u}che et~al., 2005]{buche2005accelerating}
B\"{u}che, D., Schraudolph, N.~N., and Koumoutsakos, P. (2005).
\newblock Accelerating evolutionary algorithms with {G}aussian process fitness
  function models.
\newblock {\em IEEE Transactions on Systems, Man, and Cybernetics, Part C},
  35(2):183--194.

\bibitem[Chaudhuri et~al., 1994]{chaudhuri1994piecewise}
Chaudhuri, P., Huang, M.-C., Loh, W.-Y., and Yao, R. (1994).
\newblock Piecewise-polynomial regression trees.
\newblock {\em Statistica Sinica}, 4(1):143--167.

\bibitem[Chen and Guestrin, 2016]{chen2016xgboost}
Chen, T. and Guestrin, C. (2016).
\newblock {XGB}oost: A scalable tree boosting system.
\newblock KDD '16, pages 785--794. ACM.

\bibitem[Derbel et~al., 2019]{derbel2019new}
Derbel, B., Liefooghe, A., Verel, S., Aguirre, H., and Tanaka, K. (2019).
\newblock New features for continuous exploratory landscape analysis based on
  the {SOO} tree.
\newblock FOGA '19, pages 72--86. ACM.

\bibitem[Dobra and Gehrke, 2002]{dobra2002secret}
Dobra, A. and Gehrke, J. (2002).
\newblock {SECRET}: A scalable linear regression tree algorithm.
\newblock KDD '02, pages 481--487. ACM.

\bibitem[Flamm et~al., 2002]{flamm2002barrier}
Flamm, C., Hofacker, I.~L., Stadler, P.~F., and Wolfinger, M.~T. (2002).
\newblock {Barrier Trees of Degenerate Landscapes}.
\newblock {\em Zeitschrift f\"{u}r Physikalische Chemie International Journal
  of Research in Physical Chemistry and Chemical Physics}, 216(2):155--173.

\bibitem[Forrester and Keane, 2009]{forrester2009recent}
Forrester, A. and Keane, A. (2009).
\newblock Recent advances in surrogate-based optimization.
\newblock {\em Progress in Aerospace Sciences}, 45:50--79.

\bibitem[Friedman, 2001]{friedman2001greedy}
Friedman, J.~H. (2001).
\newblock Greedy function approximation: A gradient boosting machine.
\newblock {\em The Annals of Statistics}, 29(5):1189--1232.

\bibitem[Gibbs, 1997]{gibbs1997bayesian}
Gibbs, M.~N. (1997).
\newblock {\em Bayesian {G}aussian Processes for Regression and
  Classification}.
\newblock PhD thesis, Department of Physics, University of Cambridge.

\bibitem[Hansen, 2006]{hansen2006cma}
Hansen, N. (2006).
\newblock The {CMA} evolution strategy: A comparing review.
\newblock In {\em Towards a {New} {Evolutionary} {Computation}}, Studies in
  {Fuzziness} and {Soft} {Comp.}, pages 75--102. Springer.

\bibitem[Hansen, 2019]{hansen2019global}
Hansen, N. (2019).
\newblock A global surrogate assisted {CMA-ES}.
\newblock GECCO '19, pages 664--672.

\bibitem[Hansen et~al., 2016]{hansen2016coco}
Hansen, N., Auger, A., Mersmann, O., Tusar, T., and Brockhoff, D. (2016).
\newblock {COCO}: A platform for comparing continuous optimizers in a black-box
  setting.
\newblock arXiv:1603.08785.

\bibitem[Hansen and Ostermeier, 1996]{hansen1996adapting}
Hansen, N. and Ostermeier, A. (1996).
\newblock Adapting arbitrary normal mutation distributions in evolution
  strategies: The covariance matrix adaptation.
\newblock CEC '96., pages 312--317. IEEE.

\bibitem[Hinton and Revow, 1996]{hinton1996using}
Hinton, G.~E. and Revow, M. (1996).
\newblock Using pairs of data-points to define splits for decision trees.
\newblock In {\em Advances in Neural Information Processing Systems}, volume~8,
  pages 507--513. MIT Press.

\bibitem[Hoos et~al., 2018]{hoos2018portfolio}
Hoos, H.~H., Peitl, T., Slivovsky, F., and Szeider, S. (2018).
\newblock Portfolio-based algorithm selection for circuit {QBF}s.
\newblock {CP} '18, pages 195--209. Springer.

\bibitem[Hosder et~al., 2001]{hosder2001polynomial}
Hosder, S., Watson, L., and Grossman, B. (2001).
\newblock Polynomial response surface approximations for the multidisciplinary
  design optimization of a high speed civil transport.
\newblock {\em Optimization and Engineering}, 2:431--452.

\bibitem[Jin, 2011]{jin2011surrogate}
Jin, Y. (2011).
\newblock Surrogate-assisted evolutionary computation: Recent advances and
  future challenges.
\newblock {\em Swarm and Evolutionary Computation}, 1:61--70.

\bibitem[Jin et~al., 2001]{jin2001managing}
Jin, Y., Olhofer, M., and Sendhoff, B. (2001).
\newblock Managing approximate models in evolutionary aerodynamic design
  optimization.
\newblock {CEC} '01, pages 592--599. {IEEE}.

\bibitem[Kern et~al., 2006]{kern2006local}
Kern, S., Hansen, N., and Koumoutsakos, P. (2006).
\newblock Local {Meta}-models for {Optimization} {Using} {Evolution}
  {Strategies}.
\newblock PPSN '06, pages 939--948. Springer.

\bibitem[Kerschke, 2017]{kerschke2017comprehensive}
Kerschke, P. (2017).
\newblock Comprehensive feature-based landscape analysis of continuous and
  constrained optimization problems using the {R}-package flacco.
\newblock {\em ArXiv e-prints}.

\bibitem[Kerschke et~al., 2019]{kerschke2019automated}
Kerschke, P., Hoos, H.~H., Neumann, F., and Trautmann, H. (2019).
\newblock {Automated Algorithm Selection: Survey and Perspectives}.
\newblock {\em Evolutionary Computation}, 27(1):3--45.

\bibitem[Kerschke et~al., 2014]{kerschke2014cell}
Kerschke, P., Preuss, M., Hern{\'a}ndez, C., Sch{\"u}tze, O., Sun, J.-Q.,
  Grimme, C., Rudolph, G., Bischl, B., and Trautmann, H. (2014).
\newblock Cell mapping techniques for exploratory landscape analysis.
\newblock EVOLVE V, pages 115--131. Springer.

\bibitem[Kerschke et~al., 2015]{kerschke2015detecting}
Kerschke, P., Preuss, M., Wessing, S., and Trautmann, H. (2015).
\newblock Detecting funnel structures by means of exploratory landscape
  analysis.
\newblock GECCO '15, pages 265--272.

\bibitem[Kruisselbrink et~al., 2010]{kruisselbrink2010robust}
Kruisselbrink, J., Emmerich, M. T.~M., Deutz, A., and Back, T. (2010).
\newblock A robust optimization approach using {Kriging} metamodels for
  robustness approximation in the {CMA}-{ES}.
\newblock {CEC} '10, pages 1--8. IEEE.

\bibitem[Lee et~al., 2016]{lee2016surrogate}
Lee, H., Jo, Y., Lee, D., and Choi, S. (2016).
\newblock Surrogate model based design optimization of multiple wing sails
  considering flow interaction effect.
\newblock {\em Ocean Engineering}, 121:422--436.

\bibitem[Lunacek and Whitley, 2006]{lunacek2006dispersion}
Lunacek, M. and Whitley, D. (2006).
\newblock The dispersion metric and the {CMA} evolution strategy.
\newblock GECCO '06, pages 477--484.

\bibitem[Mersmann et~al., 2011]{mersmann2011exploratory}
Mersmann, O., Bischl, B., Trautmann, H., Preuss, M., Weihs, C., and Rudolph, G.
  (2011).
\newblock Exploratory landscape analysis.
\newblock GECCO '11, pages 829--836.

\bibitem[Mu\~{n}oz et~al., 2015a]{munoz2015exploratory}
Mu\~{n}oz, M.~A., Kirley, M., and Halgamuge, S.~K. (2015a).
\newblock Exploratory landscape analysis of continuous space optimization
  problems using information content.
\newblock {\em IEEE Transactions on Evolutionary Computation}, 19(1):74--87.

\bibitem[Mu\~{n}oz et~al., 2015b]{munoz2015algorithm}
Mu\~{n}oz, M.~A., Sun, Y., Kirley, M., and Halgamuge, S.~K. (2015b).
\newblock Algorithm selection for black-box continuous optimization problems.
\newblock {\em Inf. Sci.}, 317(C):224--245.

\bibitem[Murthy et~al., 1994]{murthy1994system}
Murthy, S.~K., Kasif, S., and Salzberg, S. (1994).
\newblock A system for induction of oblique decision trees.
\newblock {\em J. Artif. Int. Res.}, 2(1):1--32.

\bibitem[Myers et~al., 2009]{myers2009response}
Myers, R., Montgomery, D., and Anderson-Cook, C. (2009).
\newblock {\em Response Surface Methodology: Process and Product Optimization
  Using Designed Experiments}.
\newblock John Wiley \& Sons, Hob.

\bibitem[Pitra et~al., 2016]{pitra2016doubly}
Pitra, Z., Bajer, L., and Hole{\v{n}}a, M. (2016).
\newblock Doubly trained evolution control for the {S}urrogate {CMA}-{ES}.
\newblock PPSN '16, pages 59--68. Springer.

\bibitem[Pitra et~al., 2017]{pitra2017overview}
Pitra, Z., Bajer, L., Repick\'{y}, J., and Hole\v{n}a, M. (2017).
\newblock Overview of surrogate-model versions of covariance matrix adaptation
  evolution strategy.
\newblock GECCO '17, pages 1622--1629.

\bibitem[Pitra et~al., 2021]{pitra2021interaction}
Pitra, Z., Hanu\v{s}, M., Koza, J., Tumpach, J., and Hole\v{n}a, M. (2021).
\newblock Interaction between model and its evolution control in
  surrogate-assisted {CMA} evolution strategy.
\newblock GECCO '21, pages 528--536.

\bibitem[Pitra et~al., 2018]{pitra2018boosted}
Pitra, Z., Repick{\'y}, J., and Hole{\v{n}}a, M. (2018).
\newblock Boosted regression forest for the doubly trained surrogate covariance
  matrix adaptation evolution strategy.
\newblock ITAT '18, pages 72--79. CEUR.

\bibitem[Pitra et~al., 2019]{pitra2019landscape}
Pitra, Z., Repick\'{y}, J., and Hole\v{n}a, M. (2019).
\newblock Landscape analysis of {G}aussian process surrogates for the
  covariance matrix adaptation evolution strategy.
\newblock GECCO '19, pages 691--699.

\bibitem[Rasheed et~al., 2005]{rasheed2005methods}
Rasheed, K., Ni, X., and Vattam, S. (2005).
\newblock Methods for using surrogate models to speed up genetic algorithm
  oprimization: Informed operators and genetic engineering.
\newblock In {\em Knowledge Incorporation in Evolutionary Computation}, pages
  103--123. Springer.

\bibitem[Rasmussen and Williams, 2006]{rasmussen2006gaussian}
Rasmussen, C.~E. and Williams, C. K.~I. (2006).
\newblock {\em {G}aussian Processes for Machine Learning}.
\newblock {MIT} Press.

\bibitem[Renau et~al., 2020]{renau2020exploratory}
Renau, Q., Doerr, C., Dr{\'{e}}o, J., and Doerr, B. (2020).
\newblock Exploratory landscape analysis is strongly sensitive to the sampling
  strategy.
\newblock PPSN '20, pages 139--153. Springer.

\bibitem[Renau et~al., 2019]{renau2019expressiveness}
Renau, Q., Dr{\'{e}}o, J., Doerr, C., and Doerr, B. (2019).
\newblock Expressiveness and robustness of landscape features.
\newblock GECCO '19, pages 2048--2051.

\bibitem[Renau et~al., 2021]{renau2021towards}
Renau, Q., Dr{\'{e}}o, J., Doerr, C., and Doerr, B. (2021).
\newblock Towards explainable exploratory landscape analysis: Extreme feature
  selection for classifying {BBOB} functions.
\newblock EvoStar '21, pages 17--33.

\bibitem[Saini et~al., 2019]{saini2019automatic}
Saini, B.~S., Lopez-Ibanez, M., and Miettinen, K. (2019).
\newblock Automatic surrogate modelling technique selection based on features
  of optimization problems.
\newblock GECCO '19, pages 1765--1772.

\bibitem[Schweizer and Wolff, 1981]{schweizer1981nonparametric}
Schweizer, B. and Wolff, E.~F. (1981).
\newblock On nonparametric measures of dependence for random variables.
\newblock {\em Annals of Statistics}, 9(4):879--885.

\bibitem[Zaefferer et~al., 2016]{zaefferer2016multi}
Zaefferer, M., Gaida, D., and Bartz-Beielstein, T. (2016).
\newblock Multi-fidelity modeling and optimization of biogas plants.
\newblock {\em Applied Soft Computing}, 48:13--28.

\end{thebibliography}


\newcommand\onlysort[1]{}
\begin{thebibliography}{}

\bibitem[Bajer et~al., 2019]{bajer2019gaussian}
Bajer, L., Pitra, Z., Repick\'{y}, J., and Hole\v{n}a, M. (2019).
\newblock {G}aussian process surrogate models for the {CMA} {E}volution
  {S}trategy.
\newblock {\em Evolutionary Computation}, 27(4):665--697.

\bibitem[Hansen, 2006]{hansen2006cma}
Hansen, N. (2006).
\newblock The {CMA} evolution strategy: A comparing review.
\newblock In {\em Towards a {New} {Evolutionary} {Computation}}, Studies in
  {Fuzziness} and {Soft} {Comp.}, pages 75--102. Springer.

\bibitem[Hansen et~al., 2016]{hansen2016coco}
Hansen, N., Auger, A., Mersmann, O., Tusar, T., and Brockhoff, D. (2016).
\newblock {COCO}: A platform for comparing continuous optimizers in a black-box
  setting.
\newblock arXiv:1603.08785.

\bibitem[Hansen and Ostermeier, 1996]{hansen1996adapting}
Hansen, N. and Ostermeier, A. (1996).
\newblock Adapting arbitrary normal mutation distributions in evolution
  strategies: The covariance matrix adaptation.
\newblock CEC '96., pages 312--317. IEEE.

\bibitem[Kern et~al., 2006]{kern2006local}
Kern, S., Hansen, N., and Koumoutsakos, P. (2006).
\newblock Local {Meta}-models for {Optimization} {Using} {Evolution}
  {Strategies}.
\newblock PPSN '06, pages 939--948. Springer.

\bibitem[Kerschke, 2017a]{kerschke2017automated}
Kerschke, P. (2017a).
\newblock {\em Automated and Feature-Based Problem Characterization and
  Algorithm Selection Through Machine Learning}.
\newblock Dissertation, University of M\"{u}nster.
\newblock Publication status: Published.

\bibitem[Kerschke, 2017b]{kerschke2017comprehensive}
Kerschke, P. (2017b).
\newblock Comprehensive feature-based landscape analysis of continuous and
  constrained optimization problems using the {R}-package flacco.
\newblock {\em ArXiv e-prints}.

\bibitem[Kerschke and Dagefoerde, 2017]{kerschke2017flacco}
Kerschke, P. and Dagefoerde, J. (2017).
\newblock {\em flacco: Feature-Based Landscape Analysis of Continuous and
  Constraint Optimization Problems}.
\newblock {R}-package v. 1.7.

\bibitem[Kerschke et~al., 2015]{kerschke2015detecting}
Kerschke, P., Preuss, M., Wessing, S., and Trautmann, H. (2015).
\newblock Detecting funnel structures by means of exploratory landscape
  analysis.
\newblock GECCO '15, pages 265--272.

\bibitem[Lunacek and Whitley, 2006]{lunacek2006dispersion}
Lunacek, M. and Whitley, D. (2006).
\newblock The dispersion metric and the {CMA} evolution strategy.
\newblock GECCO '06, pages 477--484.

\bibitem[Mersmann et~al., 2011]{mersmann2011exploratory}
Mersmann, O., Bischl, B., Trautmann, H., Preuss, M., Weihs, C., and Rudolph, G.
  (2011).
\newblock Exploratory landscape analysis.
\newblock GECCO '11, pages 829--836.

\bibitem[Mu\~{n}oz et~al., 2015]{munoz2015exploratory}
Mu\~{n}oz, M.~A., Kirley, M., and Halgamuge, S.~K. (2015).
\newblock Exploratory landscape analysis of continuous space optimization
  problems using information content.
\newblock {\em IEEE Transactions on Evolutionary Computation}, 19(1):74--87.

\bibitem[Pitra et~al., 2016]{pitra2016doubly}
Pitra, Z., Bajer, L., and Hole{\v{n}}a, M. (2016).
\newblock Doubly trained evolution control for the {S}urrogate {CMA}-{ES}.
\newblock PPSN '16, pages 59--68. Springer.

\bibitem[Pitra et~al., 2022]{pitra2022landscape}
Pitra, Z., Koza, J., Tumpach, J., and Hole\v{n}a, M. (2022).
\newblock Landscape analysis for surrogate models in the evolutionary black-box
  context.
\newblock {\em arXiv}, 2203.11315:1--34.
\newblock Under review in journal.

\bibitem[Pitra et~al., 2019]{pitra2019landscape}
Pitra, Z., Repick\'{y}, J., and Hole\v{n}a, M. (2019).
\newblock Landscape analysis of {G}aussian process surrogates for the
  covariance matrix adaptation evolution strategy.
\newblock GECCO '19, pages 691--699.

\end{thebibliography}
